\documentclass{article}
\usepackage[nonatbib,preprint]{neurips_2022}

\usepackage[utf8]{inputenc} % allow utf-8 input
\usepackage[T1]{fontenc}    % use 8-bit T1 fonts
\usepackage{url}            % simple URL typesetting
\usepackage{booktabs}       % professional-quality tables
\usepackage{amsfonts}       % blackboard math symbols
\usepackage{nicefrac}       % compact symbols for 1/2, etc.
\usepackage{microtype}      % microtypography
\usepackage{xcolor}         % colors

\usepackage{graphicx}
\usepackage{amsmath}
\usepackage{amssymb}
\usepackage{booktabs}
\usepackage{multicol}
\usepackage{multirow}
\usepackage{tabularx}
\usepackage{diagbox}
\usepackage{makecell}

\usepackage{enumitem}
\usepackage{wrapfig}

\newcolumntype{Y}{>{\centering\arraybackslash}X}
\newcommand\Tstrut{\rule{0pt}{2.4ex}}         % = `top' strut
\usepackage{hyperref}       % hyperlinks

\usepackage[capitalize]{cleveref}
\crefname{section}{Sec.}{Secs.}
\Crefname{section}{Section}{Sections}
\Crefname{table}{Table}{Tables}
\crefname{table}{Tab.}{Tabs.}

\title{Diagnosing and Remedying Shot Sensitivity with Cosine Few-Shot Learners}

\author{
  Davis Wertheimer\thanks{Work performed while a graduate student at Cornell University}\textsuperscript{\-\ \-\ $\dagger$} \\ 
  IBM\\
  \texttt{davis.wertheimer@ibm.com} 
   \And
   Luming Tang\thanks{Equal contribution} \\
   Cornell University \\
   \texttt{lt453@cornell.edu} 
   \And
   Bharath Hariharan \\
   Cornell University \\
   \texttt{bh497@cornell.edu} 
}

\begin{document}

\maketitle

\begin{abstract}
Few-shot recognition involves training an image classifier to distinguish novel concepts at test time using few examples (shot). 
Existing approaches generally assume that the shot number at test time is known in advance. 
This is not realistic, and the performance of a popular and foundational method has been shown to suffer when train and test shots do not match. 
We conduct a systematic empirical study of this phenomenon.
In line with prior work, we find that shot sensitivity is broadly present across metric-based few-shot learners, but in contrast to prior work, larger neural architectures provide a degree of built-in robustness to varying test shot. 
More importantly, a simple, previously known but greatly overlooked class of approaches based on cosine distance consistently and greatly improves robustness to shot variation, 
by removing sensitivity to sample noise. 
We derive cosine alternatives to popular and recent few-shot classifiers, broadening their applicability to realistic settings. 
These cosine models consistently improve shot-robustness, outperform prior shot-robust state of the art, and provide competitive accuracy on a range of benchmarks and architectures, including notable gains in the very-low-shot regime. 
\end{abstract}

\section{Introduction}
\label{sec:intro}

    Modern ConvNets have achieved impressive performance~\cite{he2016deep, ronneberger2015u,ren2015faster} but rely on millions of labelled images~\cite{imgnet, OpenImages2, lin2014microsoft}. 
    In many contexts this high level of data availability cannot be assumed. 
    Rare or fine-grained categories may require prohibitively expensive expert gatherers or annotators, respectively. 
    High-stakes, open-world settings like robotics compound the issue: models will likely need to learn and adapt rapidly in deployment, without waiting for offline data collection or annotation. 
    
    These issues have spawned interest in neural models that rapidly adapt to novel concepts using few labelled examples, or \textit{few-shot learning}.
    In the most highly-studied incarnation of this problem (few-shot image classification),
    a classifier is pre-trained on a set of training \textit{classes}, and then must adapt to a distinct set of \textit{unseen} test classes, with only a few (typically 1 or 5) labelled images representing each novel class. 
    Most approaches to few-shot training involve repeated simulations (\textit{episodes}) of this test-time task: batches include 1 or 5 labelled images, which the network uses to make predictions. 
    Optimizing over repeated trials, networks learn to generalize across classes, eventually including novel ones.
    The episodic training scheme has proven overwhelmingly popular~\cite{finn2017model, snell2017prototypical, lee2019meta, chen2021meta, Wertheimer_2021_CVPR}.
    
\begin{wrapfigure}{r}{0.5\linewidth}
    \vspace{-0.4cm}
      \centering
      \includegraphics[width=\linewidth]{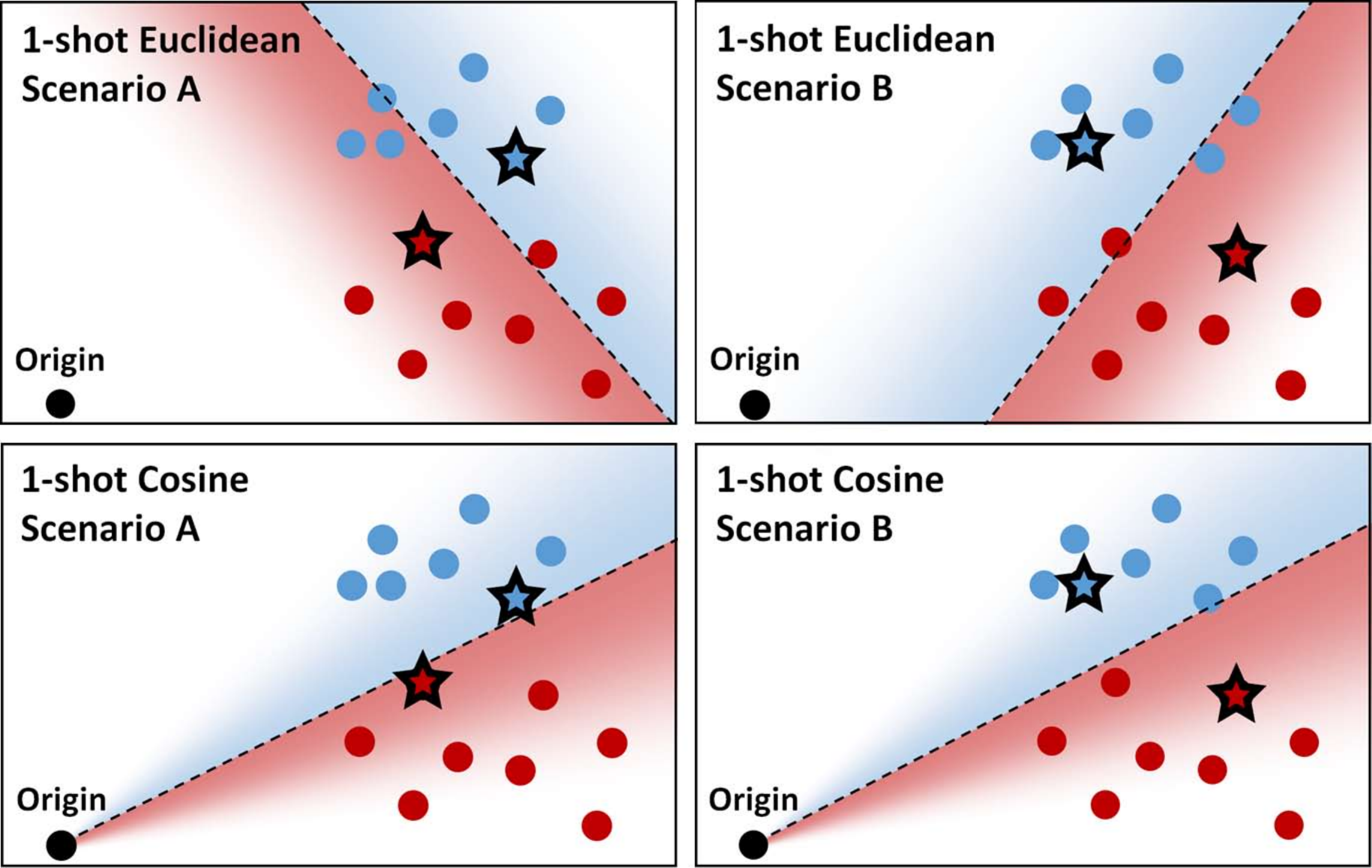}
       \caption{A synthetic illustration of the robustness of cosine decision boundaries to sample noise. In scenarios A and B the starred points represent the classes of interest, and the max-margin decision boundary under the chosen metric is displayed. The Euclidean decision boundary fluctuates wildly as representing points are resampled; cosine is highly stable and accurate. This stability is crucial for ensuring shot-robust behavior at test time.}
       \label{fig:fig1}

    \vspace{-0.4cm}
    \end{wrapfigure}

    Inherent to the episodic training scheme is the assumption that the number of images per class, or \textit{shot}, is known in advance. 
    Models for 5-shot problems are trained on 5-shot episodes; models for 1-shot problems use 1-shot episodes. 
    Clearly this is not realistic, as the practitioner cannot control or know in advance how available as yet unseen data will be in deployment. 
    Prior work shows that this assumption does impact performance. 
    Prototypical networks \cite{snell2017prototypical} (ProtoNet), a simple, foundational, and enduringly popular approach to few-shot classification, overfit severely to training shot and suffer when test shot does not match \cite{est}.
    This is because models trained with different shot numbers require different degrees of clustering to be effective.
    Models trained on low-shot episodes must learn tight class clusters, otherwise different training samples will shift decision boundaries greatly (Fig.~\ref{fig:fig1}, top).
High shot models aggregate class statistics over many images, regressing nicely to the mean, and so learn diffuse clusters that are information-rich but susceptible to sample noise when shot is low \cite{est}.

    We propose a simple yet greatly overlooked solution to the problem of shot-sensitivity: simply replace Euclidean distance logits with the cosine distance. 
    The existence of cosine alternatives to Euclidean-based few-shot learners has long been known - the original ProtoNet paper even contains a discussion of this topic \cite{snell2017prototypical, banerjee2005clustering}. 
    However, when cosine and Euclidean alternatives are compared, they are always evaluated in the episodic setting without shot mismatch, and reported differences are generally small \cite{snell2017prototypical, ye2020fewshot}. 
    We find that when train and test shot differ, these differences become dramatic: cosine models are much more robust to shot variation at test time. 
    This is because the cosine distance provides ample robustness to sample noise, by anchoring all decisions boundaries through the origin. 
    This makes decision boundaries more stable (Fig.~\ref{fig:fig1}, bottom), and by training in this stabilized regime, cosine models become effectively agnostic to shot.

    Motivated by this finding, we implement cosine alternatives to two popular few-shot classifiers (ProtoNet~\cite{snell2017prototypical} and FEAT~\cite{ye2020fewshot}), and derive a novel alternative for a recent state of the art (FRN~\cite{Wertheimer_2021_CVPR}). 
    We then conduct a systematic comparison across benchmarks and network backbones. 
    We find that:
    \begin{enumerate}[leftmargin=1.0em, topsep=0.0mm, itemsep=0.0mm, parsep=0.0mm]
        \item Similar to \cite{est}, overfitting to training shot is a consistent problem for metric few-shot learning. %techniques. 
        \item Contrary to \cite{est}, larger feature extractor %architectures 
        backbones offer some degree of built-in shot robustness. 
        \item Our cosine alternatives provide consistent and superior test time shot sensitivity reduction...
        \item ...without sacrificing model accuracy in expectation, and in fact frequently yield substantial (up to double-digit) accuracy gains in the very-low-shot regime. 

    \end{enumerate}
    
    These results suggest that the shot-overfitting problem, while ubiquitous to few-shot learning, also has a simple, elegant, and widely applicable solution. 
\section{Background and Prior Work}
\label{sec:prior}
    \subsection{Few-Shot Classification}
    
        In the few-shot classification problem we are given labelled images from training classes, and asked to train a classifier that can adapt at test time to a disjoint set of novel classes, using only a few examples.  
Typical \textit{episodic} training involves sampling tasks from an overarching task distribution - usually, by selecting subsets of the training classes to populate each batch or episode. 
        Episodes are partitioned into a small \textit{support set} and larger \textit{query set}, and the classifier uses labels from the support set to classify the query set. 
        During training, query set performance produces a loss used to optimize the model. 
        At test time, queries are used to evaluate model accuracy. 
        The number of support images per class is referred to as the \textit{shot}, while the number of classes in an episode is the \textit{way}. 
        Thus an episode with five classes and a single labelled image for each class is a \textit{5-way/1-shot} problem. 
        
        At this point many approaches exist for tackling few-shot classification, broadly falling into two categories. \textit{Optimization-based} approaches learn a good initialization for the classifier, such that rapid adaptation at test time over few examples is effective \cite{finn2017model, rusu2018meta, nichol2018first}. \textit{Metric-based} approaches, meanwhile, learn a class-agnostic feature extractor such that under a chosen distance metric, images with high semantic similarity cluster together and become separable from those without. 
        Established metrics include Euclidean distance \cite{snell2017prototypical, ye2020fewshot}, cosine distance \cite{vinyals2016matching, gidaris2018dynamic}, hyperbolic distance \cite{khrulkov2020hyperbolic}, Earth Mover's Distance \cite{Zhang_2020_CVPR}, projection distance \cite{simon2020adaptive, Wertheimer_2021_CVPR} and many others. 
        For a more detailed overview of the many approaches to few-shot classification we refer the reader to \cite{wang2020generalizing}. 
        
        We focus on the metric family of approaches, as these have proven both popular and successful at the few-shot classification task. A representative baseline is ProtoNet~\cite{snell2017prototypical}, where each class is represented by the average latent embedding over all the images belonging to that class. 
        We begin our investigation with ProtoNet and then expand to more sophisticated architectures (see Sec.~\ref{sec:method}).

    \subsection{Shot-Robustness}
        Despite the obvious impracticality of knowing test shot in advance, prior research on robustness to shot variation is limited. 
        Most relevant is \cite{est}, which
        shows that ProtoNet overfits to training shot because different shots demand different ratios of inter- to intra-class variance.
        Low-shot models cluster points tightly to overcome sample noise; test-time performance saturates rapidly with respect to shot, as class centroids regress to the mean and effectively ignore further samples. 
        High-shot models tolerate high intra-class variance, as the average prototype over many images regresses nicely to the mean. The result is a higher ceiling for performance but a lower floor, as prototypes do not converge stably when test shot is low. 
        \cite{est} proposes a test-time projection layer (EST) maximizing the inter-/intra-class variance ratio, removing these concerns and serving as our competitive baseline. 
        
        Both \cite{laenen2021} and \cite{goldblum2020unraveling} also analyze training shot as a hyperparameter, with \cite{goldblum2020unraveling} again identifying clustering behavior as the mediating mechanism. 
        But both compare to non-episodic training and ignore shot mismatch.  
        \cite{wertheimer2019few} investigates generalization of few-shot metric-based approaches to much higher way and shot, but does not disentangle these factors or identify any shot-overfitting behavior, besides generally lower performance on a more difficult task. 
        In our work we undertake a systematic evaluation of shot mismatch across architectures, benchmarks, and metric-based approaches. 
        We find that cosine metrics are consistently effective at reducing sensitivity to test-time shot variation. 
        
    \subsection{Cosine Models}
        Cosine alternatives to Euclidean-based metric approaches have been known since the earliest formulations of the few-shot image recognition problem \cite{snell2017prototypical, vinyals2016matching}. Cosine models with Euclidean counterparts have also been proposed \cite{gidaris2018dynamic, chen2021meta}. This duality has been explored in more recent approaches \cite{ye2020fewshot}. To date, all comparisons have been based on accuracy, and are performed in the standard episodic setting, where train and test shot match. Differences are generally small, and the model with slightly better results is reported \cite{snell2017prototypical, ye2020fewshot, gidaris2018dynamic, goldblum2020unraveling}. 
        While our results do not contradict these findings, we discover vast yet previously overlooked differences in behavior when train and test shot are not the same. We also extend the cosine distance beyond simple latent point-to-point comparisons, deriving a novel cosine alternative to the recently proposed FRN~\cite{Wertheimer_2021_CVPR}. The following section provides further detail. 
    
\section{Method}
\label{sec:method}

    We focus on three popular and/or recent metric techniques: ProtoNet~\cite{snell2017prototypical}, FEAT~\cite{ye2020fewshot}, and FRN~\cite{Wertheimer_2021_CVPR}. We describe each model and our corresponding cosine alternatives in the following sections. 
    
    \subsection{ProtoNet}
    \label{sec:proto}
        ProtoNet is a popular and straightforward few-shot classification method. 
        Given a set of $nk$ labelled support images representing $n$ classes of interest, the ProtoNet takes the latent embeddings for the $k$ images in each class and computes an average class centroid, or prototype.
        Class probability logits are assigned to query images based on the negative squared Euclidean distance between query image embeddings and class centroids.
        We multiply the logits by a learned temperature scaling factor before the softmax layer, as prior work has shown this to have a large positive impact on performance~\cite{gidaris2018dynamic, chen2021meta, ye2020fewshot, Wertheimer_2021_CVPR}. 
        Since squared Euclidean distance grows linearly with latent dimensionality, we also normalize logits by the network width to ensure consistent performance across architectures. 
        Thus if $\mu_c$ is the class centroid for class $c$, $C$ is the set of support classes, $\sigma$ is the learned temperature scaling term, and $d$ is the network width, then ProtoNet predictions are given by:
        \begin{equation}
        \label{eq:protopred}
            p(y|x) = \frac{\exp(-\frac{\sigma}{d} ||x-\mu_y||_2^2)}{\sum_{c\in C} \exp(-\frac{\sigma}{d} ||x-\mu_c||_2^2)}
        \end{equation}
        
        \begin{wraptable}{r}{0.6\linewidth}
        \vspace{-0.4cm}
        \caption{5-way/1-shot ProtoNet prediction consistency: when support images change, what proportion of predictions are preserved? Rows represent test-time prediction mechanism; columns represent training scheme. Cosine prediction is more robust to sample noise, and cosine training shows reduced sensitivity to shot value.}
        \vspace{.1cm}
          \centering
          \footnotesize
          \begin{tabular}{l|c c c c}
            \toprule
            \multirow{2}{*}{\backslashbox{Prediction}{Training}} & \multicolumn{2}{c}{Euclidean} & \multicolumn{2}{c}{Cosine} \\
            \-& 4-shot & 32-shot & 4-shot & 32-shot\\
            \midrule
            Euclidean & 48\% & 32\% & 37\% & 32\% \\
            Cosine & 53\% & 40\% & 55\% & 51\% \\
            \bottomrule
          \end{tabular}
          \label{tab:consistency}
          \vspace{-0.4cm}
        \end{wraptable}

        The cosine alternative to this model is straightforward. 
        We remove the $1/d$ normalization since cosine distance does not grow with dimensionality, but keep the learned temperature scale, as cosine distance on its own is bounded by $\pm 1$. 
        Cosine ProtoNet predictions are then given by:
        \begin{equation}
        \label{eq:cospred}
            p(y|x) = \frac{\exp(\sigma \bar{x}^T\bar{\mu_y})}{\sum_{c\in C} \exp(\sigma \bar{x}^T\bar{\mu_c})}
        \end{equation}
        where $\bar{v}$ represents L2 normalization on the vector $v$.

        As a preliminary experiment, we verify our intuition from Fig.~\ref{fig:fig1} that cosine distance stabilizes decision boundaries and unifies behavior across training shot regimes.
        Given a 5-way 1-shot episode, we observe the class predictions on query points from a simple 4-layer ProtoNet. 
        We then resample just the support images, make a new set of predictions, and count how many change. 
        If decision boundaries are stable and resistant to sample noise, then few predictions should change. 
        After 1000 trials, agreement rates are compared to the same model but with cosine predictions (Eq.~\ref{eq:cospred}). 
        To avoid model bias, we repeat the comparison across ProtoNets under four training regimes: Euclidean vs cosine, and 4- vs 32-shot. 
        Models are trained on meta-iNat (details in Sec.~\ref{sec:details}) and results appear in Table~\ref{tab:consistency}.
        Our intuition is correct: cosine predictions are more consistent than Euclidean across all models, regardless of training (bottom row vs top row). 
        We also see that cosine training stabilizes model behavior across different shots. 
        Agreement rates for 4- vs 32-shot Euclidean models vary widely (col 1 vs 2), while agreement rates for cosine models are much closer (col 3 vs 4). 
        Motivated by these findings, we now derive cosine alternative to more sophisticated few-shot learners.

    \subsection{FEAT}
        FEAT~\cite{ye2020fewshot}, or Few-shot Embedding Adaptation with Transformer, is a more recent and powerful approach based on ProtoNet. 
        FEAT task-conditions the prototypes by passing them through a learned attention layer. 
        Predictions are then made using Eq.~\ref{eq:protopred}, but with the refined centroids. 
        The original paper performs a comparison to the cosine alternative using Eq.~\ref{eq:cospred}, and finds only small differences in performance. 
        This is not surprising, as we find that the implementation of FEAT's attentional transformer largely trivializes this distinction. 
        The transformer head refines the centroids and then uses LayerNorm before classification; this operation recenters and L2-normalizes the input. 
        To restore the distinction between Euclidean and cosine FEAT, therefore, we remove the LayerNorm from the FEAT attention head. 
        This change did not greatly impact performance - we found that LayerNorm scores generally interpolate between cosine and Euclidean scores.

    \subsection{FRN}
    \label{sec:frn}
        Feature map Reconstruction Networks (FRN) are a more recent state-of-the-art approach recasting the classification problem as reconstruction in feature space \cite{Wertheimer_2021_CVPR}. 
        Rather than compare individual feature vectors, FRN aggregates feature maps into pools of features for each class, and attempts to reconstruct query features as weighted sums over features in the individual support pools. 
        The support pool with the best reconstruction is the most likely class candidate. 
        Reconstruction quality is measured via mean squared error from the target query features. 
        FRN is chosen as a high performer, but also illustrates that cosine methods can generalize broadly beyond simple point to point comparisons. 
        
        FRN formulates the reconstruction problem as linear regression: given a matrix of class $c$ support features $S_c$, and a matrix of query features $Q$, find $\bar{W}$ such that $\bar{W}S_c \approx Q$. 
        For practical reasons we replace the original FRN regularizer, so our latent reconstruction problem is formulated as follows:
        
        \begin{equation}
        \begin{aligned}
        \label{eq:mainrecon}
            \bar W = \underset{W}{\text{arg min\ \ }} ||Q-WS_c||_2^2 + (\lambda||S_c^TS_c||_2) ||W||_2^2
        \end{aligned}
        \end{equation}
        where $||\cdot||_2$ is the Frobenius norm and $\lambda$ a learned weight. 
        Discussion is provided in appendix~\ref{sec:cosfrn}.
        
        Deriving a cosine FRN is not straightforward. Calculating reconstruction quality with cosine distance leads to poor convergence: the reconstruction problem in Eq.~\ref{eq:mainrecon} is optimized for mean squared error. 
        Instead, we reformulate FRN as a comparison between support and query covariance matrices. 
        Let $\Sigma_S = S_c^TS_c$ and $\Sigma_Q = Q^TQ$, and let the regularizer matrix $\Lambda = (\lambda||S_c^TS_c||_2)I$. 
        \cite{Wertheimer_2021_CVPR} uses the Normal Equation to efficiently solve the reconstruction in closed form. 
        Plugging in Eq.~\ref{eq:mainrecon} yields the following FRN logit formula, 
        where $\otimes$ is elementwise multiplication (derivation in appendix~\ref{sec:cosfrn}):
        \begin{equation}
        \begin{aligned}
            \bar{W} &= Q(\Sigma_S+\Lambda)^{-1}S_c^T\\ 
            z(Q;S_c) &= -||\bar{W}S_c-Q||^2\\
            &= 2||(\Sigma_S+\Lambda)^{-1}\Sigma_S \otimes \Sigma_Q||_1 - tr(\Sigma_Q) - tr((\Sigma_S+\Lambda)^{-1}\Sigma_S\Sigma_Q\Sigma_S(\Sigma_S+\Lambda)^{-1})
        \end{aligned}
        \end{equation}
        
        Most of the terms in the last line can be discarded. 
        The second term $tr(\Sigma_Q)$ is constant across all classes and so vanishes in the softmax layer, and in practice the third term is almost exactly proportional to the first. 
        For example, in a 4-layer FRN model trained with shot 4 on meta-iNat, 1000 5-way 5-shot predictions yielded a term-1 to term-2 ratio of 2.026 with standard deviation only .059. 
        Comparing class logits for each query, the standard deviation shrinks further to .009. 
        We therefore use only the first term, with no impact on performance (discussion in appendix~\ref{sec:frncompare}):
        \begin{equation}
        \label{eq:frnfinal}
            z(Q;S_c) = ||(\Sigma_S+\Lambda)^{-1}\Sigma_S \otimes \Sigma_Q||_1
        \end{equation}
        
        This formula is a dot product of flattened normalized covariance matrices. The cosine alternative is now clear: replace the FRN normalization with L2 normalization (or Frobenius norm, for matrices):
        \begin{equation}
            z(Q;S_c) = ||\frac{\Sigma_S}{||\Sigma_S||_2}\otimes \frac{\Sigma_Q}{||\Sigma_Q||_2}||_1
        \end{equation}
        
        Like other cosine models, the novel Cosine FRN provides consistent improvement in shot robustness while maintaining competitive accuracy. These results are detailed in the following section.

\section{Results}
\label{sec:results}
    We undertake a systematic empirical study across benchmarks, network backbones, learning approaches, and training shot values.  
    Classifiers perform as expected, with FRN outperforming FEAT, and ProtoNet trailing behind (see appendix~\ref{sec:acctables}). 
    Cosine alternatives match or reduce shot-sensitivity relative to their Euclidean counterparts in all settings, while providing a notable bump in very-low-shot accuracy (though frequently trading off high-shot accuracy to a lesser degree).
    
    \subsection{Experimental Setup}
    \label{sec:details}
    
        We perform our comparisons on two standard network backbones (Conv-4 and ResNet-12), and four popular few-shot benchmarks (CUB~\cite{WahCUB_200_2011, tang2020revisiting}, meta-iNat~\cite{van2018inaturalist, wertheimer2019few}, mini-ImageNet~\cite{vinyals2016matching} and tiered-ImageNet~\cite{ren18fewshotssl}). 
        The backbones match \cite{lee2019meta, ye2020fewshot, Wertheimer_2021_CVPR} and our hyperparameters largely follow \cite{Wertheimer_2021_CVPR}. Dataset, implementation and training details can be found in appendix~\ref{sec:appdetails}.
        
    \subsection{Measuring Shot Sensitivity}
        For each dataset and classifier, we train four networks with different shot numbers (4, 8, 16, 32).
        Each model is then evaluated on a range of test shots (1, 2, 4, 8, 16, 32) to measure sensitivity to test shot variation. 
        All test-time evaluation is 5-way and results are averaged over 10K trials. 
        We do not report 95\% confidence intervals for each result, as they are uniformly small (about .20, .17, .14, .12, .11, .10 for 1, 2, 4, 8, 16, 32 test shot respectively, regardless of model or setting).
        We employ batch folding \cite{wertheimer2019few} during training, 
        to ensure that batch sizes do not explode as shot increases. 
        For practical reasons we hold batch size constant at 256, so as shot increases, way decreases. 
        This ensures that all models receive the same amount of data and have the same maximum number of updates. 
        
        \begin{figure}
        \centering
        \includegraphics[width=\linewidth]{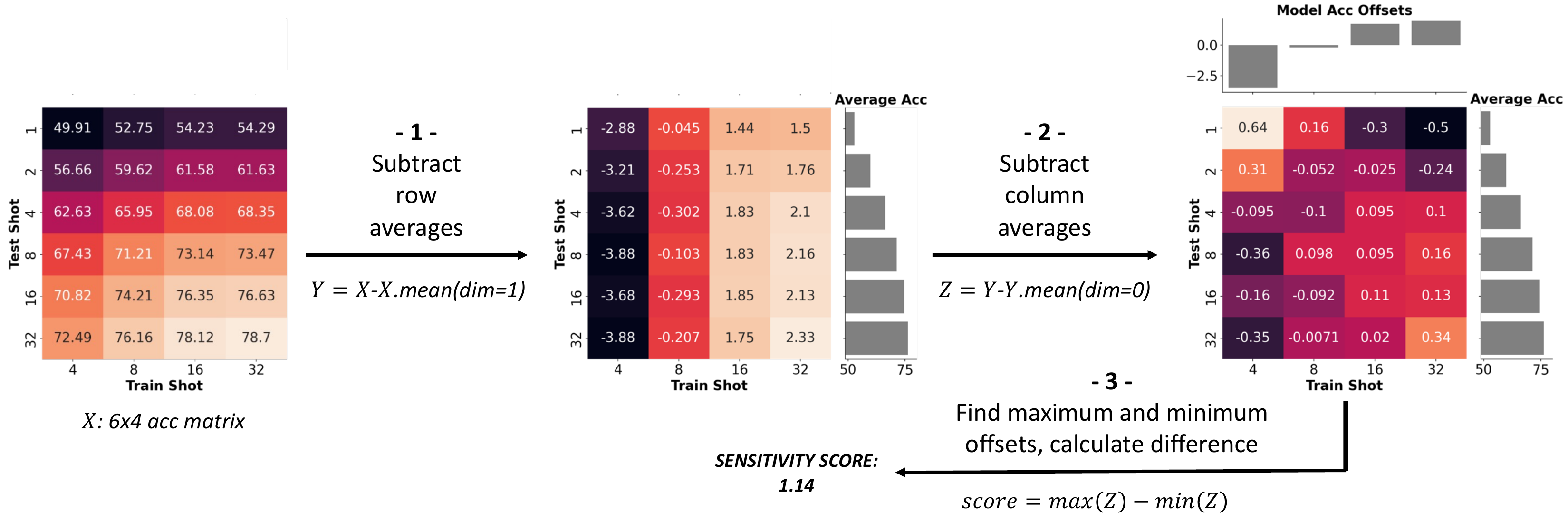}
        \caption{The process for calculating our shot sensitivity scores and heatmaps. The raw scores (left) are dominated by test shot bias: as shot increases, accuracy increases. The model collection's mean accuracy is therefore subtracted from each row to produce accuracy offsets, with subtracted averages displayed to the side (middle). Now behavior is dominated by model bias: certain members of the group are always better than others. The model bias from each column is therefore subtracted to produce a corrected heatmap of shot-sensitive behavior (right). Subtracted model biases are displayed above. Weak overfitting to shot (XOR patterning) is now visible in the heatmap.}
        \label{fig:calculations}
        \end{figure}

        \textbf{Decomposing accuracy and isolating shot-sensitive behavior}:
        The above training/evaluation scheme produces a table of raw accuracy scores organized by cross-referenced train and test shot values (Fig.~\ref{fig:calculations}, left). 
        Unfortunately, this table of raw scores is dominated by expected and uninteresting behavior. 
        We know that accuracy will increase with testing shot; this alone is not evidence of shot sensitivity. 
        Similarly, we do not expect all models to perform equally well in expectation: some hyperparameter combinations are simply more or less effective across the board. 
        A very low way, high shot training regime may saturate performance too quickly to provide a useful loss signal, while very high way, low shot training may be too difficult for proper convergence.
        This behavior is also not interesting, or evidence of shot sensitivity. 
        We therefore decompose the raw accuracy scores:
        \begin{equation}
        \label{eq:decomp}
            \begin{aligned}
                {\rm Accuracy} &= {\rm bias}_{\textit{test shot}} + {\rm bias}_{\textit{model}} + \textit{shot sensitive behavior}
            \end{aligned}
        \end{equation}
        
        By marginalizing out the first two terms, we can isolate and extract meaningful metrics and representations for shot-sensitivity.

        \begin{wrapfigure}{r}{0.59\linewidth}
        \vspace{-0.4cm}
        \centering
        \includegraphics[width=\linewidth]{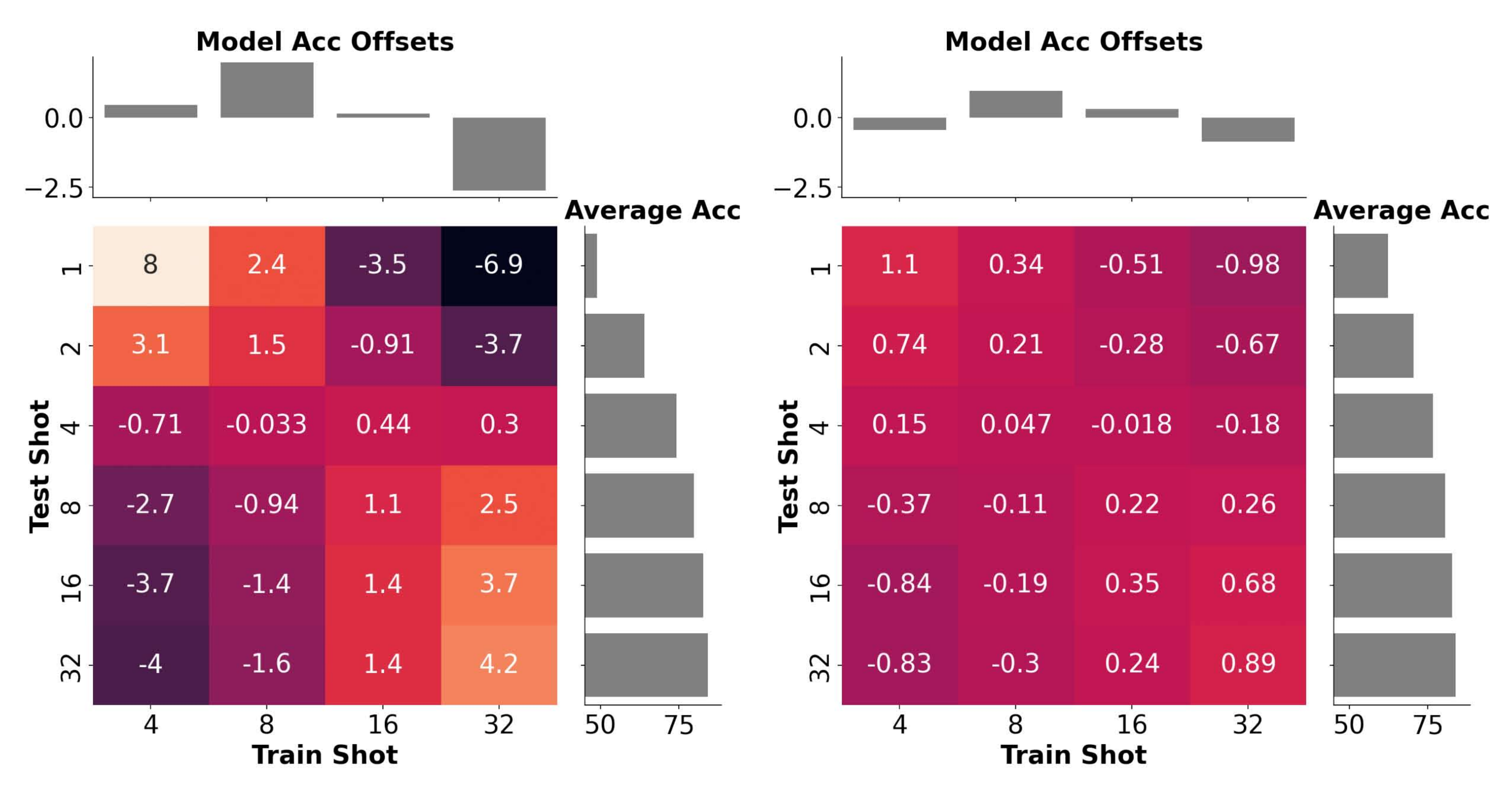}
        \caption{An illustrative example of overfitting to shot in a Euclidean (left) and cosine (right) Conv-4 ProtoNet (meta-iNat). Average group accuracy is plotted to the right of each heatmap; model performance relative to its group is plotted above. Both maps display typical XOR-style shot-overfitting, though the cosine model greatly reduces it (shot sensitivity of 14.86 vs 2.13).}
        \label{fig:heatmap_conv}
        \vspace{-0.4cm}
        \end{wrapfigure}
        
        We first marginalize out ${\rm bias}_{\textit{test shot}}$ by subtracting the average accuracy over the collection of networks for each test shot (Fig.~\ref{fig:calculations} step 1). 
        This produces a table of model offsets relative to the behavior of the group in expectation (Fig.~\ref{fig:calculations} middle). 
        For a learner entirely insensitive to test shot, we would expect each column of offsets to be a repeated constant: model bias is still present, but networks are equally insensitive to test shot changes. 
        We then marginalize out ${\rm bias}_{\textit{model}}$ to isolate shot-sensitive behavior specifically, by subtracting the average difference between each model's accuracy and the whole collection's accuracy (Fig.~\ref{fig:calculations} step 2). This produces a final heatmap of isolated shot-sensitive behavior (Fig.~\ref{fig:calculations}, right), with all behavior linked to train shot in isolation, or test shot in isolation, removed. 
        Shot overfitting manifests in these heatmaps as XOR-style patterns: increasing positive values on the diagonal indicate good performance when shots match, and decreasing negative values on the anti-diagonal indicate poor performance when shots differ (Fig.~\ref{fig:heatmap_conv} contains more examples). 
        The full set of heatmaps is displayed in appendix~\ref{sec:fullresults}.

        \textbf{Sensitivity score}:
        Shot-overfitting produces large positive diagonal values and negative off-diagonal values, while a perfectly shot-robust approach produces entirely zeros. 
        A straightforward sensitivity metric is then the difference between the largest and smallest offsets in the corrected heatmap. 
        This score provides a rough bound on how much better or worse a given approach is expected to behave as train and test shot vary: a score of 10 indicates that for the given problem setup, a model will always be within 10 points of its best- or worst-case performance. 
        
        Note that shot sensitivity is orthogonal to accuracy. This is by design: a few-shot classifier may gain 2 points of accuracy, but if sensitivity also goes up by 2, shot mismatch may wipe those gains entirely. Ideally we would like both high accuracy and low sensitivity; we now investigate how to attain this.

    \subsection{Reducing Shot Sensitivity with Cosine}
    
        \begin{wraptable}{r}{0.6\linewidth}
        \vspace{-0.8cm}
        \caption{Shot sensitivity scores for all models and datasets. Cosine alternatives reliably provide superior robustness to test-time shot.}
        \vspace{.1cm}
          \centering
          \footnotesize
          \begin{tabular}{l | c c c | c c c}
            \toprule
            \-\ & \multicolumn{3}{c}{\textbf{Conv-4}} & \multicolumn{3}{c}{\textbf{ResNet-12}}\\
            
            \textbf{Model} & CUB & mIN & iNat & CUB & mIN & tIN\\
            \midrule
            Proto~\cite{snell2017prototypical} &9.33 &7.32 &14.86 &1.14 &1.21 &1.01\\
            PCA~\cite{est} &4.46 &3.76 &2.26 &0.89 &1.43 &0.80\\
            EST~\cite{est} &4.07 &3.11  &3.95 &0.84 &1.24 &0.98\\
            Cosine Proto &\textbf{1.60} &\textbf{1.39} &\textbf{2.13} &\textbf{0.78} &\textbf{1.09} &\textbf{0.46}\\
            \hline
            FEAT~\cite{ye2020fewshot} &2.70 &3.77 &4.16 &6.10 &2.13 &0.63 \Tstrut\\
            Cosine FEAT &\textbf{0.72} &\textbf{1.31} &\textbf{2.43} &\textbf{1.27} &\textbf{0.19} &0.63\\
            \hline
            FRN~\cite{Wertheimer_2021_CVPR} &6.71 &5.09 &5.46 &2.80 &1.35 &1.07 \Tstrut\\
            Cosine FRN &\textbf{1.29} &\textbf{1.14} &\textbf{2.12} &\textbf{0.21} &\textbf{0.30} &\textbf{0.39}\\
            \bottomrule
          \end{tabular}
          \label{tab:sensitivities}
          \vspace{-0.4cm}
        \end{wraptable}
        
        Sensitivity scores are compared across settings in Table~\ref{tab:sensitivities}. 
        We make four observations. 
        First, ProtoNet is sensitive to shot, especially when using Conv-4. 
        For these models, scores fall within 7 to 14 points of best- or worst-case behavior, \textit{after} model bias correction. 
        This is not a promising guarantee, especially on such well-studied benchmarks where state-of-the-art advancements may be two points or smaller. 
        This sensitivity stems from shot overfitting;  Fig.~\ref{fig:heatmap_conv} provides an example.
        
        Second, the overfitting problem is not unique to ProtoNet. 
        FEAT and FRN display similar shot sensitivity (controlling for network architecture), though Conv-4 ProtoNet is the worst offender. 
        In line with \cite{goldblum2020unraveling}, shot sensitive behavior is broadly inherent to metric few-shot learners. 
        
        Third, shot-overfitting is generally more dramatic for Conv-4 models than for ResNet-12. 
        Overall patterns of behavior are the same,
        but the larger ResNet-12 feature extractor, with over 100x the parameters of Conv-4, possesses a similarly large degree of built-in insensitivity to test shot. 
        This makes any improvement on ResNet-12 models particularly noteworthy. 
        
        Fourth, and most crucially, cosine models consistently and reliably reduce shot sensitivity (as shown in Fig.~\ref{fig:heatmap_conv}) 
        compared to Euclidean counterparts, with only a \textit{single} exception in Table~\ref{tab:sensitivities}, where sensitivity is unchanged. 
        This is regardless of benchmark, architecture, or few-shot learner. 
        In fact, for Conv-4 architectures, the worst cosine model is still more robust to shot than the best Euclidean model (2.43 vs 2.70). 
        On average, cosine models reduce the spread of accuracy scores due to shot by a \textbf{factor of 4.5} compared to their Euclidean counterparts.
        
    \subsection{Comparison to Prior State of the Art}
        
        In addition to straightforward Cosine/Euclidean comparisons, we also evaluate two state-of-the-art shot robustness baselines: EST, a projection layer for maximal clustering, and a baseline PCA projection to the same number of dimensions \cite{est}. 
        Interestingly, EST does not always improve upon the shot sensitivity of PCA. This does not \textit{directly} contradict the findings of \cite{est}: while EST and PCA do have similar shot sensitivity scores, EST has generally higher accuracy (for detailed analysis of PCA/EST/cosine accuracy, see appendix~\ref{sec:estpcacos}), which \cite{est} does not decorrelate from shot robustness. 
        Regardless, while both techniques reduce ProtoNet shot sensitivity, cosine is still superior.

        Finally, we note that this comparison is only relevant for ProtoNet, the overall worst performer of our few-shot learners. 
        For FEAT and FRN, cosine models win by default, as the EST down-projection layer is based on class prototypes, and it is not clear what the corresponding layer would be for more sophisticated techniques that a) already exist in a smaller feature space, as feature maps are pooled not flattened, and b) do not use class prototypes directly for classification.

\section{Analysis and Discussion}
        
\begin{figure}
    \centering
    \includegraphics[width=\linewidth]{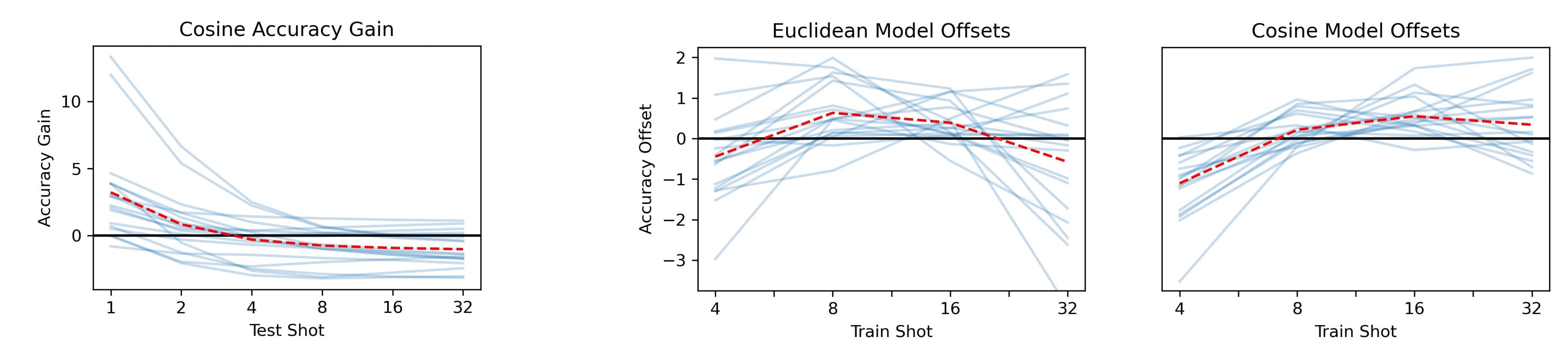}
    \caption{Model group behavior by shot. 
    Each line represents a model group from one dataset, architecture, and choice of few-shot learner.
    Left: cosine accuracy gains over Euclidean alternatives by test shot. Right: accuracy offsets from group performance due to training shot. 
    }
    \label{fig:offsets}
    \end{figure}

    \textbf{Impact on accuracy}:
        Accuracy and shot-robustness are orthogonal, so we might expect a model that provides consistent shot robustness to trade off accuracy to achieve this. 
        Interestingly, this is not the case: 
        compared to Euclidean models, cosine instead trades off high-shot performance for low-shot performance, usually at a beneficial rate. 
        This is made clear in Fig.~\ref{fig:offsets}~(left), where accuracy gains as a function of test shot are plotted for cosine models across all settings.
        1-shot performance improves for cosine by 3.2 points on average, while sacrificed high-shot performance saturates to 1 point on average (for more detailed comparisons, see appendix~\ref{sec:acctables}). 
        This behavior likely stems from the stabilized decision boundaries discussed in Sec.~\ref{sec:proto}.  
        Anchoring decision boundaries to the origin 
        is useful in low-shot regimes with high sample noise. 
        When support images are plentiful, it becomes less a useful prior and more an arbitrary constraint.

    One issue with accuracy gains as reported in  Fig.~\ref{fig:offsets}~(left) is that they are aggregate scores from collections of networks. 
    As such, these averages are not realized by any single model. 
    To better examine individual model behavior within collections, we plot model accuracy offsets (model biases displayed above the heatmaps in Fig.~\ref{fig:heatmap_conv}) in Fig.~\ref{fig:offsets}~(right). 
    Average trendlines represent the general impact of training shot on Euclidean and Cosine models.  
    Hyperparameter sensitivity is similar, but where Euclidean models do not consistently prefer high or low shot training, cosine models gather closer around their trend line.
    The more distinct behavior includes a preference for higher-shot training, as no 4-shot cosine model exceeds the mean accuracy of its group (offsets are negative or zero).
    This could explain why our 1-shot gains go unobserved in prior work: in 1- or 5-shot settings, gains from cosine are cancelled out by an aversion to low-shot training. 
    Removing these underperforming 4-shot models boosts cosine performance by $1.47$ points on average.

    \begin{figure}
    \centering
    \includegraphics[width=\linewidth]{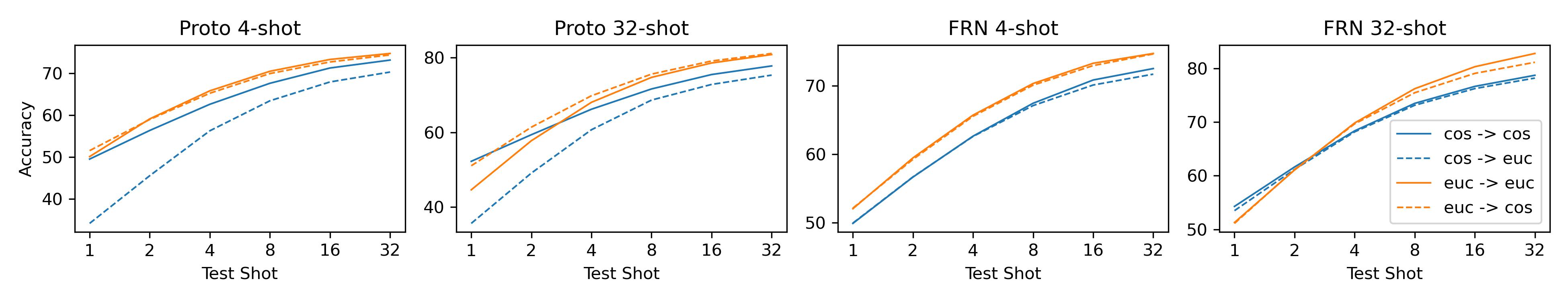}
    \caption{Euclidean models using cosine prediction, and cosine models using Euclidean prediction, tend to mimic their training counterparts rather than their testing counterparts.}
    \label{fig:swaps}
    \end{figure}
    
    \textbf{Is training with cosine distance necessary?}
    Our findings in Sec.~\ref{sec:proto} and Table~\ref{tab:consistency} suggest that cosine distance stabilizes decision boundaries regardless of training mechanism, and so could possibly robustify Euclidean models at test time without any specialized training. 
    We therefore re-evaluate eight of our Conv-4 Mini-ImageNet models with predictors swapped at test time: Euclidean models evaluate using cosine, and vice versa. 
    Fig.~\ref{fig:swaps} compares the results to the same models evaluated with the correct prediction types. 
    Swapped models mostly track the accuracy of the model with the corresponding training predictor, not the test predictor. 
    This suggests that training with cosine distance is the crucial aspect, and shot overfitting is a learned response to sample noise at train time, as predicted in prior work~\cite{est, goldblum2020unraveling}.

    \textbf{Limitations:}
    In our study we expand the scope of evaluation for few-shot classification to more realistic settings. This includes the obvious scenario where train and test shot do not match, but also a higher range of test shot (1-32) compared to most prior work, which typically caps at 5 (even \cite{est} uses only 1/5/10). 
    However, there are orthogonal directions that our benchmark (as well as most others for few-shot classification) does not capture, such as class imbalance. 
    Because we study the impact of shot mismatch, we must assume for practicality that train and test shot always exist, and are set to constant, comparable values. In the real world, however, class distributions are frequently long-tailed \cite{wertheimer2019few, gupta2019lvis}. 
    The degree to which shot-robust cosine classifiers can handle such problems is an interesting direction for future work. 
    
    Second, while cosine models confer impressive robustness to test shot, they are not reliably any more or less robust to the training shot hyperparameter, as shown in Fig.~\ref{fig:offsets}~(right). 
    In the cases where our cosine models trades off training shot sensitivity for test shot robustness,
    we view it as a worthwhile trade,  as the practitioner has control over the training shot hyperparameter but no control over shot at test time. 
    Therefore overcoming any increased sensitivity is a simple matter of careful validation. 
    
    While our analysis primarily concerns shot mismatch, which does not feature heavily in existing few-shot literature, we do not view this as a limitation of scope. Rather, shot mismatch is an inherent feature of real-world problems, and as such, we view our work as broadening the practical scope of deployment for few-shot classifiers. 
    
    \textbf{Societal impact:}
    By facilitating few-shot learning in more realistic settings, our work carries the same potential negative impacts as few-shot learning overall. 
    This includes inadvertently exacerbating problematic trends in data collection, such as overfitting to spurious factors and lack of diversity or minority representation, by allowing practitioners to ``settle'' for smaller training datasets. 
    However, these are issues inherent to few-shot and deep learning generally, and not specific to our work. 

\section{Conclusion}
    We conduct a systematic empirical study of test-time shot variation across datasets, neural architectures, and few-shot learning techniques. 
    A simple, general, and powerfully effective solution to shot overfitting is proposed: 
    replacing Euclidean distance logits with cosine distance reliably stabilizes test time performance without sacrificing accuracy, even on ResNet-12 where starting methods are already quite robust. 
    These cosine models make beneficial tradeoffs between high- and low-shot performance, achieving up to double-digit gains in 1-shot accuracy while sacrificing less than 1 point of high-shot accuracy on average. 
    By addressing a critical weakness in metric few-shot learning, we hope that these cosine methods will 
    serve as general and valuable tools for more realistic settings.

\paragraph{Acknowledgements:} This work was funded by NSF IIS-2144117 and the DARPA Learning with Less Labels program (HR001118S0044).
    
%%%%%%%%% REFERENCES
{\small
\bibliographystyle{abbrv}
\bibliography{egbib}
}

\newpage
\appendix

\section{Supplementary Materials}

\subsection{Datasets}
A summary of dataset statistics is given in Table~\ref{tab:datasets}, with details on each dataset below. 

\textbf{meta-iNat}~\cite{wertheimer2019few,van2018inaturalist} is a benchmark of animal species in the wild. This benchmark is particularly difficult, as classes are unbalanced, distinctions are fine-grained, and images are not cropped or centered, and may contain multiple animal instances. We follow the class split proposed by~\cite{wertheimer2019few}: of 1135 classes with between 50 and 1000 images, one fifth (227) are assigned to evaluation and the rest to training.
While the original meta-iNat involves all-way, imbalanced shot classification at test time \cite{wertheimer2019few}, we instead employ an episodic evaluation scheme as in \cite{Wertheimer_2021_CVPR}. 

\textbf{CUB}~\cite{WahCUB_200_2011} consists of 11,788 images from 200 bird classes. Following~\cite{chen2019closerfewshot, ye2020fewshot, tang2020revisiting, Wertheimer_2021_CVPR}, we randomly split categories into 100 classes for training, 50 for validation and 50 for evaluation. Our split is identical to ~\cite{tang2020revisiting, Wertheimer_2021_CVPR}. During training, we use raw images as input and do not crop to the provided bounding boxes.

\textbf{mini-ImageNet}~\cite{vinyals2016matching} is a subset of ImageNet containing 100 classes in total, with 600 examples per class. Following~\cite{ravi2016optimization}, we split categories into 64 classes for training, 16 for validation and 20 for test.

\textbf{tiered-ImageNet}~\cite{ren18fewshotssl} is a larger subset of ImageNet with 351-97-160 categories for training-validation-testing, respectively. 
Tiered-ImageNet ensures larger domain differences between training and evaluation compared to mini-ImageNet. Most works~\cite{wang2019simpleshot,tian2020rethinking,chen2020new,Wertheimer_2021_CVPR} use images from~\cite{ren18fewshotssl}\footnote{{\scriptsize\url{https://github.com/renmengye/few-shot-ssl-public}}} or~\cite{lee2019meta}\footnote{{\scriptsize\url{https://github.com/kjunelee/MetaOptNet}}}, which have 84$\times$84 resolution, and we do the same.

\begin{table}[h]
    \caption{Information on the four datasets used.}
          \centering
          \footnotesize
          \begin{tabular}{l | c | c | c}
            \toprule
            Dataset & \#Images & \#Classes & Train/Val/Test\\
            \midrule
            CUB~\cite{WahCUB_200_2011, tang2020revisiting} & 11,788 & 200 & 100/50/50\\
            meta-iNat~\cite{van2018inaturalist, wertheimer2019few} & 243,986 & 1135 & 908/0/227\\
            mini-ImageNet~\cite{vinyals2016matching} (mIN) & 60,000 & 100 & 64/16/20\\
            tiered-ImageNet~\cite{ren18fewshotssl} (tIN) & 779,165 & 608 & 351/97/160\\
            \bottomrule
          \end{tabular}
          \label{tab:datasets}
        \end{table}

\subsection{Training Details}
\label{sec:appdetails}
Our training setup and hyperparameters for the most part match \cite{Wertheimer_2021_CVPR} (available under MIT License), on which we base our code.
Hyperparameters particular to each method are kept the same as in the original paper, so for example we keep the same attention dropout rates for FEAT as in \cite{ye2020fewshot}, and reduce to 60 dimensions for all PCA/EST models as in \cite{est}. 
For practical reasons, we reduce the way (and therefore batch size) by $25\%$ for all ResNet-12 ProtoNet models trained on CUB, including ProtoNet, Cosine ProtoNet, PCA and EST, as well as all ResNet-12 FEAT models. 
We use the pretrained feature extractor from~\cite{Wertheimer_2021_CVPR} for mini-ImageNet and tiered-ImageNet ResNet-12 FRNs (standard and cosine). 
Following \cite{ye2020fewshot}, we pre-train our ResNet-12 FEAT models linearly before episodic fine-tuning. For mini-ImageNet and tiered-ImageNet we use the pretrained models provided by \cite{ye2020fewshot}; for CUB we pre-train our own feature extractor using the FRN pre-training hyperparameters.    
All other models are trained from scratch. 

For Conv-4 FEAT models that do not train with Adam (CUB and mini-ImageNet), we found that the standard hyperparameters from~\cite{Wertheimer_2021_CVPR} produced divergent behavior during training.
We reduced the initial learning rate for these FEAT models by a factor of 10 to recover stable behavior.

Likewise, for ResNet-12 FEAT models, we found that the default hyperparameters from~\cite{Wertheimer_2021_CVPR} do not work well for fine-tuning. We instead adopt the hyperparameters from \cite{ye2020fewshot}, but found that this caused convergence to be very slow. We therefore increased the learning rate: train with SGD and momentum 0.9 for 200 epochs (15 for the much larger tiered-ImageNet), with initial learning rate 1e-3 for the feature extractor and 1e-2 for the attention head, dividing learning rate by 10 halfway through. 

All models were trained on-premise on a range of NVIDIA GPU types. Each run used a single GPU and finished within hours to days, depending on dataset size, backbone architecture, and training procedure. 
We will release all the detailed training hyperparameters along with code upon acceptance, such that the audience will be able to reproduce our results.

\subsection{Cosine FRN Derivation}
\label{sec:cosfrn}
    \subsubsection{Motivating the new FRN regularizer term}
        The original FRN formulates the reconstruction problem as linear regression: given a matrix of class $c$ support features $S_c$, and a matrix of query features $Q$, find $\bar{W}$ such that $\bar{W}S_c \approx Q$. 
        This is formalized as: 
        \begin{equation}
        \begin{aligned}
        \label{eq:oldrecon}
            \bar W = \underset{W}{\text{arg min\ \ }} ||Q-WS_c||^2_2 + \frac{\lambda n}{d} ||W||^2_2
        \end{aligned}
        \end{equation}
        where $n$ is the number of support features, $d$ is the latent dimensionality, and $\lambda$ is a learned scaling term. 
        \cite{Wertheimer_2021_CVPR} justifies this choice of regularizer parameterization on the grounds that when $n/d$ is high, support features begin to span the feature space and the reconstruction becomes easy, so higher regularization is required. 
        The intuition is sensible, and the formulation comes with the additional benefit of being invariant to shot: expanding the support pool by repeating the existing entries has no effect on the final reconstruction. 
        
        Unfortunately, this parameterization leads to other, unintuitive behavior - for example, reconstruction quality is not equivariant to, or even monotonic with, feature scale or dimensionality. 
        Multiplying both the support and query features by a constant factor $c$ should yield the original reconstruction multiplied by $c$. 
        Similarly, expanding the dimensionality by concatenating all support and query features to themselves should yield the original reconstruction concatenated to itself. 
        When these behaviors are absent, shifting to different feature scales or network widths may require manual hyperparameter adjustment to compensate for the unpredictable shift in reconstruction behavior. 
        Indeed, the authors of \cite{Wertheimer_2021_CVPR} found it necessary to manually normalize latent values when implementing FRN on wider networks to prevent divergence during training.
        Instead, we reformulate the reconstruction problem as:
        \begin{equation}
        \begin{aligned}
        \label{eq:newrecon}
            \bar W = \underset{W}{\text{arg min\ \ }} ||Q-WS_c||_2^2 + (\lambda||S_c^TS_c||_2) ||W||_2^2
        \end{aligned}
        \end{equation}
        where $||\cdot||_2$ is the Frobenius norm.
        Eq.~\ref{eq:newrecon} provides all of the above desirable invariances and equivariances. 
        Demonstrating this is relatively straightforward. 
        
        \textbf{Invariance to shot}: expanding the support pool by repeating the individual features should not change the reconstruction.  
        Given the original $n\times d$ support feature matrix $S_c$, let $S_c'$ be the $2n\times d$ feature matrix where every feature in $S_c$ occurs twice. It must then be the case that $S_c'^TS_c' = 2S_c^TS_c$. Therefore the closed-form reconstruction from $S_c'$ is given via the Normal Equation and Woodbury Identity by:
        \begin{equation}
        \begin{aligned}
            \bar{W}S_c' &= Q(S_c'^TS_c' + \lambda||S_c'^TS_c'||_2)^{-1}S_c'TS_c'\\
            &= Q(2S_c^TS_c + 2\lambda ||S_c^TS_c||_2)^{-1} * 2S_c^TS_c\\
            &= Q(S_c^TS_c+\lambda||S_c^TS_c||_2)^{-1}S_c^TS_c
        \end{aligned}
        \end{equation}
        which is the original reconstruction. We refer the reader to \cite{Wertheimer_2021_CVPR,bertinetto2018metalearning} for background on the Woodbury normal formulation. 
        
        \textbf{Equivariance to scale}: scaling all features by $\alpha$ should yield the original reconstruction times $\alpha$. This follows an analogous argument to shot invariance. Let $S_c'=\alpha S_c'$ and $Q'=\alpha Q$. It must then be the case that $S_c'^TS_c' = \alpha^2S_c^TS_c$. Therefore the closed-form reconstruction from $S_c'$ is given by:
        \begin{equation}
        \begin{aligned}
            \bar{W}S_c' &= Q'(S_c'^TS_c' + \lambda||S_c'^TS_c'||_2)^{-1}S_c'TS_c'\\
            &= \alpha Q(\alpha^2S_c^TS_c + \alpha^2\lambda ||S_c^TS_c||_2)^{-1} \alpha^2S_c^TS_c\\
            &= \alpha Q(S_c^TS_c+\lambda||S_c^TS_c||_2)^{-1}S_c^TS_c
        \end{aligned}
        \end{equation}
        which is the original reconstruction times $\alpha$. 
        
        \textbf{Equivariance to dimensionality}: expanding the dimensionality by repeating all existing features should yield the original reconstruction, also with repeated features. Let $S_c'=S_c \oplus S_c$ and $Q'=Q\oplus Q$ where $\oplus$ represents concatenation. It must then be the case that $S_c'S_c'^T = 2S_cS_c^T$ and $Q'S_c'^T = 2QS_c^T$. Therefore the closed-form reconstruction from $S_c'$ is given by (discarding the Woodbury Identity in this case):
        \begin{equation}
        \begin{aligned}
            \bar{W}S_c' &= Q'S_c'^T(S_c'S_c'^T + \lambda||S_c'^TS_c'||_2)^{-1}S_c'\\
            &= 2QS_c^T(2S_cS_c^T + 2\lambda ||S_c^TS_c||_2)^{-1} (S_c\oplus S_c)\\
            &= QS_c^T(S_cS_c^T+\lambda||S_c^TS_c||_2)^{-1}(S_c\oplus S_c)
        \end{aligned}
        \end{equation}
        Distributing over the concatenation in the last line yields the original reconstruction, concatenated with itself. The second line uses the fact that $||X^TX||_2 = ||XX^T||_2$ for all $X$. 
        
        Equipped with this more intuitive and stable behavior, our updated FRN regularizer works out of the box in all settings, without requiring any manual hyperparameter tuning based on neural architecture. 
        We also found it to be more robust to pretraining scheme: unlike the original, the updated regularizer is amenable to vanilla linear pretraining (though we do not report these results in the main paper). 
    
    \subsubsection{Deriving the expanded FRN prediction formula}
    We use the same notation as above. Let $\Sigma_S = S_c^TS_c$ and $\Sigma_Q = Q^TQ$, and let the regularizer matrix $\Lambda = (\lambda||S_c^TS_c||_2)I$. The prediction logit for a given class is then:
    \begin{equation}
        \begin{aligned}
            \bar{W} &= Q(\Sigma_S+\Lambda)^{-1}S_c^T\\ 
            z(Q;S_c) &= -||\bar{W}S_c-Q||^2\\
            &= -tr((\bar{W}S_c-Q)^T(\bar{W}S_c-Q))\\
            &= 2tr(Q^T\bar{W}S_c) - tr(Q^TQ) - tr(S_c^T\bar{W}^T\bar{W}S_c)\\
            &= 2tr(\Sigma_Q(\Sigma_S+\Lambda)^{-1}\Sigma_S) - tr(\Sigma_Q) - tr((\Sigma_S+\Lambda)^{-1}\Sigma_S\Sigma_Q\Sigma_S(\Sigma_S+\Lambda)^{-1})\\
            &= 2||(\Sigma_S+\Lambda)^{-1}\Sigma_S \otimes \Sigma_Q||_1 - tr(\Sigma_Q) -tr((\Sigma_S+\Lambda)^{-1}\Sigma_S\Sigma_Q\Sigma_S(\Sigma_S+\Lambda)^{-1})\\
            &\approx ||(\Sigma_S+\Lambda)^{-1}\Sigma_S \otimes \Sigma_Q||_1
        \end{aligned}
        \end{equation}
        where $\otimes$ represents elementwise multiplication, which matches the result in Eq.~\ref{eq:frnfinal} of the main paper. The second-to-last line uses the fact that $tr(AB)=\langle A,B\rangle_F$, the Frobenius inner product. Motivation for removing the extra terms in the final line is discussed in Sec.~\ref{sec:frn}. The FRN models in our experiments use the updated regularizer, and train using the last line of the above formula. 
    
    \subsubsection{Comparing the new and old FRN models}
    \label{sec:frncompare}
        \begin{table}
      \caption{Accuracy comparison for the old (left) and new (right) FRN formulations on meta-iNat. Average group accuracy is displayed in the rightmost column of each table, model offsets are given in the bottom row. Bottom right cell (intersection of Offset and Mean) shows the sensitivity score.}
      \centering
      \footnotesize
      \begin{tabular}{c c | c c c c | c}
        \toprule
        \multicolumn{2}{c|}{\textbf{meta-iNat Conv-4}} & \multicolumn{5}{c}{\textbf{Train Shot}} \\
        \multicolumn{2}{c|}{\textbf{FRN (old)}} & 4 & 8 & 16 & 32 & Mean \\
        \hline
        \Tstrut\multirow{7}{*}{\makecell{\textbf{Test}\\\textbf{Shot}}} 
        & 1 & 64.25 & 63.55 & 60.58 & 57.67 & 61.51 \\
        & 2 & 73.17 & 73.32 & 71.34 & 68.64 & 71.62 \\
        & 4 & 78.73 & 79.81 & 79.10 & 77.36 & 78.75 \\
        & 8 & 81.87 & 83.82 & 84.01 & 83.13 & 83.21 \\
        & 16 & 83.81 & 86.16 & 86.79 & 86.35 & 85.78 \\
        & 32 & 84.60 & 87.12 & 88.33 & 88.33 & 87.10 \\
        \cline{2-7}
        \Tstrut & Offset & -0.26 & 0.97 & 0.36 & -1.08 & 2.60 \\
        \bottomrule
      \end{tabular}
      \quad
      \begin{tabular}{c c | c c c c | c}
        \toprule
        \multicolumn{2}{c|}{\textbf{meta-iNat Conv-4}} & \multicolumn{5}{c}{\textbf{Train Shot}} \\
        \multicolumn{2}{c|}{\textbf{FRN (new)}} & 4 & 8 & 16 & 32 & Mean \\
        \hline
        \Tstrut\multirow{7}{*}{\makecell{\textbf{Test}\\\textbf{Shot}}} 
        & 1 & 65.30 & 63.77 & 61.03 & 58.56 & 62.17 \\
        & 2 & 74.01 & 73.52 & 71.78 & 69.58 & 72.22 \\
        & 4 & 79.37 & 80.15 & 79.42 & 77.95 & 79.22 \\
        & 8 & 82.78 & 84.21 & 83.98 & 83.10 & 83.52 \\
        & 16 & 84.36 & 86.05 & 86.55 & 86.28 & 85.81 \\
        & 32 & 85.37 & 87.29 & 87.97 & 88.07 & 87.18 \\
        \cline{2-7}
        \Tstrut & Offset & 0.18 & 0.81 & 0.10 & -1.10 & 2.30 \\
        \bottomrule
      \end{tabular}
      \label{tab:supp_frn_acc}
    \end{table}
    Table~\ref{tab:supp_frn_acc} displays example accuracy scores from two Conv-4 FRN models trained on meta-iNat with the old (left) and new (right) prediction and regularizer formulations. Behavior is very similar, with the updated and simplified FRN formula actually showing slight improvements overall.

\subsection{Full Results}
\subsubsection{Detailed EST/PCA/Cosine Comparison}
\label{sec:estpcacos}
\begin{figure}
        \vspace{-0.4cm}
        \centering
        \includegraphics[width=.5\linewidth]{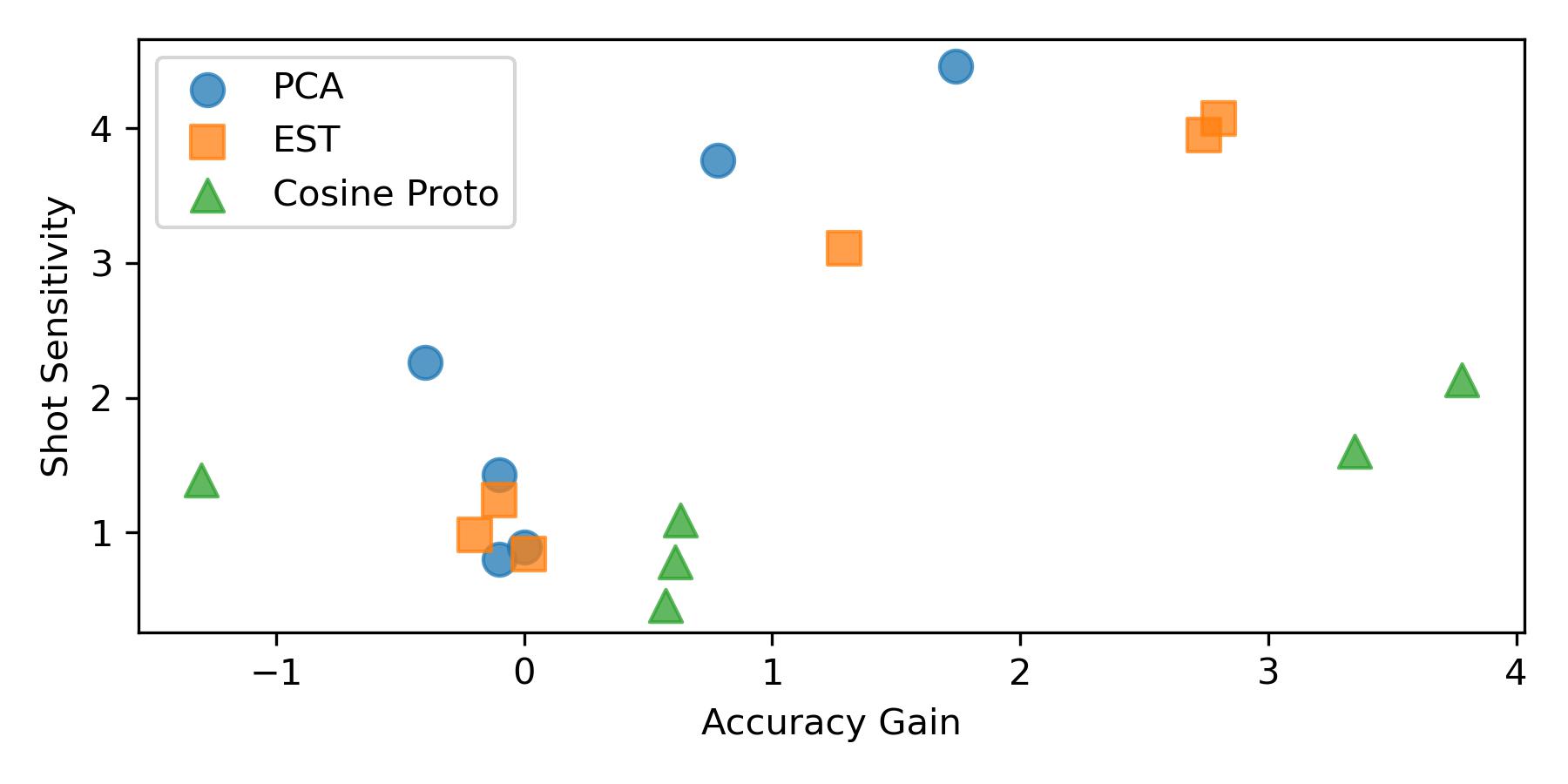}
        \caption{Accuracy/sensitivity tradeoffs for shot-robust methods based on ProtoNet. Each dot corresponds to a single benchmark and neural architecture. Values lower and to the right are desirable (low sensitivity, high accuracy).}
        \label{fig:scatter}
        \vspace{-0.4cm}
        \end{figure}

        For the ProtoNet setting, we compare our cosine classifiers to both a Euclidean ProtoNet and state-of-the-art shot robustness baselines: EST, a projection layer for maximal clustering, and a baseline PCA projection to the same number of dimensions \cite{est}. 
        Detailed comparisons of accuracy and shot-robustness between these methods are given in rows 2-4 and 10-12 of Tables~\ref{tab:ss_gain}, \ref{tab:ss_gain_cub} and \ref{tab:ss_gain_mini}. 
        Cosine models not only provide better robustness than EST and PCA, they also generally produce higher accuracy scores across the range of test shots. 
        This is demonstrated in Fig.~\ref{fig:scatter}, where all EST, PCA, and cosine accuracy gains, averaged over test shots, are plotted against shot sensitivity. 
        We wish to attain low sensitivity and high accuracy (bottom right corner), and cosine models form a clear frontier in this direction. 
        
\subsubsection{Additional Accuracy Gain Tables}
\label{sec:acctables}
Sensitivity scores and average group accuracy gains for all models on CUB, mini-ImageNet, and meta-iNat/tiered-ImageNet are displayed in Tables~\ref{tab:ss_gain_cub}, \ref{tab:ss_gain_mini} and \ref{tab:ss_gain} respectively.
\begin{table*}
          \caption{Sensitivity scores and average group accuracy gains for all models on CUB, illustrating sensitivity and accuracy tradeoffs.}
          \centering
          \tiny
          \begin{tabularx}{\linewidth}{X | l | Y | l l l l l l}
            \toprule
            \-\ & \-\ & \-\ & \multicolumn{6}{c}{\textbf{test shot}}\\
            \textbf{Network} & \textbf{Model} & \textbf{SS} ($\downarrow$) & 1 & 2 & 4 & 8 & 16 & 32\\
            \midrule
            \multirow{8}{*}{Conv-4}&Proto~\cite{snell2017prototypical} &9.33 & 50.86 & 65.40 & 74.54 & 80.02 & 82.81 & 84.22 \\
            &PCA~\cite{est} &4.46 & 57.32 (\textcolor{blue}{+6.46}) & 68.48 (\textcolor{blue}{+3.09}) & 75.92 (\textcolor{blue}{+1.38}) & 80.21 (\textcolor{blue}{+0.19}) & 82.58 (\textcolor{red}{-0.24}) & 83.77 (\textcolor{red}{-0.44}) \\
            &EST~\cite{est} &4.07 & 59.94 (\textcolor{blue}{+9.08}) & 70.23 (\textcolor{blue}{+4.84}) & 76.99 (\textcolor{blue}{+2.44}) & 80.78 (\textcolor{blue}{+0.76}) & 82.85 (\textcolor{blue}{+0.03}) & 83.37 (\textcolor{red}{-0.35})\\
            &Cosine Proto &\textbf{1.60} & 62.85 (\textcolor{blue}{+11.99}) &70.79 (\textcolor{blue}{+5.39}) &76.78 (\textcolor{blue}{+2.23}) &80.62 (\textcolor{blue}{+0.59}) &82.86 (\textcolor{blue}{+0.05}) &84.09 (\textcolor{red}{-0.13})\\
            \cline{2-9}
            &FEAT~\cite{ye2020fewshot} &2.70 & 59.85 & 70.38 & 77.24 & 81.28 & 83.50 & 84.65\Tstrut\\
            &Cosine FEAT &\textbf{0.72} & 63.75 (\textcolor{blue}{+3.90}) & 71.71 (\textcolor{blue}{+1.34}) & 77.11 (\textcolor{red}{-0.13}) & 80.27 (\textcolor{red}{-1.02}) & 82.04 (\textcolor{red}{-1.46}) & 82.88 (\textcolor{red}{-1.77}) \\
            \cline{2-9}
            &FRN~\cite{Wertheimer_2021_CVPR} &6.71 & 67.74 & 77.15 & 83.29 & 86.62 & 88.49 & 89.49\Tstrut\\
            &Cosine FRN &\textbf{1.29} & 70.78 (\textcolor{blue}{+3.05}) & 78.11 (\textcolor{blue}{+0.97}) & 82.95 (\textcolor{red}{-0.34}) & 85.62 (\textcolor{red}{-1.00}) & 87.06 (\textcolor{red}{-1.43}) & 87.77 (\textcolor{red}{-1.73}) \\
            \midrule
            \multirow{8}{*}{ResNet-12}&Proto~\cite{snell2017prototypical} &1.14 & 74.44 & 82.70 & 87.15 & 89.36 & 90.45 & 91.01 \\
            &PCA~\cite{est} &0.89 & 74.53 (\textcolor{blue}{+0.09}) & 82.62 (\textcolor{red}{-0.07}) & 87.14 (\textcolor{red}{-0.01}) & 89.27 (\textcolor{red}{-0.10}) & 90.45 (\textcolor{red}{-0.00}) & 90.99 (\textcolor{red}{-0.01}) \\
            &EST~\cite{est} &0.84 & 74.59 (\textcolor{blue}{+0.16}) & 82.69 (\textcolor{red}{-0.01}) & 87.18 (\textcolor{blue}{+0.03}) & 89.34 (\textcolor{red}{-0.02}) & 90.45 (\textcolor{red}{-0.01}) & 90.96 (\textcolor{red}{-0.05})\\
            &Cosine Proto &\textbf{0.78} & 76.67 (\textcolor{blue}{+2.24}) &83.50 (\textcolor{blue}{+0.80}) &87.51 (\textcolor{blue}{+0.35}) &89.51 (\textcolor{blue}{+0.15}) &90.52 (\textcolor{blue}{+0.07}) &91.05 (\textcolor{blue}{+0.04})\\
            \cline{2-9}
            &FEAT~\cite{ye2020fewshot} &6.10 & 72.55 & 82.71 & 87.97 & 90.58 & 91.87 & 92.54\Tstrut\\
            &Cosine FEAT &\textbf{1.27} & 75.43 (\textcolor{blue}{+2.88}) & 84.43 (\textcolor{blue}{+1.72}) & 89.38 (\textcolor{blue}{+1.41}) & 91.85 (\textcolor{blue}{+1.27}) & 93.03 (\textcolor{blue}{+1.16}) & 93.64 (\textcolor{blue}{+1.10}) \\
            \cline{2-9}
            &FRN~\cite{Wertheimer_2021_CVPR} &2.80 & 80.78 & 87.47 & 90.86 & 92.72 & 93.67 & 94.26\Tstrut\\
            &Cosine FRN &\textbf{0.21} & 79.96 (\textcolor{red}{-0.82}) & 86.12 (\textcolor{red}{-1.36}) & 89.41 (\textcolor{red}{-1.45}) & 91.04 (\textcolor{red}{-1.69}) & 91.83 (\textcolor{red}{-1.84}) & 92.18 (\textcolor{red}{-2.08}) \\
            \bottomrule
          \end{tabularx}
          \label{tab:ss_gain_cub}
        \end{table*}
        
    \begin{table*}
          \caption{Sensitivity scores and average group accuracy gains for all models on mini-ImageNet, illustrating sensitivity and accuracy tradeoffs.}
          \centering
          \tiny
          \begin{tabularx}{\linewidth}{X | l | Y | l l l l l l}
            \toprule
            \-\ & \-\ & \-\ & \multicolumn{6}{c}{\textbf{test shot}}\\
            \textbf{Network} & \textbf{Model} & \textbf{SS} ($\downarrow$) & 1 & 2 & 4 & 8 & 16 & 32\\
            \midrule
            \multirow{8}{*}{Conv-4}&Proto~\cite{snell2017prototypical} &7.32 & 47.00 & 58.34 & 67.07 & 72.98 & 76.39 & 78.33 \\
            &PCA~\cite{est} &3.76 & 50.60 (\textcolor{blue}{+3.60}) & 59.86 (\textcolor{blue}{+1.52}) & 67.54 (\textcolor{blue}{+0.47}) & 72.93 (\textcolor{red}{-0.66}) & 76.02 (\textcolor{red}{-0.36}) & 77.85 (\textcolor{red}{-0.48}) \\
            &EST~\cite{est} &3.11 & 51.60 (\textcolor{blue}{+4.61}) & 60.89 (\textcolor{blue}{+2.56}) & 68.20 (\textcolor{blue}{+1.13}) & 73.17 (\textcolor{blue}{+0.19}) & 76.18 (\textcolor{red}{-0.21}) & 77.77 (\textcolor{red}{-0.56})\\
            &Cosine Proto &\textbf{1.39} & 50.85 (\textcolor{blue}{+3.86}) &57.80 (\textcolor{red}{-0.54}) &64.43 (\textcolor{red}{-2.64}) &69.86 (\textcolor{red}{-3.12}) &73.61 (\textcolor{red}{-2.78}) &75.88 (\textcolor{red}{-2.45})\\
            \cline{2-9}
            &FEAT~\cite{ye2020fewshot} & 3.77 & 50.37 & 59.58 & 67.02 & 72.25 & 75.32 & 77.11 \Tstrut\\
            &Cosine FEAT &\textbf{1.31} & 50.33 (\textcolor{red}{-0.03}) & 57.52 (\textcolor{red}{-2.07}) & 64.05 (\textcolor{red}{-2.98}) & 69.04 (\textcolor{red}{-3.21}) & 72.20 (\textcolor{red}{-3.12}) & 74.05 (\textcolor{red}{-3.06}) \\
            \cline{2-9}
            &FRN~\cite{Wertheimer_2021_CVPR} &5.09 & 52.11 &61.16 &68.75 &74.20 &77.57 &79.53\Tstrut\\
            &Cosine FRN &\textbf{1.14} & 52.80 (\textcolor{blue}{+0.69})& 59.87 (\textcolor{red}{-1.29}) & 66.25 (\textcolor{red}{-2.50}) & 71.31 (\textcolor{red}{-2.88}) & 74.50 (\textcolor{red}{-3.07}) & 76.37 (\textcolor{red}{-3.16}) \\
            \midrule
            \multirow{8}{*}{ResNet-12}&Proto~\cite{snell2017prototypical} &1.21 & 58.69 & 68.49 & 75.30 & 79.64 & 82.09 & 83.48 \\
            &PCA~\cite{est} &1.43 &58.75 (\textcolor{blue}{+0.06}) & 68.41 (\textcolor{red}{-0.08}) & 75.17 (\textcolor{red}{-0.14}) & 79.54 (\textcolor{red}{-0.10}) & 81.97 (\textcolor{red}{-0.12}) & 83.32 (\textcolor{red}{-0.16}) \\
            &EST~\cite{est} &1.24 & 58.89 (\textcolor{blue}{+0.20}) & 68.47 (\textcolor{red}{-0.02}) & 75.17 (\textcolor{red}{-0.14}) & 79.47 (\textcolor{red}{-0.17}) & 81.98 (\textcolor{red}{-0.11}) & 83.32 (\textcolor{red}{-0.16})\\
            &Cosine Proto &\textbf{1.09} & 61.65 (\textcolor{blue}{+2.96}) &69.14 (\textcolor{blue}{+0.65}) &75.22 (\textcolor{red}{-0.09}) &79.60 (\textcolor{red}{-0.04}) &82.20 (\textcolor{blue}{+0.11}) &83.67 (\textcolor{blue}{+0.19})\\
            \cline{2-9}
            &FEAT~\cite{ye2020fewshot} & 2.13 & 57.50 & 69.14 & 77.08 & 81.79 & 84.55 & 85.97 \Tstrut\\
            &Cosine FEAT &\textbf{0.19} & 59.39 (\textcolor{blue}{1.89}) & 69.62 (\textcolor{blue}{0.48}) & 77.45 (\textcolor{blue}{0.37}) & 82.37 (\textcolor{blue}{0.57}) & 85.29 (\textcolor{blue}{0.74}) & 86.87 (\textcolor{blue}{0.90}) \\
            \cline{2-9}
            &FRN~\cite{Wertheimer_2021_CVPR} &1.35 &65.82 &75.18 &82.30 &86.85 &89.62 &91.12\Tstrut\\
            &Cosine FRN &\textbf{0.30} & 66.73 (\textcolor{blue}{+0.91})& 75.29 (\textcolor{blue}{+0.11}) & 81.82 (\textcolor{red}{-0.49}) & 86.19 (\textcolor{red}{-0.66}) & 88.53 (\textcolor{red}{-1.09}) & 89.75 (\textcolor{red}{-1.37}) \\
            \bottomrule
          \end{tabularx}
          \label{tab:ss_gain_mini}
        \end{table*}
        
        \begin{table*}
        \caption{Sensitivity scores and average group accuracy gains for all models across two datasets, illustrating sensitivity and accuracy tradeoffs.}
          \centering
          \tiny
          \begin{tabularx}{\linewidth}{X | l | Y | l l l l l l}
            \toprule
            \-\ & \-\ & \-\ & \multicolumn{6}{c}{\textbf{Test Shot}}\\
            \textbf{Setting} & \textbf{Model} & \textbf{SS} ($\downarrow$) & 1 & 2 & 4 & 8 & 16 & 32\\
            \midrule
            \multirow{8}{*}{\shortstack{Conv-4\\iNat}}&Proto~\cite{snell2017prototypical} &14.86 & 49.04 & 63.95 & 74.17 & 79.86 & 82.76 & 84.30 \\
            &PCA~\cite{est} &2.26 & 58.16 (\textcolor{blue}{+9.11}) & 66.66 (\textcolor{blue}{+2.71}) & 72.67 (\textcolor{red}{-1.50}) & 76.38 (\textcolor{red}{-3.48}) & 78.66 (\textcolor{red}{-4.11}) & 79.37 (\textcolor{red}{-4.93}) \\
            &EST~\cite{est} &3.95 & 59.78 (\textcolor{blue}{+10.74}) & 69.35 (\textcolor{blue}{+5.40}) & 76.00 (\textcolor{blue}{+1.83}) & 79.90 (\textcolor{blue}{+0.05}) & 82.36 (\textcolor{red}{-0.41}) & 83.11 (\textcolor{red}{-1.19})\\
            &Cosine Proto &\textbf{2.13} & 62.37 (\textcolor{blue}{+13.33}) &70.60 (\textcolor{blue}{+6.65}) &76.65 (\textcolor{blue}{+2.48}) &80.53 (\textcolor{blue}{+0.67}) &82.69 (\textcolor{red}{-0.07}) &83.92 (\textcolor{red}{-0.39})\\
            \cline{2-9}
            &FEAT~\cite{ye2020fewshot} &4.16 & 58.58 & 69.55 & 76.95 & 81.35 & 83.81 & 85.18\Tstrut\\
            &Cosine FEAT &\textbf{2.43} & 63.22 (\textcolor{blue}{+4.64}) & 71.87 (\textcolor{blue}{+2.32}) & 77.95 (\textcolor{blue}{+0.99}) & 81.60 (\textcolor{blue}{+0.24}) & 83.63 (\textcolor{red}{-0.18}) & 84.74 (\textcolor{red}{-0.45}) \\
            \cline{2-9}
            &FRN~\cite{Wertheimer_2021_CVPR} &5.46 & 62.17 & 72.22 & 79.22 & 83.52 & 85.81 & 87.18\Tstrut\\
            &Cosine FRN &\textbf{2.12} & 66.00 (\textcolor{blue}{+3.83}) & 73.90 (\textcolor{blue}{+1.68}) & 79.45 (\textcolor{blue}{+0.23}) & 82.73 (\textcolor{red}{-0.79}) & 84.52 (\textcolor{red}{-1.29}) & 85.48 (\textcolor{red}{-1.69}) \\
            \midrule
            \multirow{8}{*}{\shortstack{ResNet-12\\tIN}}&Proto~\cite{snell2017prototypical} &1.01 & 65.04 & 74.28 & 80.26 & 83.91 & 85.85 & 86.91 \\
            &PCA~\cite{est} &0.80 & 65.00 (\textcolor{red}{-0.04}) & 74.20 (\textcolor{red}{-0.09}) & 80.20 (\textcolor{red}{-0.07}) & 83.71 (\textcolor{red}{-0.21}) & 85.71 (\textcolor{red}{-0.14}) & 86.73 (\textcolor{red}{-0.19}) \\
            &EST~\cite{est} &0.98 & 65.06 (\textcolor{blue}{+0.03}) & 74.11 (\textcolor{red}{-0.17}) & 80.08 (\textcolor{red}{-0.18}) & 83.60 (\textcolor{red}{-0.27}) & 85.58 (\textcolor{red}{-0.14}) & 86.58 (\textcolor{red}{-0.33}) \\
            &Cosine Proto &\textbf{0.46} & 67.12 (\textcolor{blue}{+2.08}) & 74.63 (\textcolor{blue}{+0.35}) & 80.25 (\textcolor{red}{-0.01}) & 84.07 (\textcolor{blue}{+0.16}) & 86.20 (\textcolor{blue}{+0.35}) & 87.40 (\textcolor{blue}{+0.49}) \\
            \cline{2-9}
            &FEAT~\cite{ye2020fewshot} & 0.63 & 66.64 & 76.08 & 81.92 & 85.34 & 87.23 & 88.21 \Tstrut\\
            &Cosine FEAT & 0.63 & 66.63 (\textcolor{red}{-0.02}) & 74.11 (\textcolor{red}{-1,97}) & 79.61 (\textcolor{red}{-2.31}) & 83.33 (\textcolor{red}{-2.00}) & 85.45 (\textcolor{red}{-1.78}) & 86.63 (\textcolor{red}{-1.58}) \\
            \cline{2-9}
            &FRN~\cite{Wertheimer_2021_CVPR} &1.07 & 69.68 & 78.37 & 84.40 & 88.20 & 90.47 & 91.71\Tstrut\\
            &Cosine FRN &\textbf{0.39} & 70.17 (\textcolor{blue}{+0.49}) & 78.06 (\textcolor{red}{-0.32}) & 83.70 (\textcolor{red}{-0.70}) & 87.23 (\textcolor{red}{-0.97}) & 89.25 (\textcolor{red}{-1.22}) & 90.29 (\textcolor{red}{-1.42}) \\
            \bottomrule
          \end{tabularx}
          \label{tab:ss_gain}
        \end{table*}

\subsubsection{Full Accuracy and Heatmap Results}
\label{sec:fullresults}

Paired accuracy tables and heatmaps are displayed for all tested models below. 
Models are listed by dataset, then by model architecture, then by learning method. 
Precise figure and table locations are given in Table~\ref{tab:supp_toc}. 

\begin{table*}
  \caption{Hyperlinked figure numbers for the accuracy tables and sensitivity heatmaps for each tested model configuration.}
  \centering
  \footnotesize
  \begin{tabularx}{0.75\linewidth}{X | c c c | c c c}
    \toprule
    \-\ & \multicolumn{3}{c}{\textbf{Conv-4}} & \multicolumn{3}{c}{\textbf{ResNet-12}}\\
    
    \textbf{Model} & CUB & mIN & iNat & CUB & mIN & tIN\\
    \midrule
    Proto~\cite{snell2017prototypical}/Cosine & fig.~\ref{fig:supp_heatmap_cub_proto} & fig.~\ref{fig:supp_heatmap_min_proto} & fig.~\ref{fig:supp_heatmap_inat_proto} & fig.~\ref{fig:supp_heatmap_cub_proto_12} & fig.~\ref{fig:supp_heatmap_min_proto_12} & fig.~\ref{fig:supp_heatmap_tin_proto}\\
    PCA/EST~\cite{est} &fig.~\ref{fig:supp_heatmap_cub_est} & fig.~\ref{fig:supp_heatmap_min_est} & fig.~\ref{fig:supp_heatmap_inat_pca_est} & fig.~\ref{fig:supp_heatmap_cub_est_12} & fig.~\ref{fig:supp_heatmap_min_est_12} & fig.~\ref{fig:supp_heatmap_tin_pca_est}\\
    \hline
FEAT~\cite{ye2020fewshot}/Cosine & fig.~\ref{fig:supp_heatmap_cub_feat} & fig.\ref{fig:supp_heatmap_min_feat} & fig.~\ref{fig:supp_heatmap_inat_feat} & fig.~\ref{fig:supp_heatmap_cub_feat_12} & fig.~\ref{fig:supp_heatmap_min_feat_12} & fig.~\ref{fig:supp_heatmap_tin_feat} \Tstrut\\
    \hline
    FRN~\cite{Wertheimer_2021_CVPR}/Cosine & fig.~\ref{fig:supp_heatmap_cub_frn} & fig.~\ref{fig:supp_heatmap_min_frn} & fig.~\ref{fig:supp_heatmap_inat_frn} & fig.~\ref{fig:supp_heatmap_cub_frn_12} & fig.~\ref{fig:supp_heatmap_min_frn_12} & fig.~\ref{fig:supp_heatmap_tin_frn} \Tstrut\\

    \bottomrule
  \end{tabularx}
  \label{tab:supp_toc}
\end{table*}

\clearpage

        \begin{figure*}
        \centering
        \includegraphics[width=.8\linewidth]{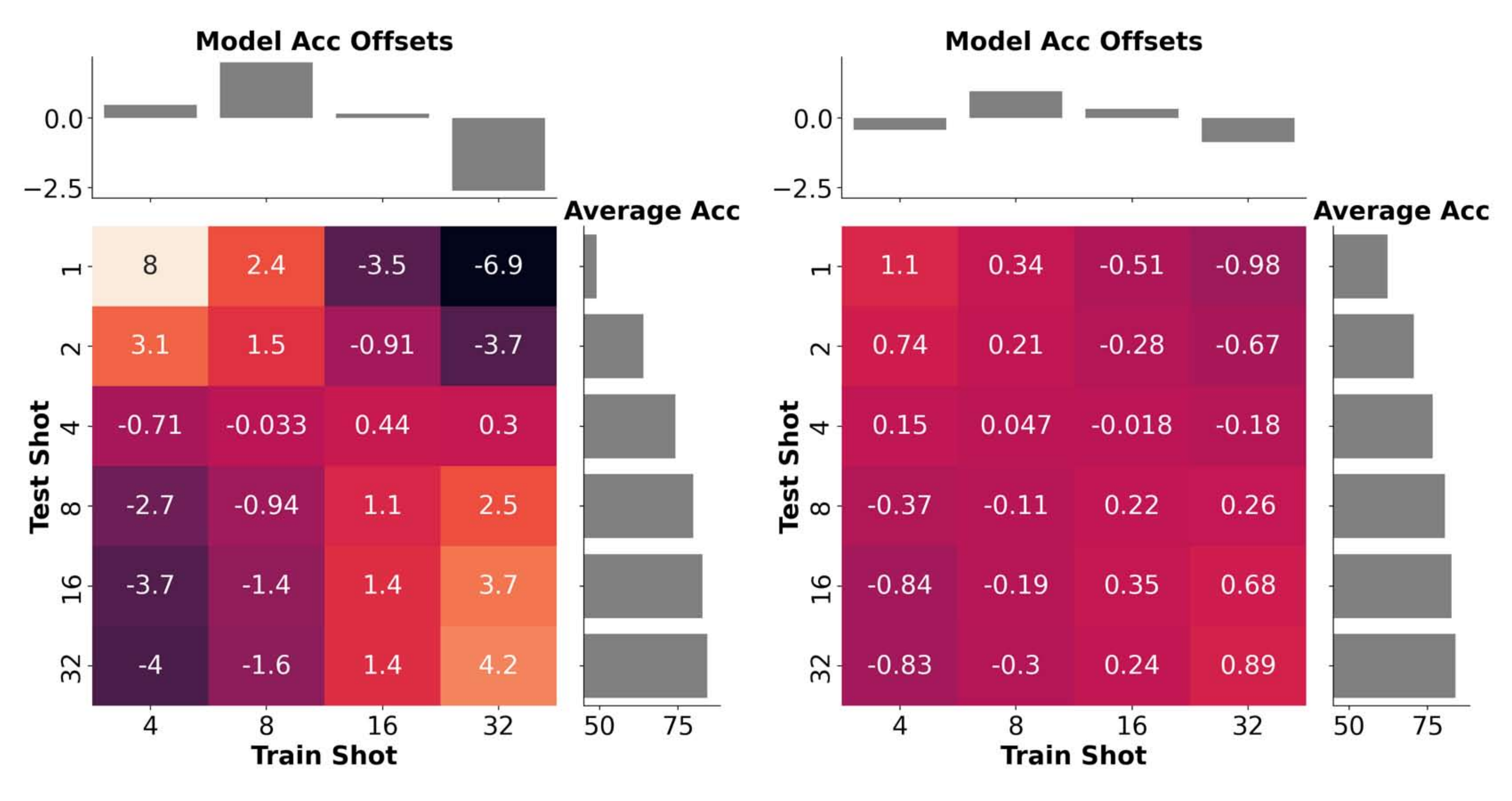}
        \caption{ProtoNet vs Cosine ProtoNet on meta-iNat, both using a Conv-4 backbone.}
        \label{fig:supp_heatmap_inat_proto}
        \end{figure*}
        
    \begin{table}
      \centering
      \scriptsize
      \tiny
      \begin{tabular}{c c | c c c c | c}
        \toprule
        \multicolumn{2}{c|}{\textbf{meta-iNat Conv-4}} & \multicolumn{5}{c}{\textbf{Train Shot}} \\
        \multicolumn{2}{c|}{\textbf{Proto}} & 4 & 8 & 16 & 32 & Mean \\
        \hline
        \Tstrut\multirow{7}{*}{\makecell{\textbf{Test}\\\textbf{Shot}}} 
        & 1 & 57.47 & 53.48 & 45.69 & 39.53 & 49.04 \\
& 2 & 67.52 & 67.47 & 63.20 & 57.61 & 63.95 \\
& 4 & 73.93 & 76.13 & 74.77 & 71.86 & 74.17 \\
& 8 & 77.64 & 80.91 & 81.16 & 79.72 & 79.86 \\
& 16 & 79.53 & 83.36 & 84.34 & 83.82 & 82.76 \\
& 32 & 80.81 & 84.68 & 85.86 & 85.86 & 84.30 \\
\cline{2-7}
\Tstrut & Offset & 0.47 & 1.99 & 0.16 & -2.61 & 14.86 \\
        \bottomrule
      \end{tabular}
      \quad
      \begin{tabular}{c c | c c c c | c}
        \toprule
        \multicolumn{2}{c|}{\textbf{meta-iNat Conv-4}} & \multicolumn{5}{c}{\textbf{Train Shot}} \\
        \multicolumn{2}{c|}{\textbf{Cosine Proto}} & 4 & 8 & 16 & 32 & Mean \\
        \hline
        \Tstrut\multirow{7}{*}{\makecell{\textbf{Test}\\\textbf{Shot}}} 
        & 1 & 63.09 & 63.68 & 62.19 & 60.53 & 62.37 \\
& 2 & 70.91 & 71.77 & 70.64 & 69.06 & 70.60 \\
& 4 & 76.37 & 77.66 & 76.96 & 75.61 & 76.65 \\
& 8 & 79.73 & 81.38 & 81.08 & 79.93 & 80.53 \\
& 16 & 81.42 & 83.46 & 83.37 & 82.51 & 82.69 \\
& 32 & 82.66 & 84.58 & 84.48 & 83.94 & 83.92 \\
\cline{2-7}
\Tstrut & Offset & -0.43 & 0.96 & 0.33 & -0.86 & 2.13 \\
        \bottomrule
      \end{tabular}
      \caption{Accuracy comparison for ProtoNet / Cosine ProtoNet on meta-iNat. Average group accuracy is displayed in the rightmost column of each table, model offsets are given in the bottom row. Bottom right cell (intersection of Offset and Mean) shows the sensitivity score.}
      \label{tab:supp_tab_inat_proto}
    \end{table}

    \begin{figure*}
        \centering
        \includegraphics[width=.8\linewidth]{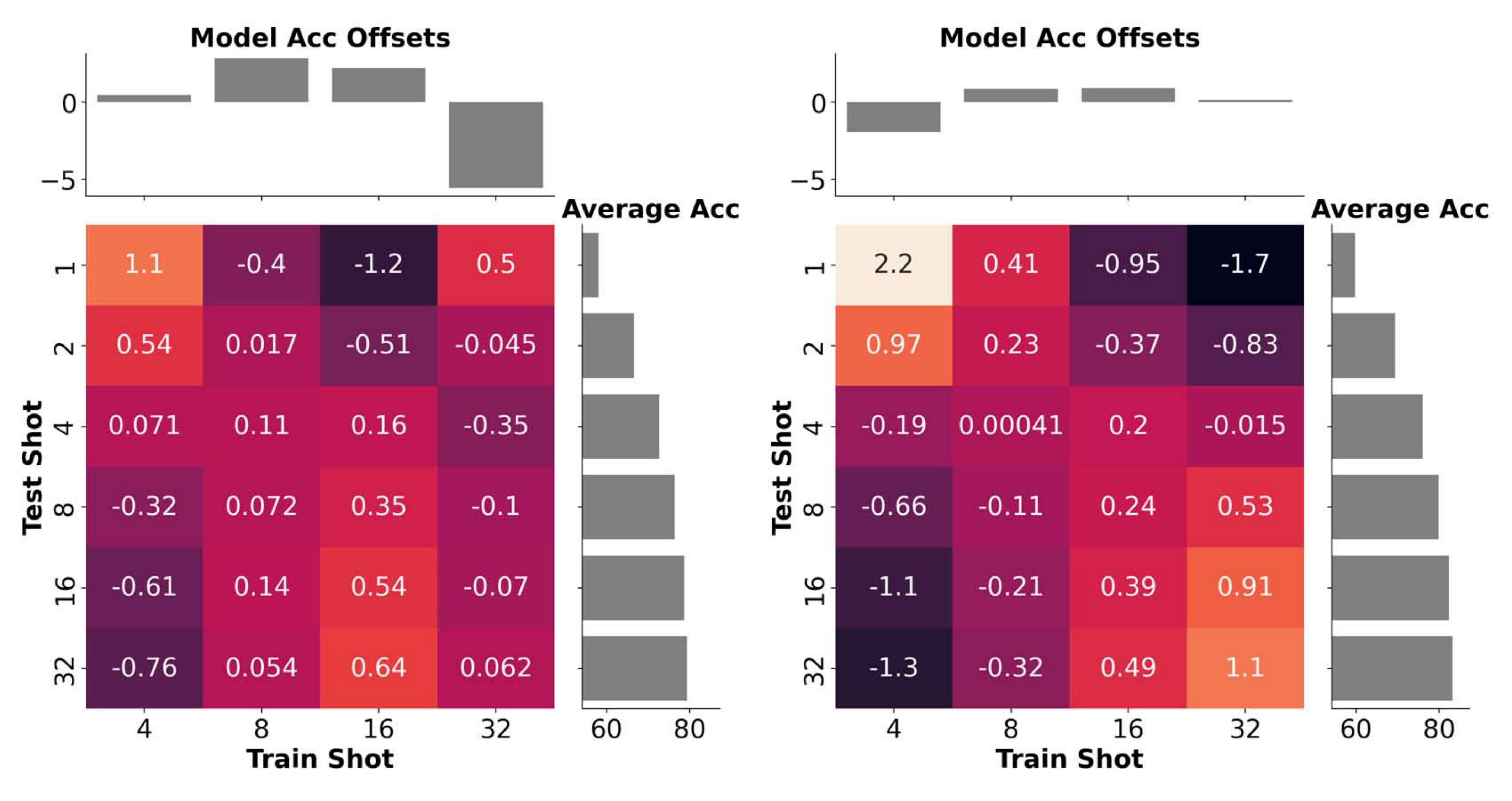}
        \caption{PCA vs EST on meta-iNat, both using a Conv-4 backbone.}
        \label{fig:supp_heatmap_inat_pca_est}
        \end{figure*}
    
       \begin{table}
      \centering
      \tiny
      \begin{tabular}{c c | c c c c | c}
        \toprule
        \multicolumn{2}{c|}{\textbf{meta-iNat Conv-4}} & \multicolumn{5}{c}{\textbf{Train Shot}} \\
        \multicolumn{2}{c|}{\textbf{PCA}} & 4 & 8 & 16 & 32 & Mean \\
        \hline
        \Tstrut\multirow{7}{*}{\makecell{\textbf{Test}\\\textbf{Shot}}} 
& 1 & 59.71 & 60.61 & 59.19 & 53.11 & 58.15 \\
& 2 & 67.68 & 69.53 & 68.36 & 61.07 & 66.66 \\
& 4 & 73.22 & 75.64 & 75.05 & 66.78 & 72.67 \\
& 8 & 76.53 & 79.30 & 78.94 & 70.73 & 76.38 \\
& 16 & 78.52 & 81.65 & 81.41 & 73.04 & 78.65 \\
& 32 & 79.09 & 82.28 & 82.23 & 73.89 & 79.37 \\
\cline{2-7}
\Tstrut & Offset & 0.48 & 2.85 & 2.22 & -5.54 & 2.26 \\
        \bottomrule
      \end{tabular}
      \quad
      \begin{tabular}{c c | c c c c | c}
        \toprule
        \multicolumn{2}{c|}{\textbf{meta-iNat Conv-4}} & \multicolumn{5}{c}{\textbf{Train Shot}} \\
        \multicolumn{2}{c|}{\textbf{EST}} & 4 & 8 & 16 & 32 & Mean \\
        \hline
        \Tstrut\multirow{7}{*}{\makecell{\textbf{Test}\\\textbf{Shot}}} 
& 1 & 60.08 & 61.07 & 59.75 & 58.23 & 59.78 \\
& 2 & 68.38 & 70.46 & 69.90 & 68.67 & 69.35 \\
& 4 & 73.87 & 76.88 & 77.12 & 76.14 & 76.00 \\
& 8 & 77.30 & 80.67 & 81.06 & 80.58 & 79.90 \\
& 16 & 79.33 & 83.02 & 83.66 & 83.42 & 82.36 \\
& 32 & 79.87 & 83.67 & 84.52 & 84.38 & 83.11 \\
\cline{2-7}
\Tstrut & Offset & -1.95 & 0.88 & 0.92 & 0.15 & 3.95 \\
        \bottomrule
      \end{tabular}
      \caption{Accuracy comparison for PCA and EST on meta-iNat. Average group accuracy is displayed in the rightmost column of each table, model offsets are given in the bottom row. Bottom right cell (intersection of Offset and Mean) shows the sensitivity score.}
      \label{tab:supp_tab_inat_pca_est}
    \end{table}

\clearpage
    
        \begin{figure*}
        \centering
        \includegraphics[width=.8\linewidth]{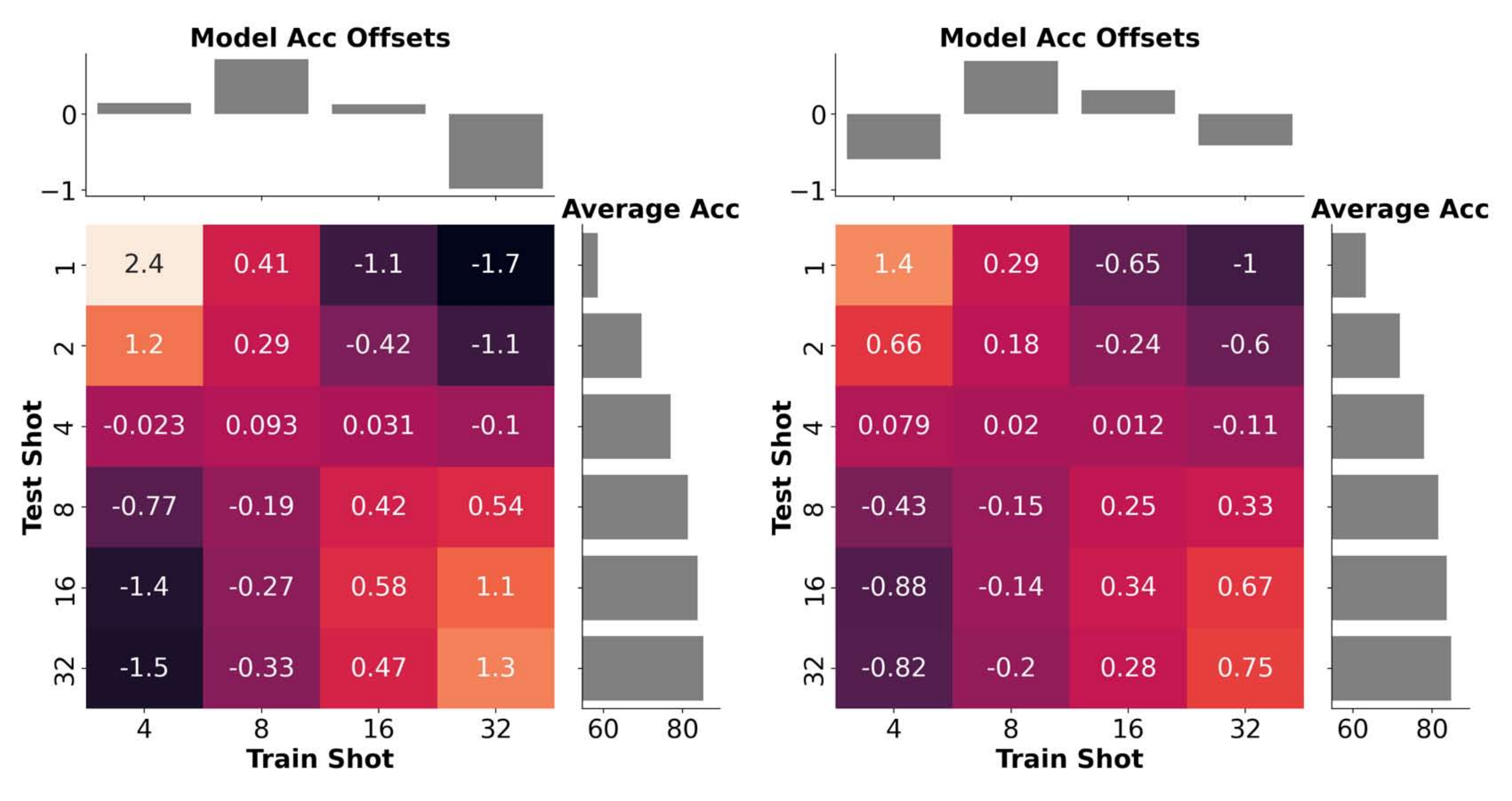}
        \caption{FEAT vs Cosine FEAT on meta-iNat, both using a Conv-4 backbone.}
        \label{fig:supp_heatmap_inat_feat}
        \end{figure*}
    
       \begin{table}
      \centering
      \tiny
      \begin{tabular}{c c | c c c c | c}
        \toprule
        \multicolumn{2}{c|}{\textbf{meta-iNat Conv-4}} & \multicolumn{5}{c}{\textbf{Train Shot}} \\
        \multicolumn{2}{c|}{\textbf{FEAT}} & 4 & 8 & 16 & 32 & Mean \\
        \hline
        \Tstrut\multirow{7}{*}{\makecell{\textbf{Test}\\\textbf{Shot}}} 
& 1 & 61.13 & 59.70 & 57.63 & 55.85 & 58.58 \\
& 2 & 70.89 & 70.55 & 69.25 & 67.50 & 69.55 \\
& 4 & 77.07 & 77.76 & 77.11 & 75.87 & 76.95 \\
& 8 & 80.72 & 81.88 & 81.90 & 80.91 & 81.35 \\
& 16 & 82.58 & 84.25 & 84.51 & 83.89 & 83.81 \\
& 32 & 83.87 & 85.57 & 85.78 & 85.51 & 85.18 \\
\cline{2-7}
\Tstrut & Offset & 0.14 & 0.72 & 0.13 & -0.98 & 4.16 \\
        \bottomrule
      \end{tabular}
      \quad
      \begin{tabular}{c c | c c c c | c}
        \toprule
        \multicolumn{2}{c|}{\textbf{meta-iNat Conv-4}} & \multicolumn{5}{c}{\textbf{Train Shot}} \\
        \multicolumn{2}{c|}{\textbf{Cosine FEAT}} & 4 & 8 & 16 & 32 & Mean \\
        \hline
        \Tstrut\multirow{7}{*}{\makecell{\textbf{Test}\\\textbf{Shot}}} 
& 1 & 64.02 & 64.20 & 62.88 & 61.77 & 63.22 \\
& 2 & 71.93 & 72.74 & 71.94 & 70.85 & 71.86 \\
& 4 & 77.43 & 78.66 & 78.27 & 77.42 & 77.94 \\
& 8 & 80.57 & 82.14 & 82.16 & 81.51 & 81.60 \\
& 16 & 82.16 & 84.19 & 84.29 & 83.89 & 83.63 \\
& 32 & 83.32 & 85.23 & 85.33 & 85.07 & 84.74 \\
\cline{2-7}
\Tstrut & Offset & -0.59 & 0.69 & 0.31 & -0.41 & 2.43 \\
        \bottomrule
      \end{tabular}
      \caption{Accuracy comparison for FEAT and Cosine FEAT on meta-iNat. Average group accuracy is displayed in the rightmost column of each table, model offsets are given in the bottom row. Bottom right cell (intersection of Offset and Mean) shows the sensitivity score.}
      \label{tab:supp_tab_inat_feat}
    \end{table}

        \begin{figure*}
        \centering
        \includegraphics[width=.8\linewidth]{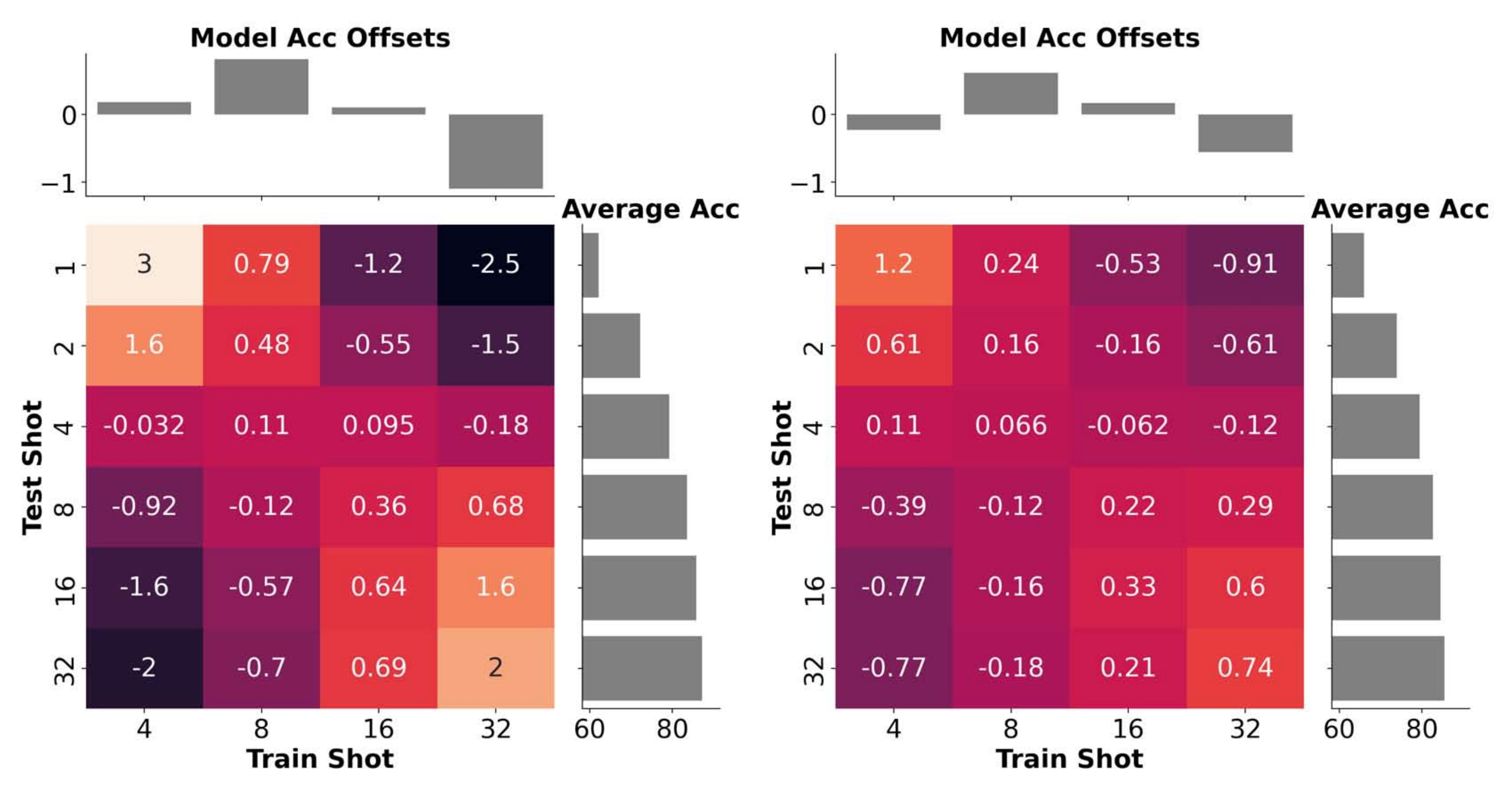}
        \caption{FRN vs Cosine FRN on meta-iNat, both using a Conv-4 backbone.}
        \label{fig:supp_heatmap_inat_frn}
        \end{figure*}
    
       \begin{table}
      \centering
      \tiny
      \begin{tabular}{c c | c c c c | c}
        \toprule
        \multicolumn{2}{c|}{\textbf{meta-iNat Conv-4}} & \multicolumn{5}{c}{\textbf{Train Shot}} \\
        \multicolumn{2}{c|}{\textbf{FRN}} & 4 & 8 & 16 & 32 & Mean \\
        \hline
        \Tstrut\multirow{7}{*}{\makecell{\textbf{Test}\\\textbf{Shot}}} 
& 1 & 65.30 & 63.77 & 61.03 & 58.56 & 62.17 \\
& 2 & 74.01 & 73.52 & 71.78 & 69.58 & 72.22 \\
& 4 & 79.37 & 80.15 & 79.42 & 77.95 & 79.22 \\
& 8 & 82.78 & 84.21 & 83.98 & 83.10 & 83.52 \\
& 16 & 84.36 & 86.05 & 86.55 & 86.28 & 85.81 \\
& 32 & 85.37 & 87.29 & 87.97 & 88.07 & 87.18 \\
\cline{2-7}
\Tstrut & Offset & 0.18 & 0.81 & 0.10 & -1.10 & 5.46 \\
        \bottomrule
      \end{tabular}
      \quad
      \begin{tabular}{c c | c c c c | c}
        \toprule
        \multicolumn{2}{c|}{\textbf{meta-iNat Conv-4}} & \multicolumn{5}{c}{\textbf{Train Shot}} \\
        \multicolumn{2}{c|}{\textbf{Cosine FRN}} & 4 & 8 & 16 & 32 & Mean \\
        \hline
        \Tstrut\multirow{7}{*}{\makecell{\textbf{Test}\\\textbf{Shot}}} 
& 1 & 66.97 & 66.85 & 65.64 & 64.53 & 66.00 \\
& 2 & 74.28 & 74.67 & 73.91 & 72.74 & 73.90 \\
& 4 & 79.33 & 80.13 & 79.56 & 78.78 & 79.45 \\
& 8 & 82.11 & 83.22 & 83.12 & 82.47 & 82.73 \\
& 16 & 83.52 & 84.97 & 85.02 & 84.56 & 84.52 \\
& 32 & 84.48 & 85.92 & 85.86 & 85.67 & 85.48 \\
\cline{2-7}
\Tstrut & Offset & -0.23 & 0.61 & 0.17 & -0.55 & 2.12 \\
        \bottomrule
      \end{tabular}
      \caption{Accuracy comparison for FRN and Cosine FRN on meta-iNat. Average group accuracy is displayed in the rightmost column of each table, model offsets are given in the bottom row. Bottom right cell (intersection of Offset and Mean) shows the sensitivity score.}
      \label{tab:supp_tab_inat_frn}
    \end{table}
    
    \clearpage

\begin{figure*}
\centering
\includegraphics[width=.8\linewidth]{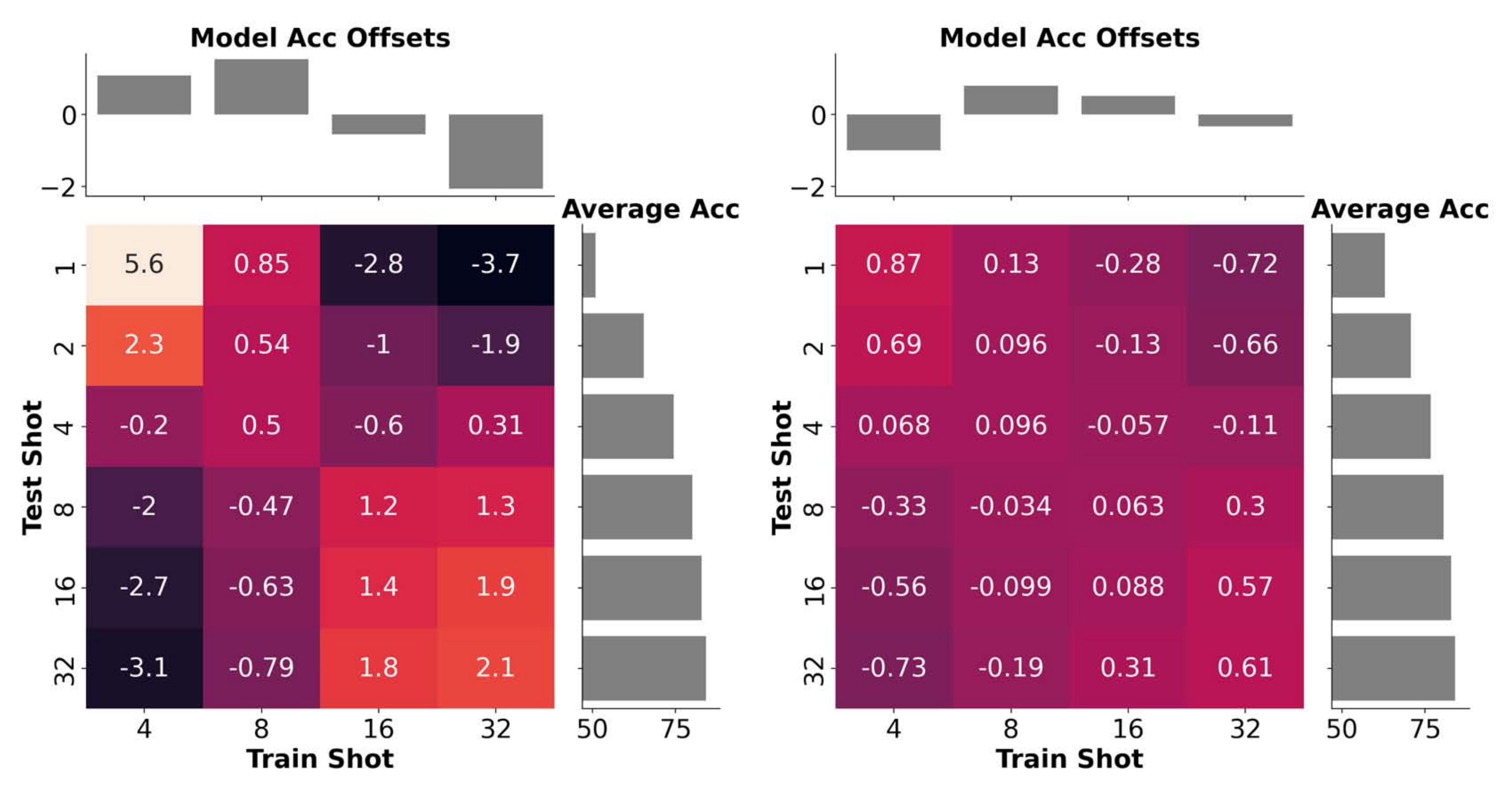}
\caption{Proto vs Cosine Proto on CUB, both using a Conv-4 backbone.}
\label{fig:supp_heatmap_cub_proto}
\end{figure*}

\begin{table}
\centering
\tiny
\begin{tabular}{c c | c c c c | c}
\toprule
\multicolumn{2}{c|}{\textbf{CUB Conv-4}} & \multicolumn{5}{c}{\textbf{Train Shot}} \\
\multicolumn{2}{c|}{\textbf{Proto}} & 4 & 8 & 16 & 32 & Mean \\
\hline
\Tstrut\multirow{7}{*}{\makecell{\textbf{Test}\\\textbf{Shot}}} 
& 1 & 57.58 & 53.24 & 47.51 & 45.10 & 50.86 \\
& 2 & 68.80 & 67.47 & 63.85 & 61.46 & 65.39 \\
& 4 & 75.42 & 76.58 & 73.39 & 72.78 & 74.54 \\
& 8 & 79.15 & 81.09 & 80.63 & 79.22 & 80.02 \\
& 16 & 81.16 & 83.72 & 83.70 & 82.67 & 82.81 \\
& 32 & 82.22 & 84.96 & 85.47 & 84.21 & 84.21 \\
\cline{2-7}
\Tstrut & Offset & 1.08 & 1.54 & -0.55 & -2.07 & 9.33 \\
\bottomrule
\end{tabular}
\quad
\begin{tabular}{c c | c c c c | c}
\toprule
\multicolumn{2}{c|}{\textbf{CUB Conv-4}} & \multicolumn{5}{c}{\textbf{Train Shot}} \\
\multicolumn{2}{c|}{\textbf{Cosine Proto}} & 4 & 8 & 16 & 32 & Mean \\
\hline
\Tstrut\multirow{7}{*}{\makecell{\textbf{Test}\\\textbf{Shot}}} 
& 1 & 62.72 & 63.77 & 63.09 & 61.80 & 62.85 \\
& 2 & 70.48 & 71.68 & 71.18 & 69.80 & 70.79 \\
& 4 & 75.85 & 77.67 & 77.24 & 76.34 & 76.77 \\
& 8 & 79.29 & 81.38 & 81.20 & 80.59 & 80.61 \\
& 16 & 81.31 & 83.56 & 83.47 & 83.10 & 82.86 \\
& 32 & 82.36 & 84.70 & 84.92 & 84.37 & 84.09 \\
\cline{2-7}
\Tstrut & Offset & -0.99 & 0.80 & 0.52 & -0.33 & 1.60 \\
\bottomrule
\end{tabular}
\caption{Accuracy comparison for ProtoNet / Cosine ProtoNet on CUB. Average group accuracy is displayed in the rightmost column of each table, model offsets are given in the bottom row. Bottom right cell (intersection of Offset and Mean) shows the sensitivity score.}
\label{tab:supp_tab_cub_proto}
\end{table}

\begin{figure*}
\centering
\includegraphics[width=.8\linewidth]{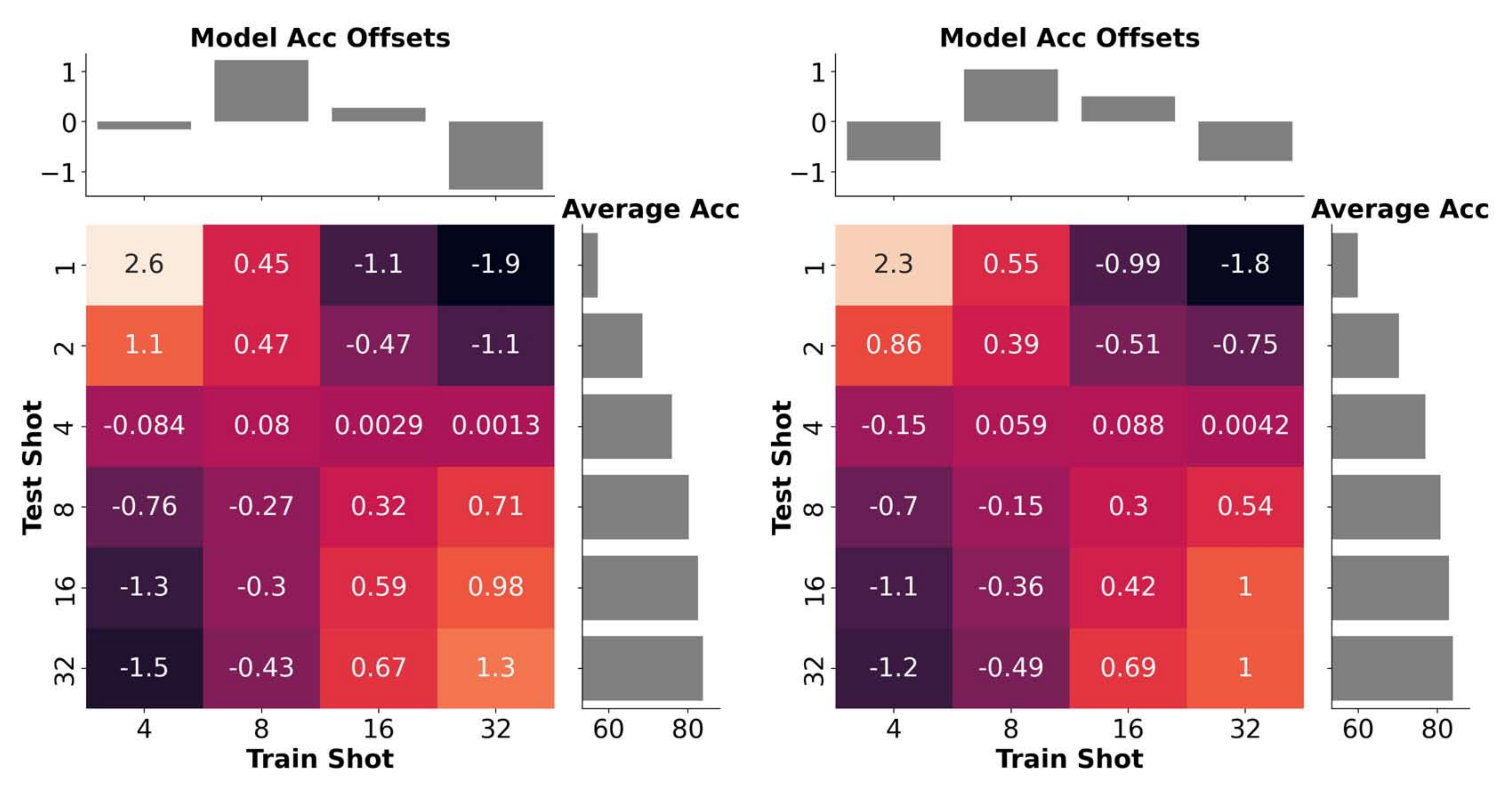}
\caption{PCA vs EST on CUB, both using a Conv-4 backbone.}
\label{fig:supp_heatmap_cub_est}
\end{figure*}
\begin{table}
\centering
\tiny
\begin{tabular}{c c | c c c c | c}
\toprule
\multicolumn{2}{c|}{\textbf{CUB Conv-4}} & \multicolumn{5}{c}{\textbf{Train Shot}} \\
\multicolumn{2}{c|}{\textbf{PCA}} & 4 & 8 & 16 & 32 & Mean \\
\hline
\Tstrut\multirow{7}{*}{\makecell{\textbf{Test}\\\textbf{Shot}}} 
& 1 & 59.72 & 59.00 & 56.48 & 54.06 & 57.31 \\
& 2 & 69.40 & 70.18 & 68.29 & 66.06 & 68.48 \\
& 4 & 75.68 & 77.23 & 76.20 & 74.57 & 75.92 \\
& 8 & 79.29 & 81.17 & 80.81 & 79.57 & 80.21 \\
& 16 & 81.15 & 83.51 & 83.44 & 82.21 & 82.58 \\
& 32 & 82.10 & 84.57 & 84.72 & 83.70 & 83.77 \\
\cline{2-7}
\Tstrut & Offset & -0.16 & 1.23 & 0.28 & -1.35 & 4.46 \\
\bottomrule
\end{tabular}
\quad
\begin{tabular}{c c | c c c c | c}
\toprule
\multicolumn{2}{c|}{\textbf{CUB Conv-4}} & \multicolumn{5}{c}{\textbf{Train Shot}} \\
\multicolumn{2}{c|}{\textbf{EST}} & 4 & 8 & 16 & 32 & Mean \\
\hline
\Tstrut\multirow{7}{*}{\makecell{\textbf{Test}\\\textbf{Shot}}} 
& 1 & 61.42 & 61.53 & 59.45 & 57.35 & 59.94 \\
& 2 & 70.32 & 71.67 & 70.23 & 68.70 & 70.23 \\
& 4 & 76.06 & 78.09 & 77.58 & 76.21 & 76.99 \\
& 8 & 79.31 & 81.68 & 81.59 & 80.54 & 80.78 \\
& 16 & 81.01 & 83.53 & 83.77 & 83.07 & 82.85 \\
& 32 & 81.88 & 84.42 & 85.07 & 84.10 & 83.87 \\
\cline{2-7}
\Tstrut & Offset & -0.77 & 1.05 & 0.51 & -0.78 & 4.07 \\
\bottomrule
\end{tabular}
\caption{Accuracy comparison for PCA and EST on CUB. Average group accuracy is displayed in the rightmost column of each table, model offsets are given in the bottom row. Bottom right cell (intersection of Offset and Mean) shows the sensitivity score.}
\label{tab:supp_tab_cub_est}
\end{table}

\clearpage

\begin{figure*}
\centering
\includegraphics[width=.8\linewidth]{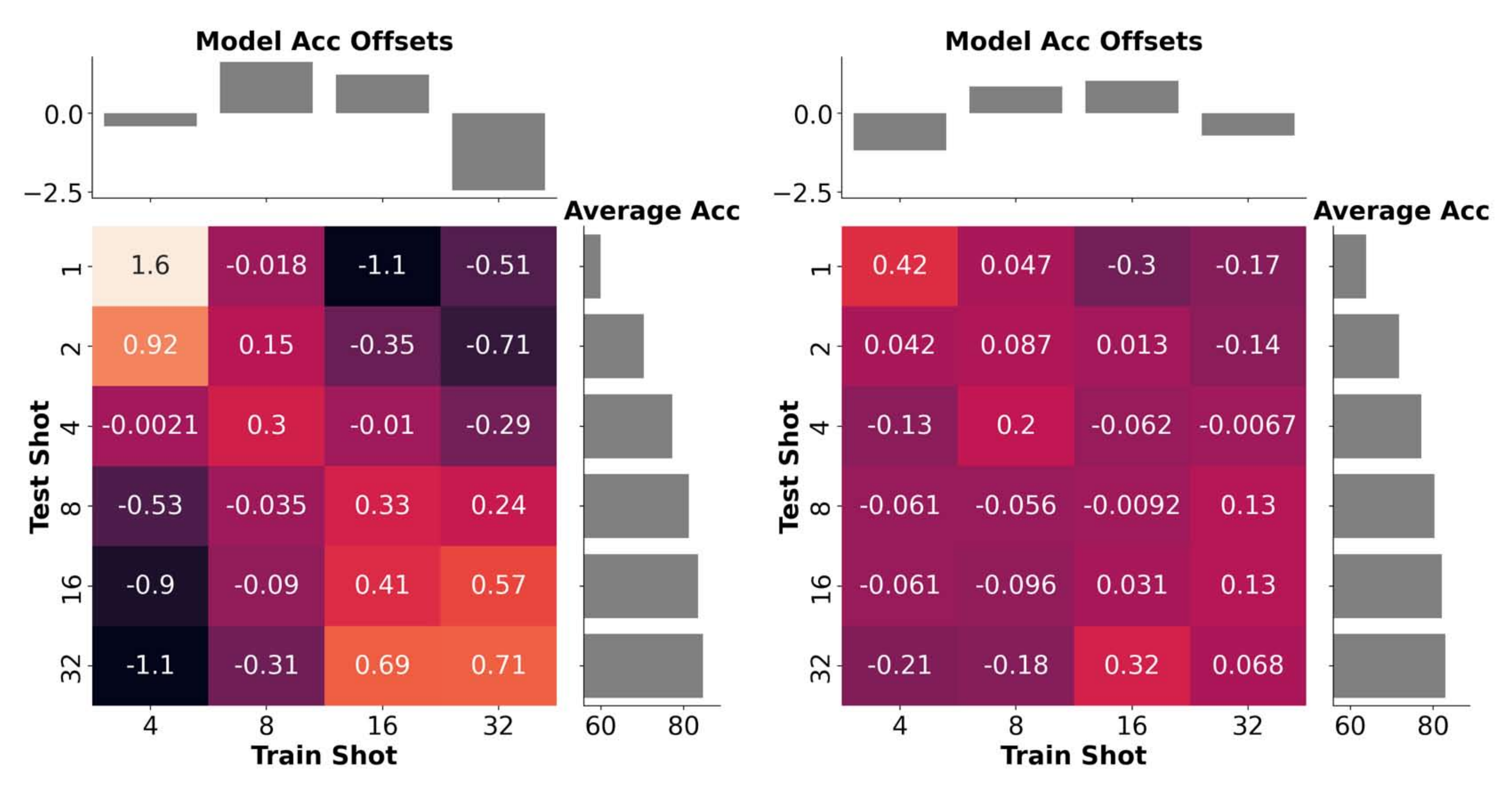}
\caption{FEAT vs Cosine FEAT on CUB, both using a Conv-4 backbone.}
\label{fig:supp_heatmap_cub_feat}
\end{figure*}
\begin{table}
\centering
\tiny
\begin{tabular}{c c | c c c c | c}
\toprule
\multicolumn{2}{c|}{\textbf{CUB Conv-4}} & \multicolumn{5}{c}{\textbf{Train Shot}} \\
\multicolumn{2}{c|}{\textbf{FEAT}} & 4 & 8 & 16 & 32 & Mean \\
\hline
\Tstrut\multirow{7}{*}{\makecell{\textbf{Test}\\\textbf{Shot}}} 
& 1 & 61.05 & 61.46 & 60.01 & 56.89 & 59.85 \\
& 2 & 70.89 & 72.16 & 71.26 & 67.22 & 70.38 \\
& 4 & 76.83 & 79.17 & 78.46 & 74.50 & 77.24 \\
& 8 & 80.34 & 82.87 & 82.84 & 79.07 & 81.28 \\
& 16 & 82.19 & 85.03 & 85.14 & 81.62 & 83.50 \\
& 32 & 83.15 & 85.96 & 86.57 & 82.91 & 84.65 \\
\cline{2-7}
\Tstrut & Offset & -0.41 & 1.63 & 1.23 & -2.45 & 2.70 \\
\bottomrule
\end{tabular}
\quad
\begin{tabular}{c c | c c c c | c}
\toprule
\multicolumn{2}{c|}{\textbf{CUB Conv-4}} & \multicolumn{5}{c}{\textbf{Train Shot}} \\
\multicolumn{2}{c|}{\textbf{Cosine FEAT}} & 4 & 8 & 16 & 32 & Mean \\
\hline
\Tstrut\multirow{7}{*}{\makecell{\textbf{Test}\\\textbf{Shot}}} 
& 1 & 62.99 & 64.65 & 64.49 & 62.88 & 63.75 \\
& 2 & 70.58 & 72.66 & 72.77 & 70.88 & 71.72 \\
& 4 & 75.79 & 78.16 & 78.08 & 76.40 & 77.11 \\
& 8 & 79.02 & 81.06 & 81.29 & 79.69 & 80.26 \\
& 16 & 80.79 & 82.79 & 83.10 & 81.46 & 82.03 \\
& 32 & 81.49 & 83.55 & 84.24 & 82.25 & 82.88 \\
\cline{2-7}
\Tstrut & Offset & -1.18 & 0.85 & 1.03 & -0.70 & 0.72 \\
\bottomrule
\end{tabular}
\caption{Accuracy comparison for FEAT and Cosine FEAT on CUB. Average group accuracy is displayed in the rightmost column of each table, model offsets are given in the bottom row. Bottom right cell (intersection of Offset and Mean) shows the sensitivity score.}
\label{tab:supp_tab_cub_feat}
\end{table}

\begin{figure*}
\centering
\includegraphics[width=.8\linewidth]{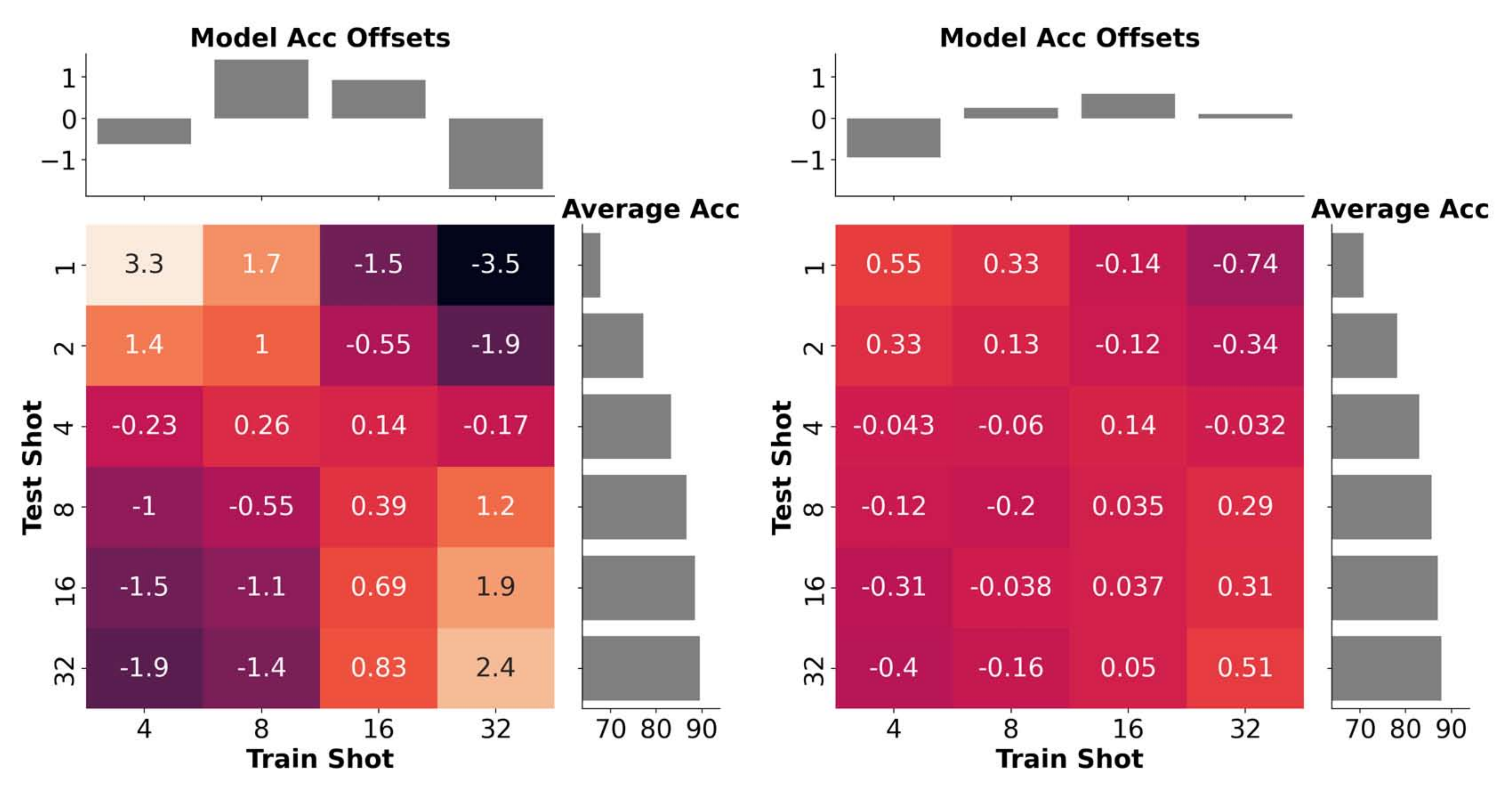}
\caption{FRN vs Cosine FRN on CUB, both using a Conv-4 backbone.}
\label{fig:supp_heatmap_cub_frn}
\end{figure*}
\begin{table}
\centering
\tiny
\begin{tabular}{c c | c c c c | c}
\toprule
\multicolumn{2}{c|}{\textbf{CUB Conv-4}} & \multicolumn{5}{c}{\textbf{Train Shot}} \\
\multicolumn{2}{c|}{\textbf{FRN}} & 4 & 8 & 16 & 32 & Mean \\
\hline
\Tstrut\multirow{7}{*}{\makecell{\textbf{Test}\\\textbf{Shot}}} 
& 1 & 70.36 & 70.86 & 67.17 & 62.56 & 67.74 \\
& 2 & 77.89 & 79.59 & 77.53 & 73.57 & 77.14 \\
& 4 & 82.43 & 84.98 & 84.36 & 81.40 & 83.29 \\
& 8 & 84.98 & 87.49 & 87.94 & 86.06 & 86.62 \\
& 16 & 86.34 & 88.83 & 90.11 & 88.67 & 88.49 \\
& 32 & 86.98 & 89.56 & 91.25 & 90.18 & 89.49 \\
\cline{2-7}
\Tstrut & Offset & -0.63 & 1.42 & 0.93 & -1.72 & 6.71 \\
\bottomrule
\end{tabular}
\quad
\begin{tabular}{c c | c c c c | c}
\toprule
\multicolumn{2}{c|}{\textbf{CUB Conv-4}} & \multicolumn{5}{c}{\textbf{Train Shot}} \\
\multicolumn{2}{c|}{\textbf{Cosine FRN}} & 4 & 8 & 16 & 32 & Mean \\
\hline
\Tstrut\multirow{7}{*}{\makecell{\textbf{Test}\\\textbf{Shot}}} 
& 1 & 70.38 & 71.37 & 71.24 & 70.14 & 70.78 \\
& 2 & 77.49 & 78.50 & 78.58 & 77.87 & 78.11 \\
& 4 & 81.96 & 83.15 & 83.68 & 83.02 & 82.95 \\
& 8 & 84.55 & 85.68 & 86.25 & 86.01 & 85.62 \\
& 16 & 85.80 & 87.28 & 87.69 & 87.47 & 87.06 \\
& 32 & 86.42 & 87.86 & 88.41 & 88.38 & 87.77 \\
\cline{2-7}
\Tstrut & Offset & -0.95 & 0.26 & 0.59 & 0.10 & 1.29 \\
\bottomrule
\end{tabular}
\caption{Accuracy comparison for FRN and Cosine FRN on CUB. Average group accuracy is displayed in the rightmost column of each table, model offsets are given in the bottom row. Bottom right cell (intersection of Offset and Mean) shows the sensitivity score.}
\label{tab:supp_tab_cub_frn}
\end{table}

\clearpage

\begin{figure*}
\centering
\includegraphics[width=.8\linewidth]{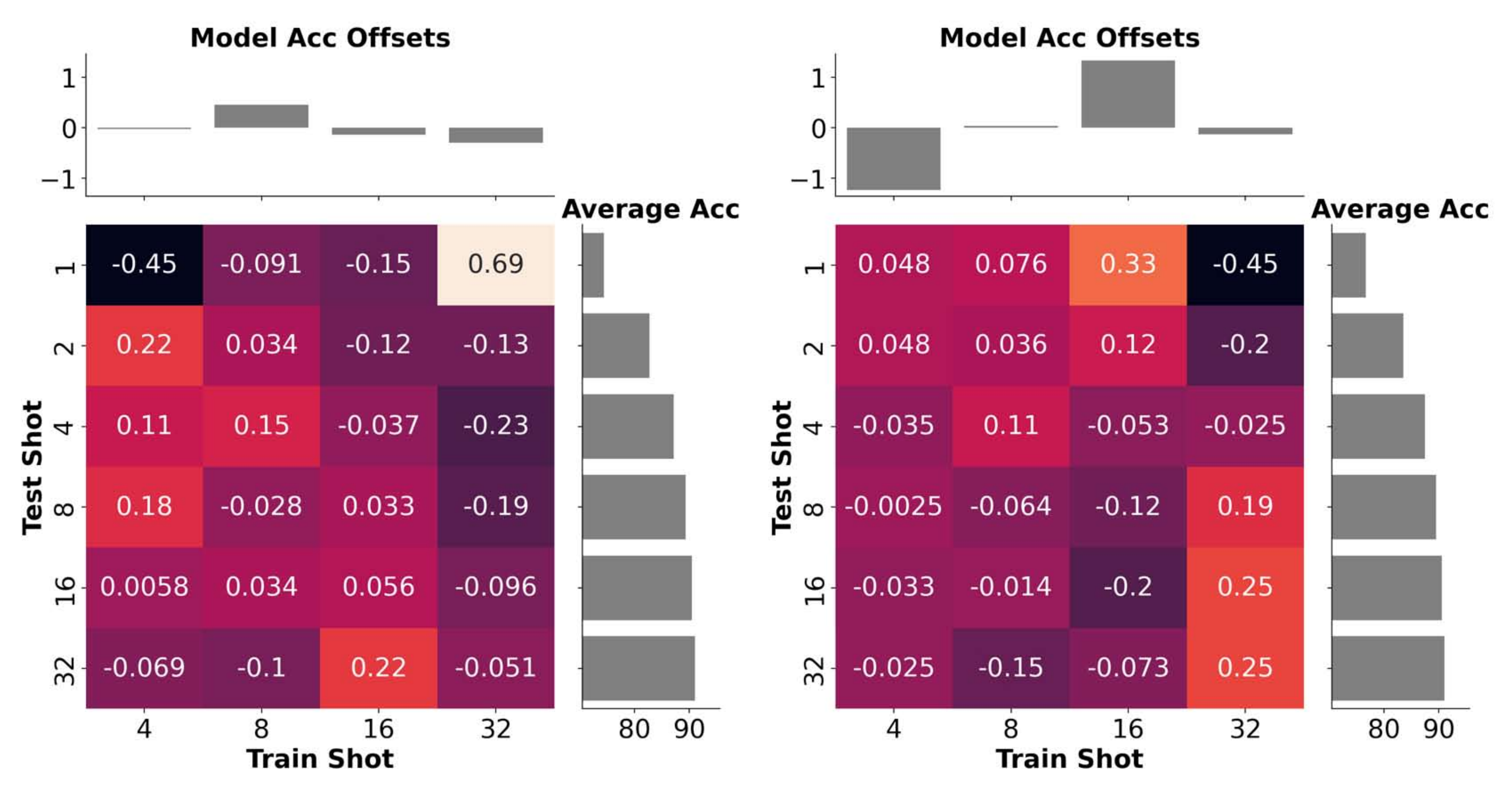}
\caption{ProtoNet vs Cosine ProtoNet on CUB, both using a ResNet-12 backbone.}
\label{fig:supp_heatmap_cub_proto_12}
\end{figure*}
\begin{table}
\centering
\tiny
\begin{tabular}{c c | c c c c | c}
\toprule
\multicolumn{2}{c|}{\textbf{CUB ResNet-12}} & \multicolumn{5}{c}{\textbf{Train Shot}} \\
\multicolumn{2}{c|}{\textbf{Proto}} & 4 & 8 & 16 & 32 & Mean \\
\hline
\Tstrut\multirow{7}{*}{\makecell{\textbf{Test}\\\textbf{Shot}}} 
& 1 & 73.96 & 74.80 & 74.15 & 74.83 & 74.43 \\
& 2 & 82.89 & 83.19 & 82.44 & 82.28 & 82.70 \\
& 4 & 87.24 & 87.76 & 86.98 & 86.63 & 87.15 \\
& 8 & 89.52 & 89.79 & 89.26 & 88.88 & 89.36 \\
& 16 & 90.43 & 90.94 & 90.37 & 90.06 & 90.45 \\
& 32 & 90.91 & 91.36 & 91.09 & 90.66 & 91.00 \\
\cline{2-7}
\Tstrut & Offset & -0.03 & 0.46 & -0.14 & -0.29 & 1.14 \\
\bottomrule
\end{tabular}
\quad
\begin{tabular}{c c | c c c c | c}
\toprule
\multicolumn{2}{c|}{\textbf{CUB ResNet-12}} & \multicolumn{5}{c}{\textbf{Train Shot}} \\
\multicolumn{2}{c|}{\textbf{Cosine Proto}} & 4 & 8 & 16 & 32 & Mean \\
\hline
\Tstrut\multirow{7}{*}{\makecell{\textbf{Test}\\\textbf{Shot}}} 
& 1 & 75.49 & 76.78 & 78.33 & 76.09 & 76.67 \\
& 2 & 82.32 & 83.57 & 84.95 & 83.17 & 83.50 \\
& 4 & 86.24 & 87.65 & 88.78 & 87.35 & 87.50 \\
& 8 & 88.28 & 89.48 & 90.72 & 89.57 & 89.51 \\
& 16 & 89.26 & 90.54 & 91.65 & 90.64 & 90.52 \\
& 32 & 89.79 & 90.93 & 92.30 & 91.16 & 91.05 \\
\cline{2-7}
\Tstrut & Offset & -1.23 & 0.03 & 1.33 & -0.13 & 0.78 \\
\bottomrule
\end{tabular}
\caption{Accuracy comparison for ProtoNet / Cosine ProtoNet on CUB. Average group accuracy is displayed in the rightmost column of each table, model offsets are given in the bottom row. Bottom right cell (intersection of Offset and Mean) shows the sensitivity score.}
\label{tab:supp_tab_cub_proto_12}
\end{table}

\begin{figure*}
\centering
\includegraphics[width=.8\linewidth]{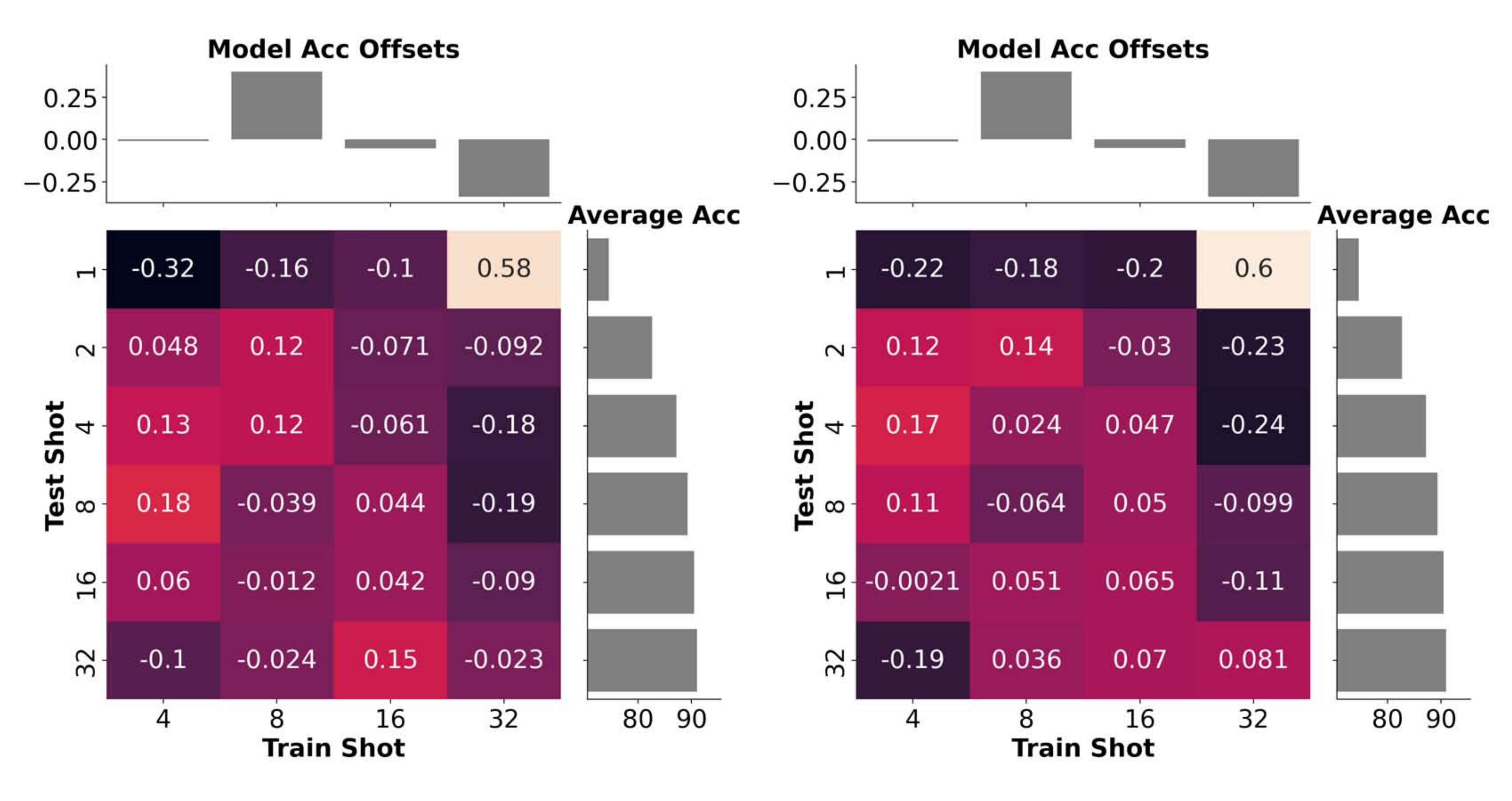}
\caption{PCA vs EST on CUB, both using a ResNet-12 backbone.}
\label{fig:supp_heatmap_cub_est_12}
\end{figure*}
\begin{table}
\centering
\tiny
\begin{tabular}{c c | c c c c | c}
\toprule
\multicolumn{2}{c|}{\textbf{CUB ResNet-12}} & \multicolumn{5}{c}{\textbf{Train Shot}} \\
\multicolumn{2}{c|}{\textbf{PCA}} & 4 & 8 & 16 & 32 & Mean \\
\hline
\Tstrut\multirow{7}{*}{\makecell{\textbf{Test}\\\textbf{Shot}}} 
& 1 & 74.20 & 74.77 & 74.37 & 74.76 & 74.53 \\
& 2 & 82.66 & 83.14 & 82.50 & 82.19 & 82.62 \\
& 4 & 87.26 & 87.66 & 87.03 & 86.62 & 87.14 \\
& 8 & 89.44 & 89.63 & 89.26 & 88.74 & 89.27 \\
& 16 & 90.50 & 90.84 & 90.44 & 90.02 & 90.45 \\
& 32 & 90.88 & 91.37 & 91.09 & 90.63 & 90.99 \\
\cline{2-7}
\Tstrut & Offset & -0.01 & 0.40 & -0.05 & -0.34 & 0.89 \\
\bottomrule
\end{tabular}
\quad
\begin{tabular}{c c | c c c c | c}
\toprule
\multicolumn{2}{c|}{\textbf{CUB ResNet-12}} & \multicolumn{5}{c}{\textbf{Train Shot}} \\
\multicolumn{2}{c|}{\textbf{EST}} & 4 & 8 & 16 & 32 & Mean \\
\hline
\Tstrut\multirow{7}{*}{\makecell{\textbf{Test}\\\textbf{Shot}}} 
& 1 & 74.36 & 74.81 & 74.34 & 74.85 & 74.59 \\
& 2 & 82.80 & 83.23 & 82.61 & 82.12 & 82.69 \\
& 4 & 87.34 & 87.61 & 87.18 & 86.60 & 87.18 \\
& 8 & 89.44 & 89.68 & 89.34 & 88.90 & 89.34 \\
& 16 & 90.43 & 90.90 & 90.46 & 89.99 & 90.44 \\
& 32 & 90.76 & 91.40 & 90.98 & 90.70 & 90.96 \\
\cline{2-7}
\Tstrut & Offset & -0.01 & 0.40 & -0.05 & -0.34 & 0.84 \\
\bottomrule
\end{tabular}
\caption{Accuracy comparison for PCA and EST on CUB. Average group accuracy is displayed in the rightmost column of each table, model offsets are given in the bottom row. Bottom right cell (intersection of Offset and Mean) shows the sensitivity score.}
\label{tab:supp_tab_cub_est_12}
\end{table}

\clearpage

\begin{figure*}
\centering
\includegraphics[width=.8\linewidth]{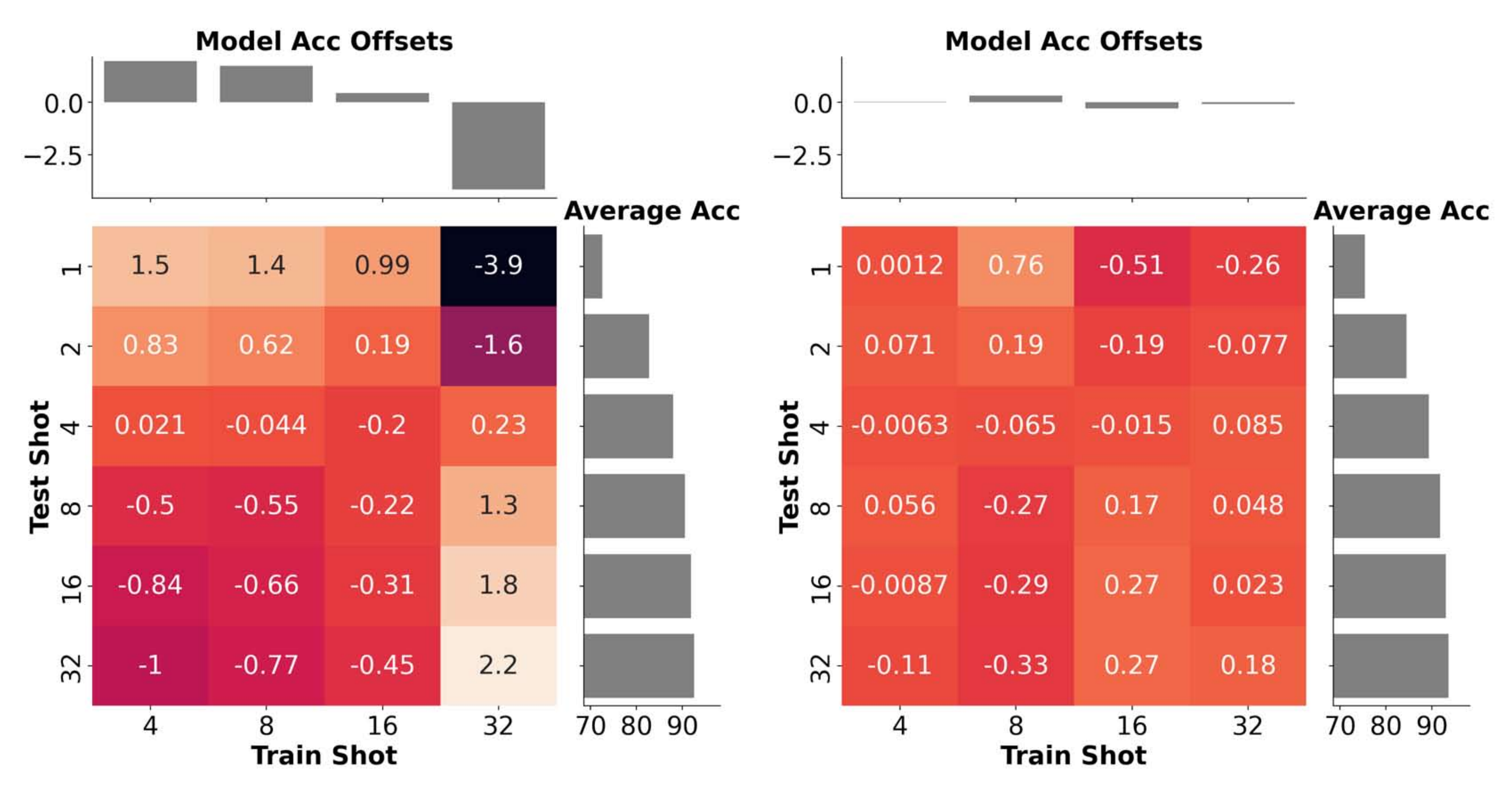}
\caption{FEAT vs Cosine FEAT on CUB, both using a ResNet-12 backbone.}
\label{fig:supp_heatmap_cub_feat_12}
\end{figure*}
\begin{table}
\centering
\tiny
\begin{tabular}{c c | c c c c | c}
\toprule
\multicolumn{2}{c|}{\textbf{CUB ResNet-12}} & \multicolumn{5}{c}{\textbf{Train Shot}} \\
\multicolumn{2}{c|}{\textbf{FEAT}} & 4 & 8 & 16 & 32 & Mean \\
\hline
\Tstrut\multirow{7}{*}{\makecell{\textbf{Test}\\\textbf{Shot}}} 
& 1 & 76.02 & 75.70 & 73.98 & 64.51 & 72.55 \\
& 2 & 85.51 & 85.08 & 83.34 & 76.91 & 82.71 \\
& 4 & 89.96 & 89.67 & 88.20 & 84.03 & 87.97 \\
& 8 & 92.05 & 91.77 & 90.79 & 87.69 & 90.58 \\
& 16 & 93.00 & 92.96 & 92.00 & 89.51 & 91.87 \\
& 32 & 93.51 & 93.52 & 92.53 & 90.59 & 92.54 \\
\cline{2-7}
\Tstrut & Offset & 1.97 & 1.75 & 0.44 & -4.16 & 6.10 \\
\bottomrule
\end{tabular}
\quad
\begin{tabular}{c c | c c c c | c}
\toprule
\multicolumn{2}{c|}{\textbf{CUB ResNet-12}} & \multicolumn{5}{c}{\textbf{Train Shot}} \\
\multicolumn{2}{c|}{\textbf{Cosine FEAT}} & 4 & 8 & 16 & 32 & Mean \\
\hline
\Tstrut\multirow{7}{*}{\makecell{\textbf{Test}\\\textbf{Shot}}} 
& 1 & 75.46 & 76.52 & 74.64 & 75.10 & 75.43 \\
& 2 & 84.53 & 84.95 & 83.96 & 84.28 & 84.43 \\
& 4 & 89.40 & 89.64 & 89.08 & 89.39 & 89.38 \\
& 8 & 91.93 & 91.90 & 91.73 & 91.82 & 91.85 \\
& 16 & 93.05 & 93.07 & 93.02 & 92.98 & 93.03 \\
& 32 & 93.55 & 93.63 & 93.62 & 93.74 & 93.63 \\
\cline{2-7}
\Tstrut & Offset & 0.03 & 0.33 & -0.28 & -0.07 & 1.27 \\
\bottomrule
\end{tabular}
\caption{Accuracy comparison for FEAT and Cosine FEAT on CUB. Average group accuracy is displayed in the rightmost column of each table, model offsets are given in the bottom row. Bottom right cell (intersection of Offset and Mean) shows the sensitivity score.}
\label{tab:supp_tab_cub_feat_12}
\end{table}

\begin{figure*}
\centering
\includegraphics[width=.8\linewidth]{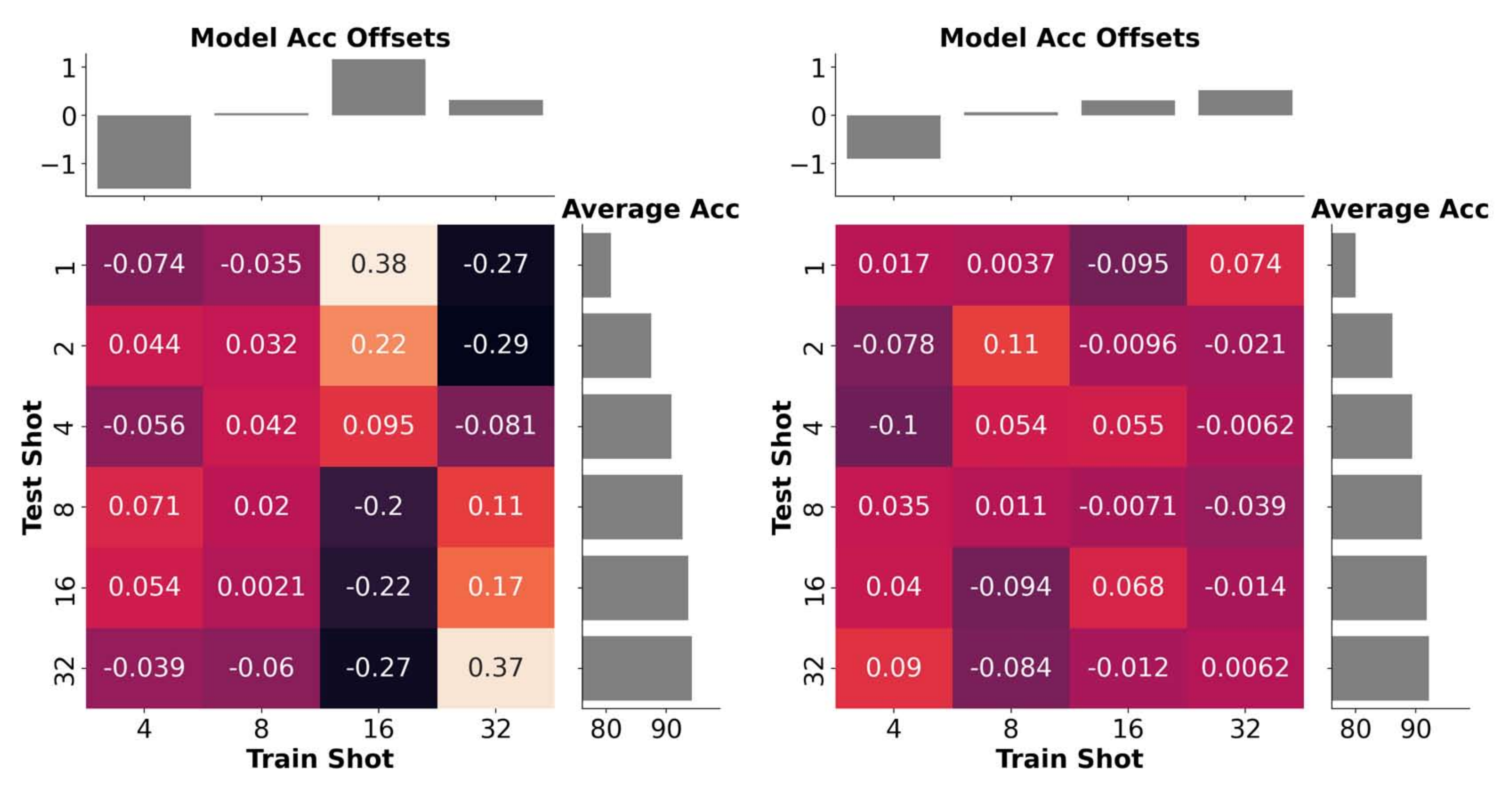}
\caption{FRN vs Cosine FRN on CUB, both using a ResNet-12 backbone.}
\label{fig:supp_heatmap_cub_frn_12}
\end{figure*}

\begin{table}
\centering
\tiny
\begin{tabular}{c c | c c c c | c}
\toprule
\multicolumn{2}{c|}{\textbf{CUB ResNet-12}} & \multicolumn{5}{c}{\textbf{Train Shot}} \\
\multicolumn{2}{c|}{\textbf{FRN}} & 4 & 8 & 16 & 32 & Mean \\
\hline
\Tstrut\multirow{7}{*}{\makecell{\textbf{Test}\\\textbf{Shot}}} 
& 1 & 79.18 & 80.79 & 82.32 & 80.83 & 80.78 \\
& 2 & 85.99 & 87.55 & 88.85 & 87.50 & 87.47 \\
& 4 & 89.28 & 90.95 & 92.12 & 91.10 & 90.86 \\
& 8 & 91.27 & 92.79 & 93.69 & 93.15 & 92.72 \\
& 16 & 92.20 & 93.72 & 94.61 & 94.16 & 93.67 \\
& 32 & 92.69 & 94.24 & 95.15 & 94.94 & 94.25 \\
\cline{2-7}
\Tstrut & Offset & -1.53 & 0.05 & 1.16 & 0.32 & 2.80 \\
\bottomrule
\end{tabular}
\quad
\begin{tabular}{c c | c c c c | c}
\toprule
\multicolumn{2}{c|}{\textbf{CUB ResNet-12}} & \multicolumn{5}{c}{\textbf{Train Shot}} \\
\multicolumn{2}{c|}{\textbf{Cosine FRN}} & 4 & 8 & 16 & 32 & Mean \\
\hline
\Tstrut\multirow{7}{*}{\makecell{\textbf{Test}\\\textbf{Shot}}} 
& 1 & 79.08 & 80.03 & 80.18 & 80.56 & 79.96 \\
& 2 & 85.14 & 86.29 & 86.42 & 86.62 & 86.12 \\
& 4 & 88.41 & 89.53 & 89.78 & 89.93 & 89.41 \\
& 8 & 90.17 & 91.11 & 91.34 & 91.52 & 91.03 \\
& 16 & 90.97 & 91.80 & 92.21 & 92.34 & 91.83 \\
& 32 & 91.37 & 92.16 & 92.48 & 92.71 & 92.18 \\
\cline{2-7}
\Tstrut & Offset & -0.90 & 0.06 & 0.31 & 0.52 & 0.21 \\
\bottomrule
\end{tabular}
\caption{Accuracy comparison for FRN and Cosine FRN on CUB. Average group accuracy is displayed in the rightmost column of each table, model offsets are given in the bottom row. Bottom right cell (intersection of Offset and Mean) shows the sensitivity score.}
\label{tab:supp_tab_cub_frn_12}
\end{table}

\clearpage

\begin{figure*}
\centering
\includegraphics[width=.8\linewidth]{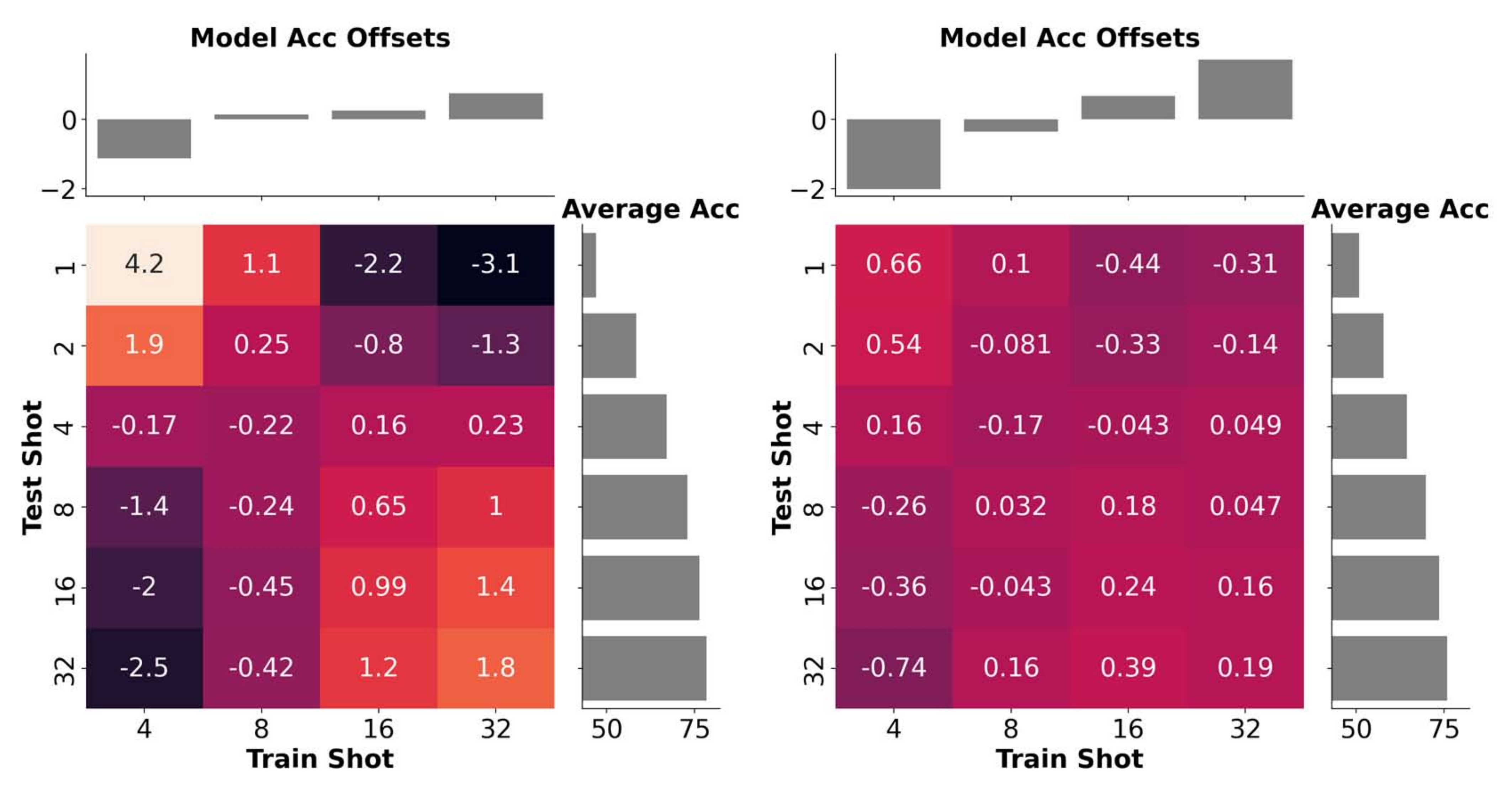}
\caption{Proto vs Cosine Proto on mini-ImageNet, both using a Conv-4 backbone.}
\label{fig:supp_heatmap_min_proto}
\end{figure*}

\begin{table}
\centering
\tiny
\begin{tabular}{c c | c c c c | c}
\toprule
\multicolumn{2}{c|}{\textbf{mIN Conv-4}} &\multicolumn{5}{c}{\textbf{Train Shot}} \\
\multicolumn{2}{c|}{\textbf{Proto}} & 4 & 8 & 16 & 32 & Mean \\
\hline
\Tstrut\multirow{7}{*}{\makecell{\textbf{Test}\\\textbf{Shot}}} 
& 1 & 50.08 & 48.20 & 45.09 & 44.62 & 47.00 \\
& 2 & 59.08 & 58.72 & 57.79 & 57.76 & 58.34 \\
& 4 & 65.78 & 66.98 & 67.49 & 68.04 & 67.07 \\
& 8 & 70.45 & 72.87 & 73.88 & 74.72 & 72.98 \\
& 16 & 73.27 & 76.07 & 77.63 & 78.57 & 76.39 \\
& 32 & 74.70 & 78.04 & 79.74 & 80.83 & 78.33 \\
\cline{2-7}
\Tstrut & Offset & -1.12 & 0.13 & 0.25 & 0.74 & 7.32 \\
\bottomrule
\end{tabular}
\quad
\begin{tabular}{c c | c c c c | c}
\toprule
\multicolumn{2}{c|}{\textbf{mIN Conv-4}}& \multicolumn{5}{c}{\textbf{Train Shot}} \\
\multicolumn{2}{c|}{\textbf{Cosine Proto}} & 4 & 8 & 16 & 32 & Mean \\
\hline
\Tstrut\multirow{7}{*}{\makecell{\textbf{Test}\\\textbf{Shot}}} 
& 1 & 49.49 & 50.59 & 51.08 & 52.25 & 50.85 \\
& 2 & 56.32 & 57.35 & 58.14 & 59.37 & 57.79 \\
& 4 & 62.57 & 63.90 & 65.06 & 66.19 & 64.43 \\
& 8 & 67.58 & 69.53 & 70.72 & 71.62 & 69.86 \\
& 16 & 71.23 & 73.20 & 74.52 & 75.48 & 73.61 \\
& 32 & 73.12 & 75.67 & 76.94 & 77.78 & 75.88 \\
\cline{2-7}
\Tstrut & Offset & -2.02 & -0.36 & 0.67 & 1.71 & 1.39 \\
\bottomrule
\end{tabular}
\caption{Accuracy comparison for ProtoNet / Cosine ProtoNet on mini-ImageNet. Average group accuracy is displayed in the rightmost column of each table, model offsets are given in the bottom row. Bottom right cell (intersection of Offset and Mean) shows the sensitivity score.}
\label{tab:supp_tab_min_proto}
\end{table}

    \begin{figure*}
    \centering
    \includegraphics[width=.8\linewidth]{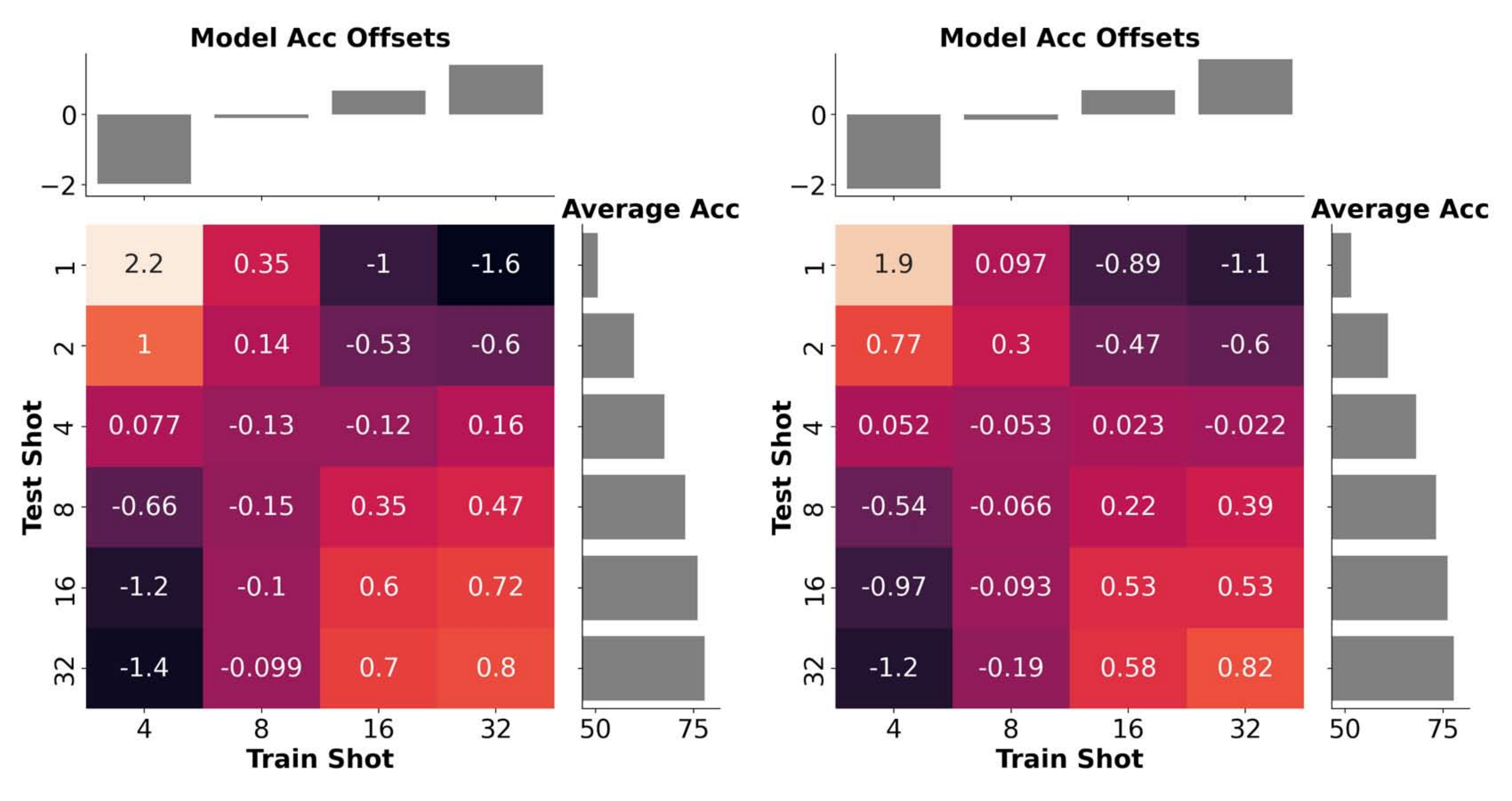}
    \caption{PCA vs EST on mini-ImageNet, both using a Conv-4 backbone.}
    \label{fig:supp_heatmap_min_est}
    \end{figure*}
    
    \begin{table}
    \centering
    \tiny
    \begin{tabular}{c c | c c c c | c}
    \toprule
    \multicolumn{2}{c|}{\textbf{mIN Conv-4}}& \multicolumn{5}{c}{\textbf{Train Shot}} \\
    \multicolumn{2}{c|}{\textbf{PCA}} & 4 & 8 & 16 & 32 & Mean \\
    \hline
    \Tstrut\multirow{7}{*}{\makecell{\textbf{Test}\\\textbf{Shot}}} 
& 1 & 50.82 & 50.84 & 50.27 & 50.45 & 50.60 \\
& 2 & 58.87 & 59.89 & 60.00 & 60.66 & 59.85 \\
& 4 & 65.63 & 67.31 & 68.10 & 69.11 & 67.54 \\
& 8 & 70.28 & 72.67 & 73.95 & 74.80 & 72.93 \\
& 16 & 72.82 & 75.82 & 77.30 & 78.15 & 76.02 \\
& 32 & 74.46 & 77.65 & 79.23 & 80.06 & 77.85 \\
\cline{2-7}
\Tstrut & Offset & -1.98 & -0.10 & 0.68 & 1.41 & 3.76 \\
    \bottomrule
    \end{tabular}
    \quad
    \begin{tabular}{c c | c c c c | c}
    \toprule
    \multicolumn{2}{c|}{\textbf{mIN Conv-4}}& \multicolumn{5}{c}{\textbf{Train Shot}} \\
    \multicolumn{2}{c|}{\textbf{EST}} & 4 & 8 & 16 & 32 & Mean \\
    \hline
    \Tstrut\multirow{7}{*}{\makecell{\textbf{Test}\\\textbf{Shot}}} 
& 1 & 51.39 & 51.55 & 51.42 & 52.07 & 51.61 \\
& 2 & 59.54 & 61.04 & 61.12 & 61.87 & 60.89 \\
& 4 & 66.13 & 67.99 & 68.92 & 69.75 & 68.20 \\
& 8 & 70.51 & 72.95 & 74.09 & 75.13 & 73.17 \\
& 16 & 73.09 & 75.93 & 77.41 & 78.28 & 76.18 \\
& 32 & 74.44 & 77.43 & 79.05 & 80.16 & 77.77 \\
\cline{2-7}
\Tstrut & Offset & -2.12 & -0.15 & 0.70 & 1.57 & 3.11 \\
    \bottomrule
    \end{tabular}
    \caption{Accuracy comparison for PCA and EST on mini-ImageNet. Average group accuracy is displayed in the rightmost column of each table, model offsets are given in the bottom row. Bottom right cell (intersection of Offset and Mean) shows the sensitivity score.}
    \label{tab:supp_tab_min_est}
    \end{table}
    
    \clearpage
    
    \begin{figure*}
    \centering
    \includegraphics[width=.8\linewidth]{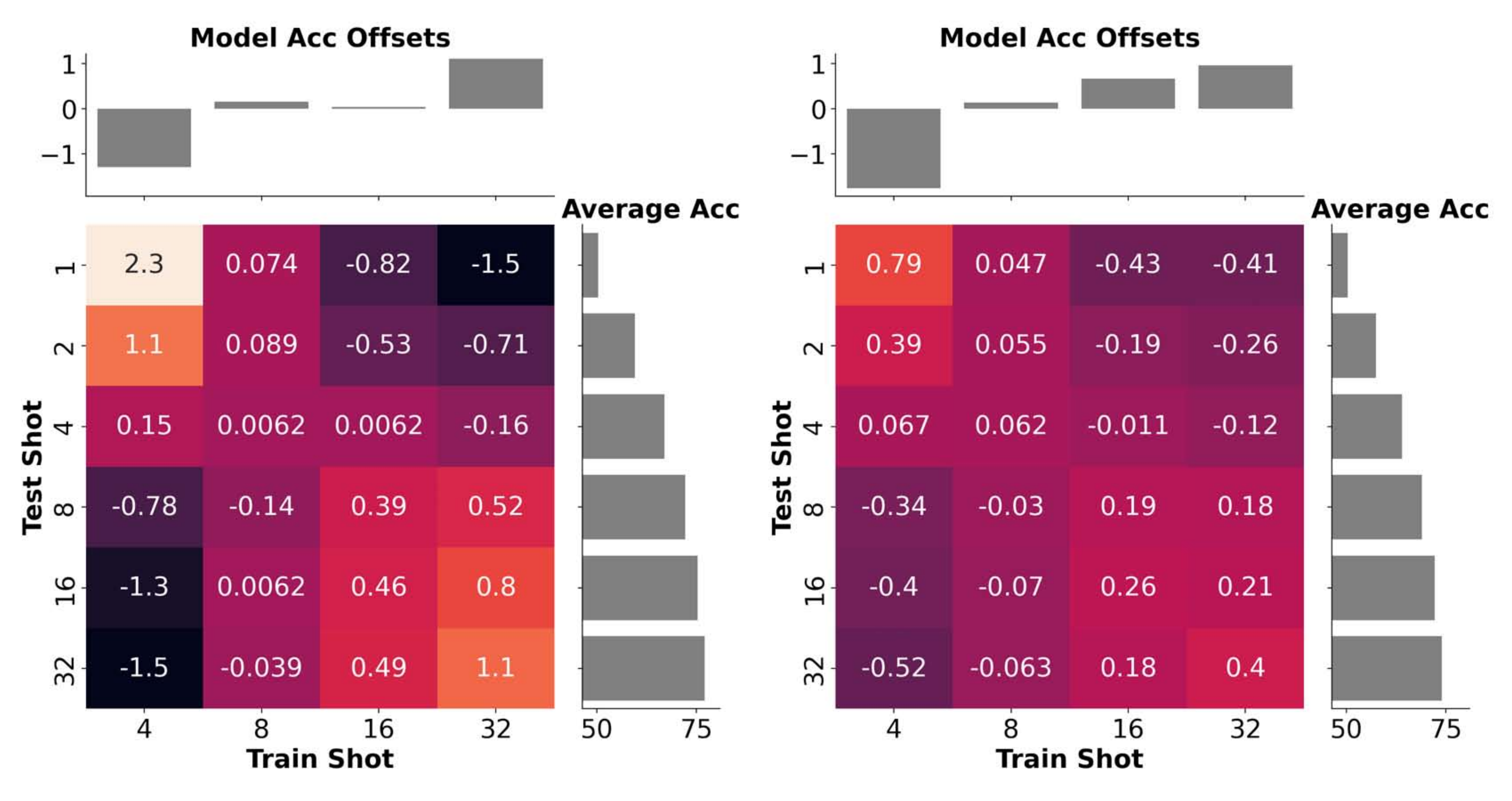}
    \caption{FEAT vs Cosine FEAT on mini-ImageNet, both using a Conv-4 backbone.}
    \label{fig:supp_heatmap_min_feat}
    \end{figure*}
    \begin{table}
    \centering
    \tiny
    \begin{tabular}{c c | c c c c | c}
    \toprule
    \multicolumn{2}{c|}{\textbf{mIN Conv-4}}&\multicolumn{5}{c}{\textbf{Train Shot}} \\
    \multicolumn{2}{c|}{\textbf{FEAT}} & 4 & 8 & 16 & 32 & Mean \\
    \hline
    \Tstrut\multirow{7}{*}{\makecell{\textbf{Test}\\\textbf{Shot}}} 
& 1 & 51.32 & 50.59 & 49.59 & 49.96 & 50.36 \\
& 2 & 59.43 & 59.82 & 59.09 & 59.98 & 59.58 \\
& 4 & 65.87 & 67.18 & 67.07 & 67.97 & 67.02 \\
& 8 & 70.17 & 72.26 & 72.68 & 73.87 & 72.24 \\
& 16 & 72.76 & 75.48 & 75.82 & 77.23 & 75.32 \\
& 32 & 74.30 & 77.22 & 77.64 & 79.27 & 77.11 \\
\cline{2-7}
\Tstrut & Offset & -1.30 & 0.15 & 0.04 & 1.11 & 3.77 \\
    \bottomrule
    \end{tabular}
    \quad
    \begin{tabular}{c c | c c c c | c}
    \toprule
    \multicolumn{2}{c|}{\textbf{mIN Conv-4}}&\multicolumn{5}{c}{\textbf{Train Shot}} \\
    \multicolumn{2}{c|}{\textbf{Cosine FEAT}} & 4 & 8 & 16 & 32 & Mean \\
    \hline
    \Tstrut\multirow{7}{*}{\makecell{\textbf{Test}\\\textbf{Shot}}} 
& 1 & 49.36 & 50.52 & 50.57 & 50.88 & 50.33 \\
& 2 & 56.14 & 57.71 & 57.99 & 58.22 & 57.51 \\
& 4 & 62.35 & 64.25 & 64.70 & 64.89 & 64.05 \\
& 8 & 66.94 & 69.15 & 69.89 & 70.18 & 69.04 \\
& 16 & 70.04 & 72.27 & 73.12 & 73.37 & 72.20 \\
& 32 & 71.77 & 74.13 & 74.90 & 75.41 & 74.05 \\
\cline{2-7}
\Tstrut & Offset & -1.76 & 0.14 & 0.66 & 0.96 & 1.31 \\
    \bottomrule
    \end{tabular}
    \caption{Accuracy comparison for FEAT and Cosine FEAT on mini-ImageNet. Average group accuracy is displayed in the rightmost column of each table, model offsets are given in the bottom row. Bottom right cell (intersection of Offset and Mean) shows the sensitivity score.}
    \label{tab:supp_tab_min_feat}
    \end{table}

    \begin{figure*}
    \centering
    \includegraphics[width=.8\linewidth]{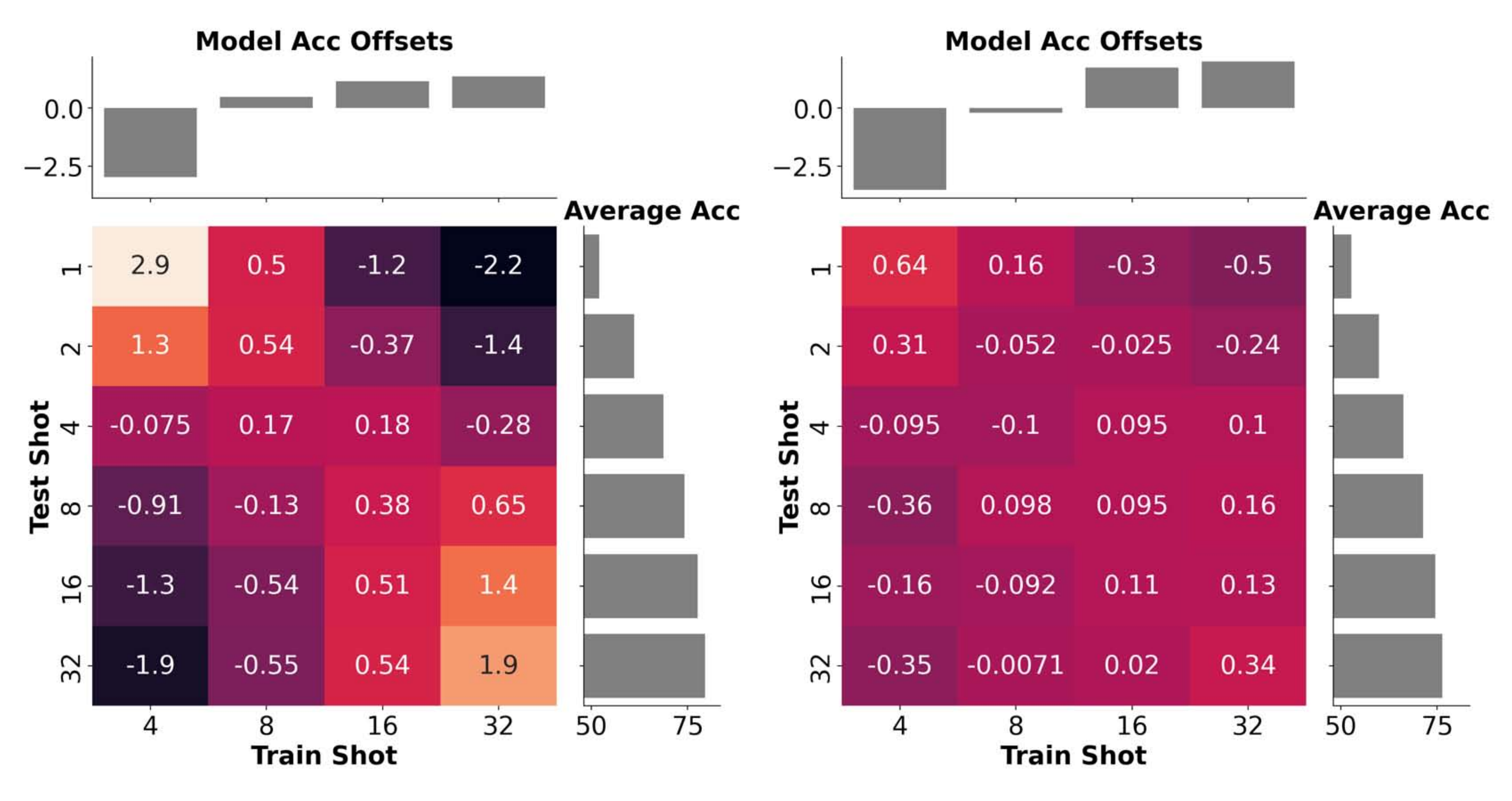}
    \caption{FRN vs Cosine FRN on mini-ImageNet, both using a Conv-4 backbone.}
    \label{fig:supp_heatmap_min_frn}
    \end{figure*}
    \begin{table}
    \centering
    \tiny
    \begin{tabular}{c c | c c c c | c}
    \toprule
    \multicolumn{2}{c|}{\textbf{mIN Conv-4}}&\multicolumn{5}{c}{\textbf{Train Shot}} \\
    \multicolumn{2}{c|}{\textbf{FRN}} & 4 & 8 & 16 & 32 & Mean \\
    \hline
    \Tstrut\multirow{7}{*}{\makecell{\textbf{Test}\\\textbf{Shot}}} 
& 1 & 52.05 & 53.08 & 52.01 & 51.29 & 52.11 \\
& 2 & 59.46 & 62.17 & 61.94 & 61.06 & 61.16 \\
& 4 & 65.70 & 69.39 & 70.08 & 69.82 & 68.75 \\
& 8 & 70.31 & 74.54 & 75.73 & 76.20 & 74.20 \\
& 16 & 73.26 & 77.50 & 79.23 & 80.29 & 77.57 \\
& 32 & 74.69 & 79.45 & 81.22 & 82.76 & 79.53 \\
\cline{2-7}
\Tstrut & Offset & -2.97 & 0.47 & 1.15 & 1.35 & 5.09 \\
    \bottomrule
    \end{tabular}
    \quad
    \begin{tabular}{c c | c c c c | c}
    \toprule
    \multicolumn{2}{c|}{\textbf{mIN Conv-4}}&\multicolumn{5}{c}{\textbf{Train Shot}} \\
    \multicolumn{2}{c|}{\textbf{Cosine FRN}} & 4 & 8 & 16 & 32 & Mean \\
    \hline
    \Tstrut\multirow{7}{*}{\makecell{\textbf{Test}\\\textbf{Shot}}} 
& 1 & 49.91 & 52.75 & 54.23 & 54.29 & 52.79 \\
& 2 & 56.66 & 59.62 & 61.58 & 61.63 & 59.87 \\
& 4 & 62.63 & 65.95 & 68.08 & 68.35 & 66.25 \\
& 8 & 67.43 & 71.21 & 73.14 & 73.47 & 71.31 \\
& 16 & 70.82 & 74.21 & 76.35 & 76.63 & 74.50 \\
& 32 & 72.49 & 76.16 & 78.12 & 78.70 & 76.37 \\
\cline{2-7}
\Tstrut & Offset & -3.53 & -0.20 & 1.73 & 1.99 & 1.14 \\
    \bottomrule
    \end{tabular}
    \caption{Accuracy comparison for FRN and Cosine FRN on mini-ImageNet. Average group accuracy is displayed in the rightmost column of each table, model offsets are given in the bottom row. Bottom right cell (intersection of Offset and Mean) shows the sensitivity score.}
    \label{tab:supp_tab_min_frn}
    \end{table}
    
    \clearpage
    
    \begin{figure*}
    \centering
    \includegraphics[width=.8\linewidth]{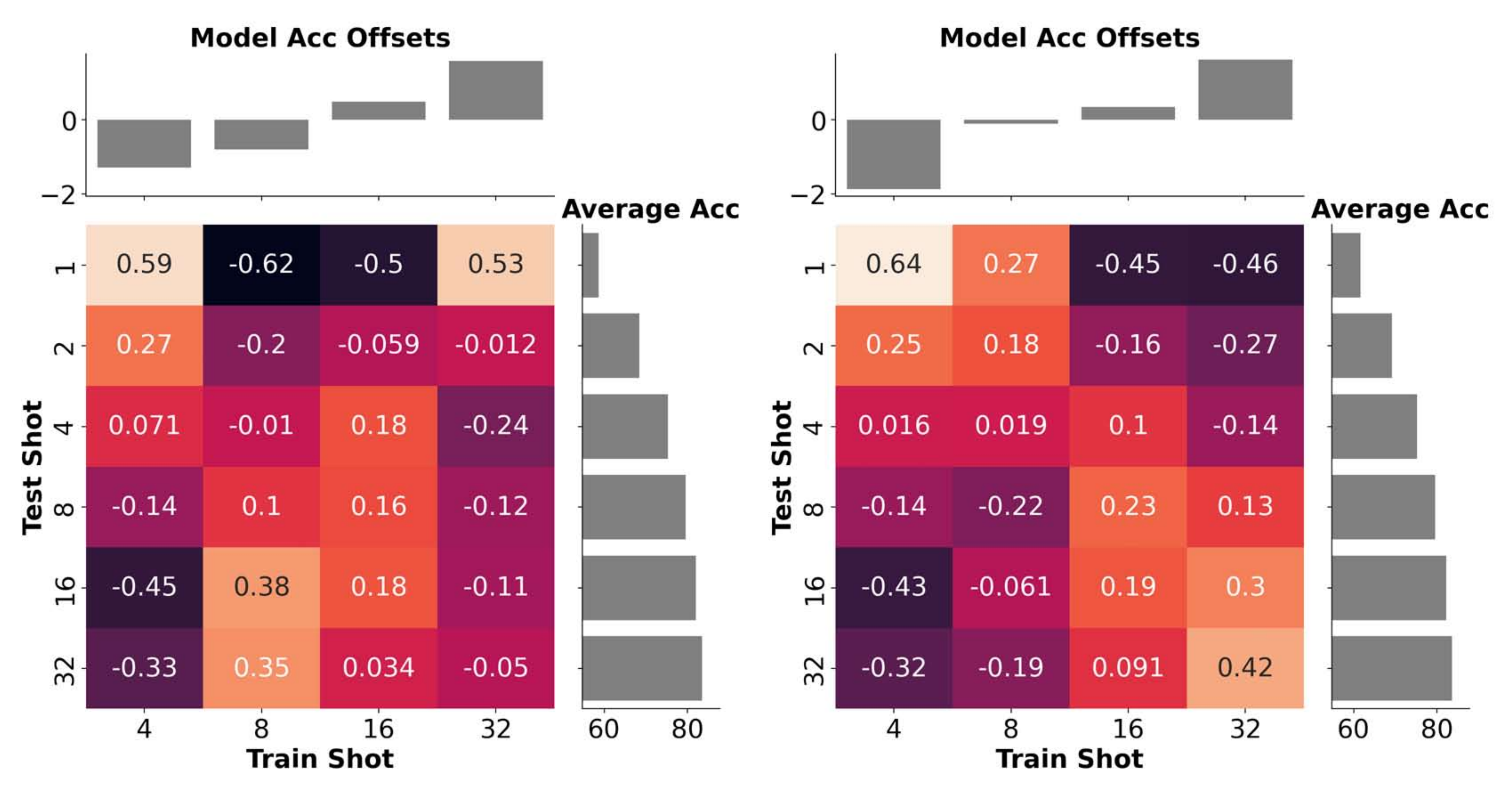}
    \caption{ProtoNet vs Cosine ProtoNet on mini-ImageNet, both using a ResNet-12 backbone.}
    \label{fig:supp_heatmap_min_proto_12}
    \end{figure*}
    \begin{table}
    \centering
    \tiny
    \begin{tabular}{c c | c c c c | c}
    \toprule
    \multicolumn{2}{c|}{\textbf{mIN ResNet-12}} &\multicolumn{5}{c}{\textbf{Train Shot}} \\
    \multicolumn{2}{c|}{\textbf{Proto}} & 4 & 8 & 16 & 32 & Mean \\
    \hline
    \Tstrut\multirow{7}{*}{\makecell{\textbf{Test}\\\textbf{Shot}}} 
& 1 & 57.99 & 57.27 & 58.68 & 60.80 & 58.69 \\
& 2 & 67.47 & 67.50 & 68.92 & 70.06 & 68.49 \\
& 4 & 74.09 & 74.50 & 75.97 & 76.65 & 75.30 \\
& 8 & 78.21 & 78.95 & 80.29 & 81.10 & 79.64 \\
& 16 & 80.35 & 81.67 & 82.76 & 83.56 & 82.08 \\
& 32 & 81.86 & 83.03 & 84.00 & 85.01 & 83.47 \\
\cline{2-7}
\Tstrut & Offset & -1.28 & -0.79 & 0.49 & 1.58 & 1.21 \\
    \bottomrule
    \end{tabular}
    \quad
    \begin{tabular}{c c | c c c c | c}
    \toprule
    \multicolumn{2}{c|}{\textbf{mIN ResNet-12}}& \multicolumn{5}{c}{\textbf{Train Shot}} \\
    \multicolumn{2}{c|}{\textbf{Cosine Proto}} & 4 & 8 & 16 & 32 & Mean \\
    \hline
    \Tstrut\multirow{7}{*}{\makecell{\textbf{Test}\\\textbf{Shot}}} 
& 1 & 60.41 & 61.81 & 61.55 & 62.81 & 61.65 \\
& 2 & 67.51 & 69.21 & 69.33 & 70.49 & 69.14 \\
& 4 & 73.36 & 75.13 & 75.67 & 76.70 & 75.21 \\
& 8 & 77.59 & 79.28 & 80.18 & 81.35 & 79.60 \\
& 16 & 79.89 & 82.03 & 82.74 & 84.12 & 82.19 \\
& 32 & 81.47 & 83.37 & 84.11 & 85.71 & 83.67 \\
\cline{2-7}
\Tstrut & Offset & -1.87 & -0.10 & 0.35 & 1.62 & 1.09 \\
    \bottomrule
    \end{tabular}
    \caption{Accuracy comparison for ProtoNet / Cosine ProtoNet on mini-ImageNet. Average group accuracy is displayed in the rightmost column of each table, model offsets are given in the bottom row. Bottom right cell (intersection of Offset and Mean) shows the sensitivity score.}
    \label{tab:supp_tab_min_proto_12}
    \end{table}

    \begin{figure*}
    \centering
    \includegraphics[width=.8\linewidth]{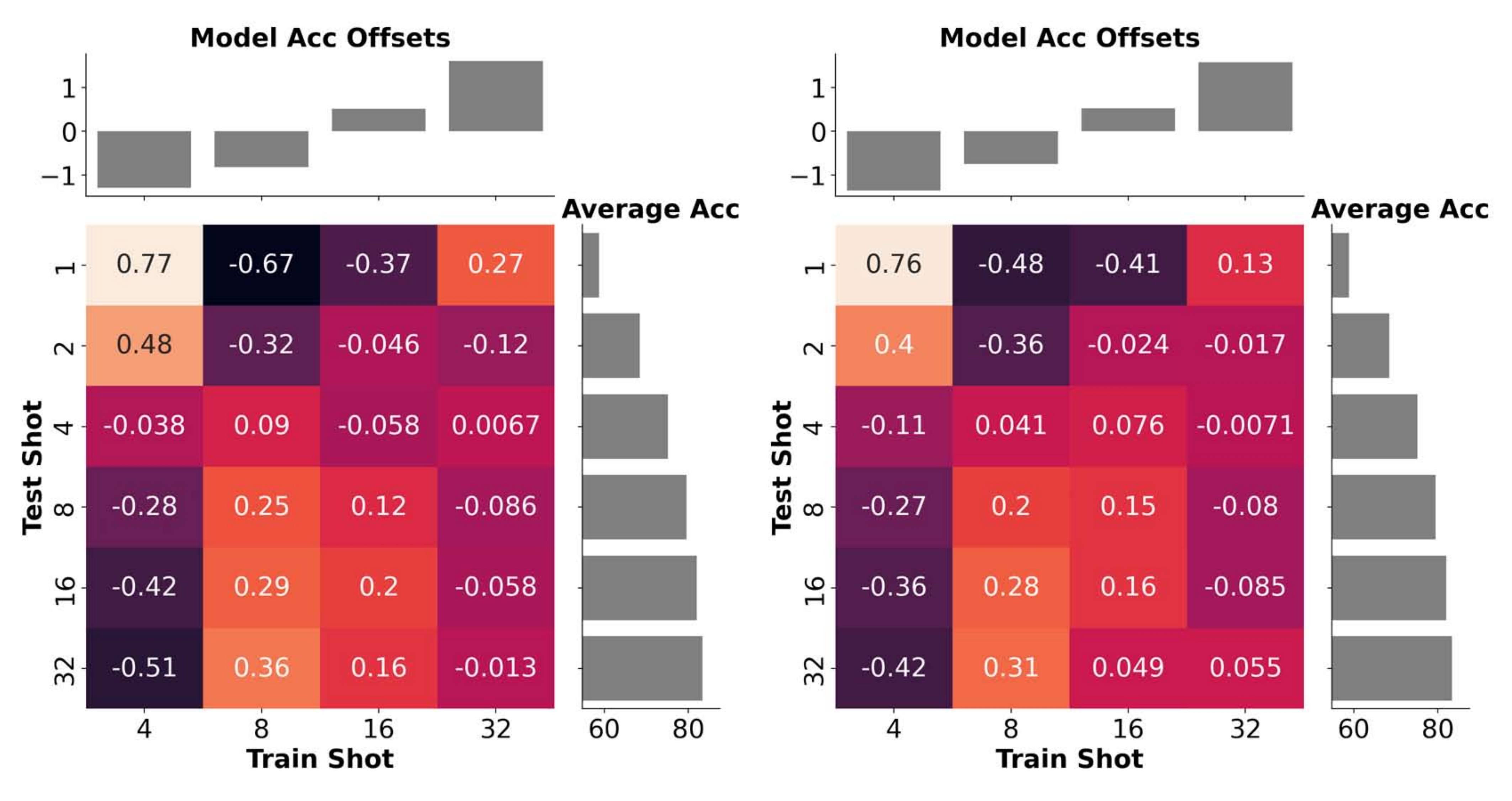}
    \caption{PCA vs EST on mini-ImageNet, both using a ResNet-12 backbone.}
    \label{fig:supp_heatmap_min_est_12}
    \end{figure*}
    \begin{table}
    \centering
    \tiny
    \begin{tabular}{c c | c c c c | c}
    \toprule
    \multicolumn{2}{c|}{\textbf{mIN ResNet-12}}&\multicolumn{5}{c}{\textbf{Train Shot}} \\
    \multicolumn{2}{c|}{\textbf{PCA}} & 4 & 8 & 16 & 32 & Mean \\
    \hline
    \Tstrut\multirow{7}{*}{\makecell{\textbf{Test}\\\textbf{Shot}}} 
& 1 & 58.22 & 57.26 & 58.88 & 60.62 & 58.74 \\
& 2 & 67.60 & 67.27 & 68.87 & 69.89 & 68.41 \\
& 4 & 73.84 & 74.44 & 75.62 & 76.78 & 75.17 \\
& 8 & 77.97 & 78.97 & 80.17 & 81.06 & 79.54 \\
& 16 & 80.25 & 81.43 & 82.67 & 83.51 & 81.96 \\
& 32 & 81.52 & 82.86 & 83.99 & 84.91 & 83.32 \\
\cline{2-7}
\Tstrut & Offset & -1.29 & -0.82 & 0.51 & 1.60 & 1.43 \\
    \bottomrule
    \end{tabular}
    \quad
    \begin{tabular}{c c | c c c c | c}
    \toprule
    \multicolumn{2}{c|}{\textbf{mIN ResNet-12}}&\multicolumn{5}{c}{\textbf{Train Shot}} \\
    \multicolumn{2}{c|}{\textbf{EST}} & 4 & 8 & 16 & 32 & Mean \\
    \hline
    \Tstrut\multirow{7}{*}{\makecell{\textbf{Test}\\\textbf{Shot}}} 
& 1 & 58.30 & 57.66 & 59.00 & 60.59 & 58.89 \\
& 2 & 67.52 & 67.36 & 68.97 & 70.02 & 68.47 \\
& 4 & 73.71 & 74.46 & 75.77 & 76.73 & 75.17 \\
& 8 & 77.85 & 78.92 & 80.15 & 80.96 & 79.47 \\
& 16 & 80.27 & 81.51 & 82.66 & 83.46 & 81.97 \\
& 32 & 81.55 & 82.88 & 83.89 & 84.94 & 83.32 \\
\cline{2-7}
\Tstrut & Offset & -1.35 & -0.75 & 0.53 & 1.57 & 1.24 \\
    \bottomrule
    \end{tabular}
    \caption{Accuracy comparison for PCA and EST on mini-ImageNet. Average group accuracy is displayed in the rightmost column of each table, model offsets are given in the bottom row. Bottom right cell (intersection of Offset and Mean) shows the sensitivity score.}
    \label{tab:supp_tab_min_est_12}
    \end{table}
    
    \clearpage
    
    \begin{figure*}
    \centering
    \includegraphics[width=.8\linewidth]{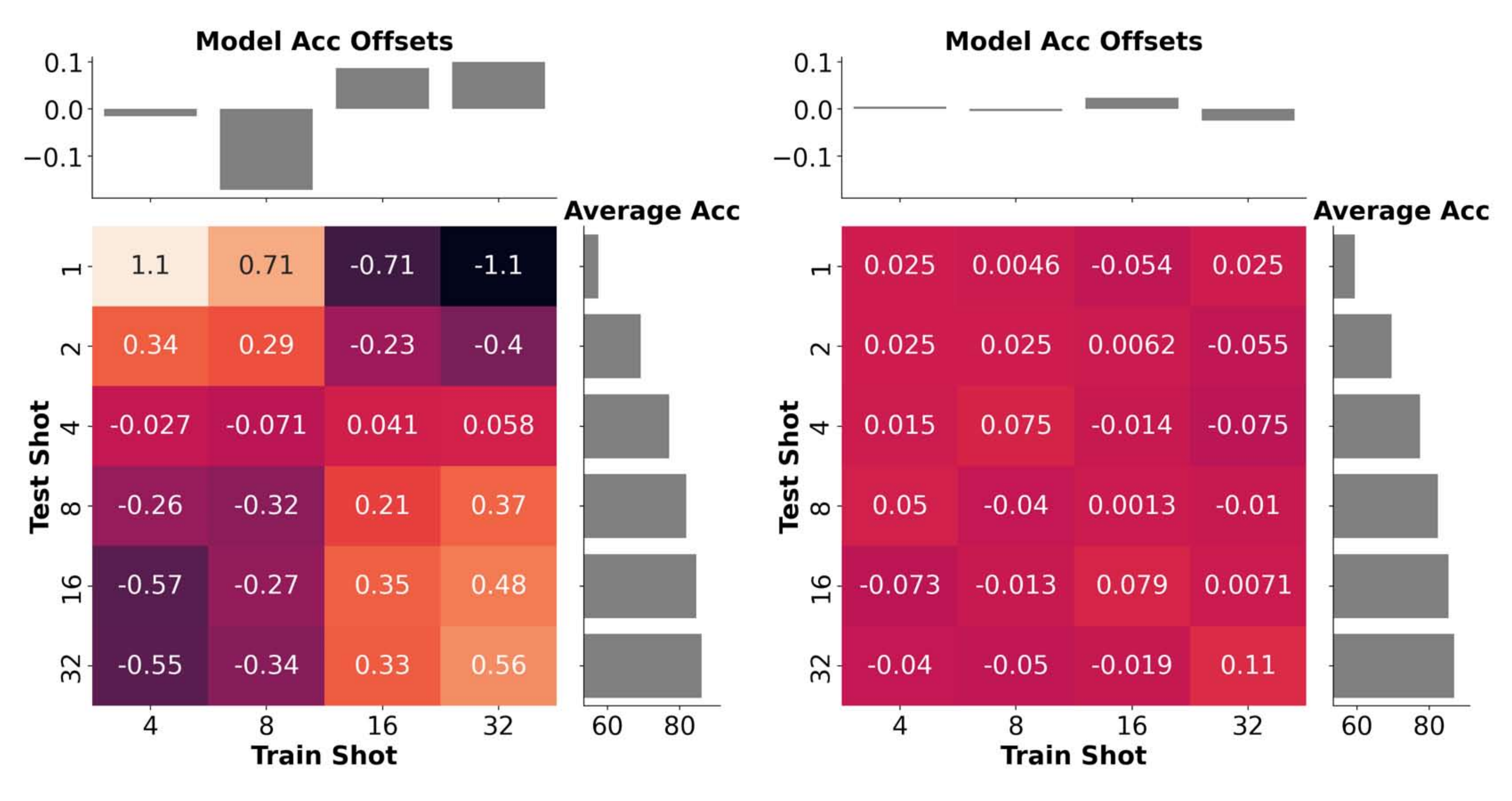}
    \caption{FEAT vs Cosine FEAT on mini-ImageNet, both using a ResNet-12 backbone.}
    \label{fig:supp_heatmap_min_feat_12}
    \end{figure*}
    \begin{table}
    \centering
    \tiny
    \begin{tabular}{c c | c c c c | c}
    \toprule
    \multicolumn{2}{c|}{\textbf{mIN ResNet-12}}&\multicolumn{5}{c}{\textbf{Train Shot}} \\
    \multicolumn{2}{c|}{\textbf{FEAT}} & 4 & 8 & 16 & 32 & Mean \\
    \hline
    \Tstrut\multirow{7}{*}{\makecell{\textbf{Test}\\\textbf{Shot}}} 
& 1 & 58.55 & 58.04 & 56.88 & 56.54 & 57.50 \\
& 2 & 69.47 & 69.26 & 69.00 & 68.84 & 69.14 \\
& 4 & 77.04 & 76.84 & 77.21 & 77.24 & 77.08 \\
& 8 & 81.52 & 81.30 & 82.09 & 82.26 & 81.79 \\
& 16 & 83.97 & 84.11 & 84.99 & 85.13 & 84.55 \\
& 32 & 85.40 & 85.46 & 86.39 & 86.63 & 85.97 \\
\cline{2-7}
\Tstrut & Offset & -0.02 & -0.17 & 0.09 & 0.10 & 2.13 \\
    \bottomrule
    \end{tabular}
    \quad
    \begin{tabular}{c c | c c c c | c}
    \toprule
    \multicolumn{2}{c|}{\textbf{mIN ResNet-12}}&\multicolumn{5}{c}{\textbf{Train Shot}} \\
    \multicolumn{2}{c|}{\textbf{Cosine FEAT}} & 4 & 8 & 16 & 32 & Mean \\
    \hline
    \Tstrut\multirow{7}{*}{\makecell{\textbf{Test}\\\textbf{Shot}}} 
& 1 & 59.42 & 59.39 & 59.36 & 59.39 & 59.39 \\
& 2 & 69.65 & 69.64 & 69.65 & 69.54 & 69.62 \\
& 4 & 77.47 & 77.52 & 77.46 & 77.35 & 77.45 \\
& 8 & 82.42 & 82.32 & 82.39 & 82.33 & 82.36 \\
& 16 & 85.22 & 85.27 & 85.39 & 85.27 & 85.29 \\
& 32 & 86.83 & 86.81 & 86.87 & 86.95 & 86.87 \\
\cline{2-7}
\Tstrut & Offset & 0.01 & -0.00 & 0.02 & -0.02 & 0.19 \\
    \bottomrule
    \end{tabular}
    \caption{Accuracy comparison for FEAT and Cosine FEAT on mini-ImageNet. Average group accuracy is displayed in the rightmost column of each table, model offsets are given in the bottom row. Bottom right cell (intersection of Offset and Mean) shows the sensitivity score.}
    \label{tab:supp_tab_min_feat_12}
    \end{table}

    \begin{figure*}
    \centering
    \includegraphics[width=.8\linewidth]{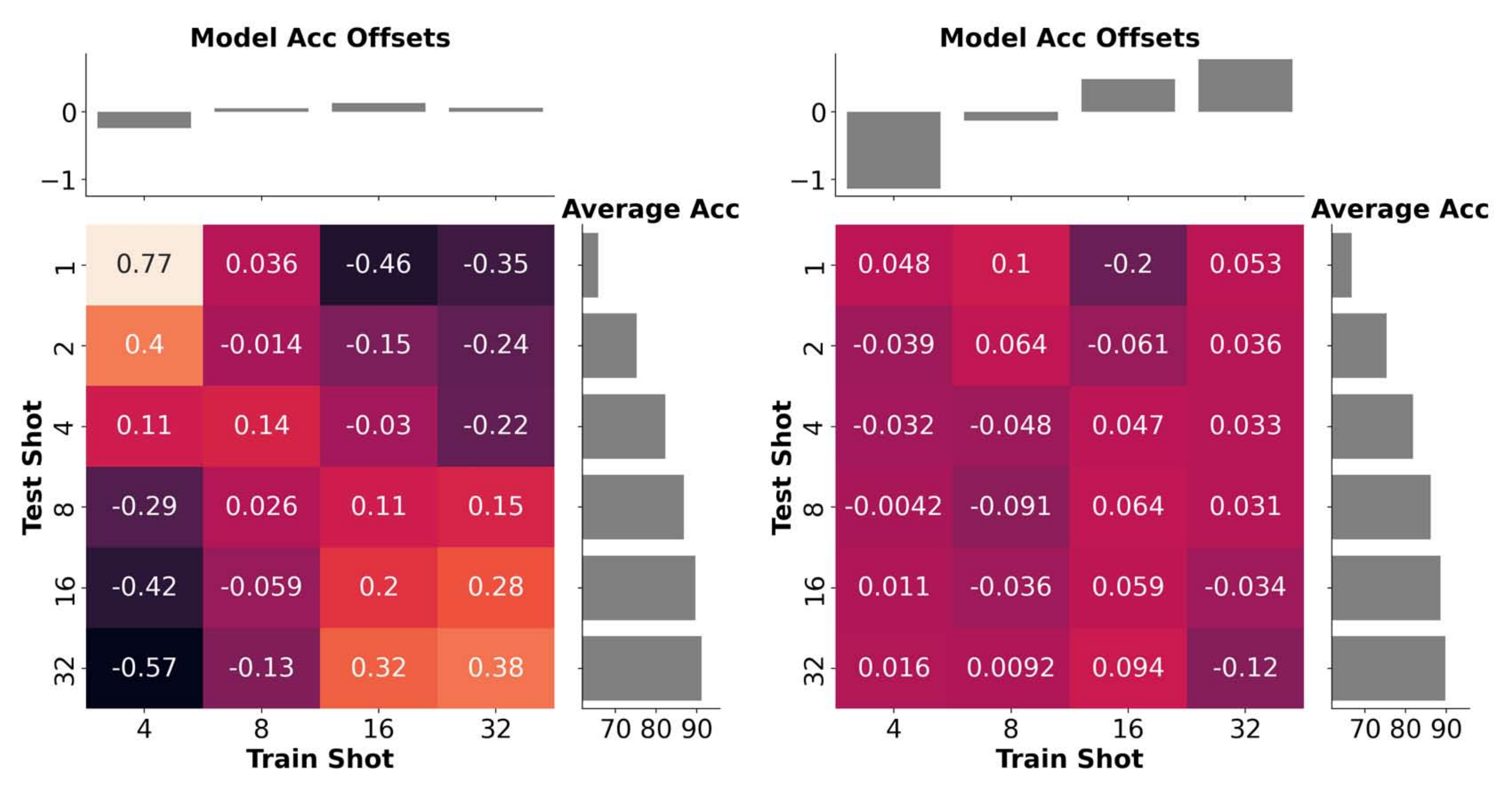}
    \caption{FRN vs Cosine FRN on mini-ImageNet, both using a ResNet-12 backbone.}
    \label{fig:supp_heatmap_min_frn_12}
    \end{figure*}
    \begin{table}
    \centering
    \tiny
    \begin{tabular}{c c | c c c c | c}
    \toprule
\multicolumn{2}{c|}{\textbf{mIN ResNet-12}}&\multicolumn{5}{c}{\textbf{Train Shot}} \\
    \multicolumn{2}{c|}{\textbf{FRN}} & 4 & 8 & 16 & 32 & Mean \\
    \hline
    \Tstrut\multirow{7}{*}{\makecell{\textbf{Test}\\\textbf{Shot}}} 
& 1 & 66.35 & 65.91 & 65.49 & 65.53 & 65.82 \\
& 2 & 75.34 & 75.22 & 75.16 & 75.00 & 75.18 \\
& 4 & 82.17 & 82.50 & 82.40 & 82.14 & 82.30 \\
& 8 & 86.32 & 86.93 & 87.09 & 87.06 & 86.85 \\
& 16 & 88.95 & 89.61 & 89.94 & 89.96 & 89.61 \\
& 32 & 90.30 & 91.04 & 91.57 & 91.56 & 91.12 \\
\cline{2-7}
\Tstrut & Offset & -0.24 & 0.05 & 0.13 & 0.06 & 1.35 \\
    \bottomrule
    \end{tabular}
    \quad
    \begin{tabular}{c c | c c c c | c}
    \toprule
    \multicolumn{2}{c|}{\textbf{mIN ResNet-12}}&\multicolumn{5}{c}{\textbf{Train Shot}} \\
    \multicolumn{2}{c|}{\textbf{Cosine FRN}} & 4 & 8 & 16 & 32 & Mean \\
    \hline
    \Tstrut\multirow{7}{*}{\makecell{\textbf{Test}\\\textbf{Shot}}} 
& 1 & 65.64 & 66.70 & 67.01 & 67.56 & 66.73 \\
& 2 & 74.11 & 75.22 & 75.71 & 76.10 & 75.29 \\
& 4 & 80.65 & 81.64 & 82.35 & 82.63 & 81.82 \\
& 8 & 85.05 & 85.97 & 86.74 & 87.00 & 86.19 \\
& 16 & 87.40 & 88.36 & 89.07 & 89.27 & 88.53 \\
& 32 & 88.63 & 89.63 & 90.33 & 90.41 & 89.75 \\
\cline{2-7}
\Tstrut & Offset & -1.14 & -0.13 & 0.49 & 0.78 & 0.30 \\
    \bottomrule
    \end{tabular}
    \caption{Accuracy comparison for FRN and Cosine FRN on mini-ImageNet. Average group accuracy is displayed in the rightmost column of each table, model offsets are given in the bottom row. Bottom right cell (intersection of Offset and Mean) shows the sensitivity score.}
    \label{tab:supp_tab_min_frn_12}
    \end{table}
    
    \clearpage

        \begin{figure*}
        \centering
        \includegraphics[width=.8\linewidth]{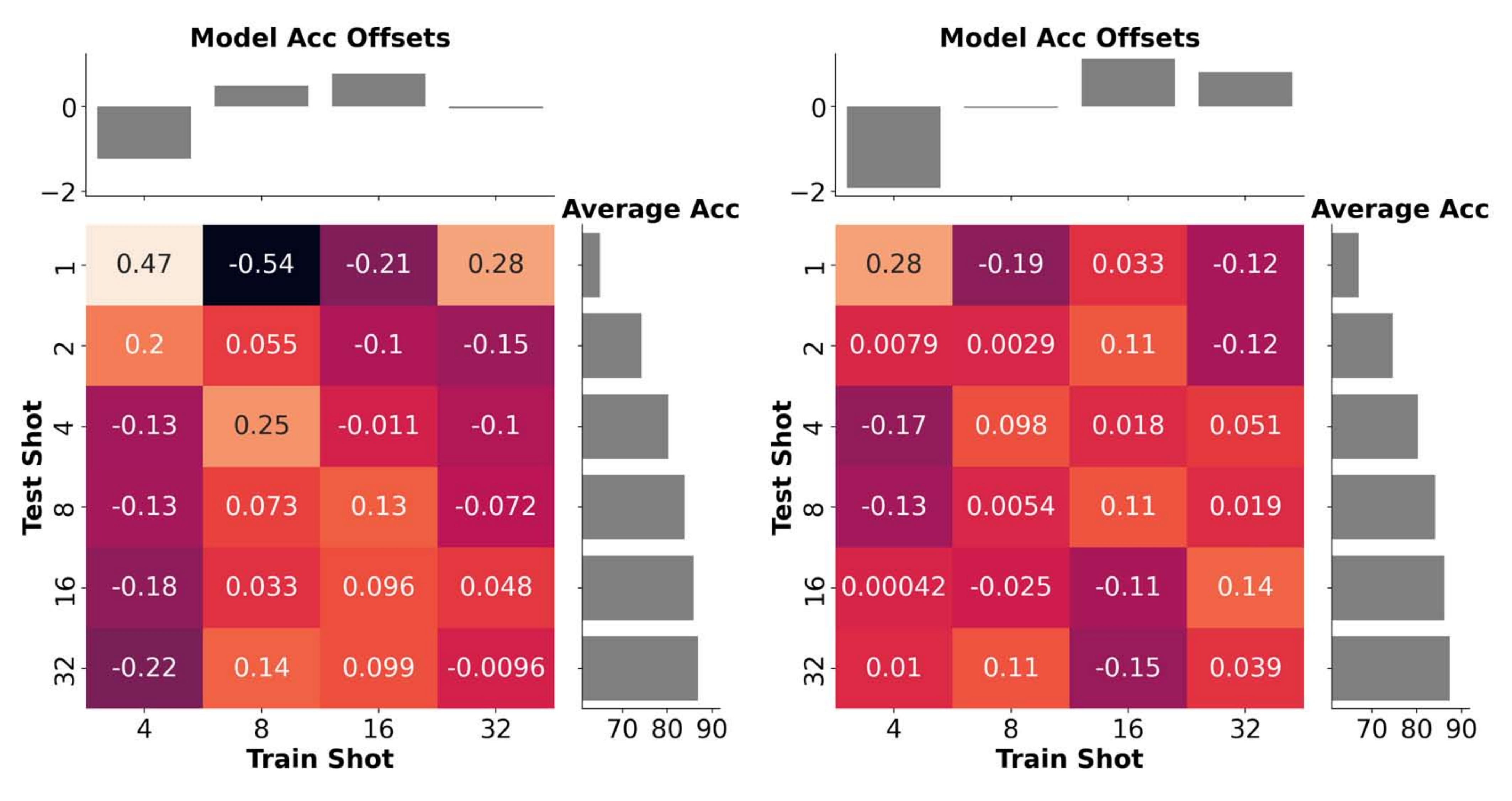}
        \caption{Proto vs Cosine Proto on tiered-ImageNet, both using a ResNet-12 backbone.}
        \label{fig:supp_heatmap_tin_proto}
        \end{figure*}
    
       \begin{table}
      \centering
      \tiny
      \begin{tabular}{c c | c c c c | c}
        \toprule
        \multicolumn{2}{c|}{\textbf{tIN ResNet-12}} &\multicolumn{5}{c}{\textbf{Train Shot}} \\
        \multicolumn{2}{c|}{\textbf{Proto}} & 4 & 8 & 16 & 32 & Mean \\
        \hline
        \Tstrut\multirow{7}{*}{\makecell{\textbf{Test}\\\textbf{Shot}}} 
& 1 & 64.28 & 64.99 & 65.60 & 65.28 & 65.04 \\
& 2 & 73.25 & 74.83 & 74.95 & 74.09 & 74.28 \\
& 4 & 78.90 & 81.00 & 81.02 & 80.12 & 80.26 \\
& 8 & 82.56 & 84.48 & 84.81 & 83.80 & 83.91 \\
& 16 & 84.45 & 86.38 & 86.72 & 85.86 & 85.85 \\
& 32 & 85.46 & 87.54 & 87.78 & 86.86 & 86.91 \\
\cline{2-7}
\Tstrut & Offset & -1.23 & 0.49 & 0.77 & -0.04 & 1.01 \\
        \bottomrule
      \end{tabular}
      \quad
      \begin{tabular}{c c | c c c c | c}
        \toprule
        \multicolumn{2}{c|}{\textbf{tIN ResNet-12}} &\multicolumn{5}{c}{\textbf{Train Shot}} \\
        \multicolumn{2}{c|}{\textbf{Cosine Proto}} & 4 & 8 & 16 & 32 & Mean \\
        \hline
        \Tstrut\multirow{7}{*}{\makecell{\textbf{Test}\\\textbf{Shot}}} 
& 1 & 65.47 & 66.90 & 68.28 & 67.81 & 67.11 \\
& 2 & 72.71 & 74.60 & 75.87 & 75.32 & 74.62 \\
& 4 & 78.16 & 80.32 & 81.40 & 81.12 & 80.25 \\
& 8 & 82.02 & 84.05 & 85.31 & 84.91 & 84.07 \\
& 16 & 84.28 & 86.15 & 87.22 & 87.16 & 86.20 \\
& 32 & 85.49 & 87.48 & 88.38 & 88.26 & 87.40 \\
\cline{2-7}
\Tstrut & Offset & -1.92 & -0.03 & 1.13 & 0.82 & 0.46 \\
        \bottomrule
      \end{tabular}
      \caption{Accuracy comparison for ProtoNet / Cosine ProtoNet on tiered-ImageNet. Average group accuracy is displayed in the rightmost column of each table, model offsets are given in the bottom row. Bottom right cell (intersection of Offset and Mean) shows the sensitivity score.}
      \label{tab:supp_tab_tin_proto}
    \end{table}

        \begin{figure*}
        \centering
        \includegraphics[width=.8\linewidth]{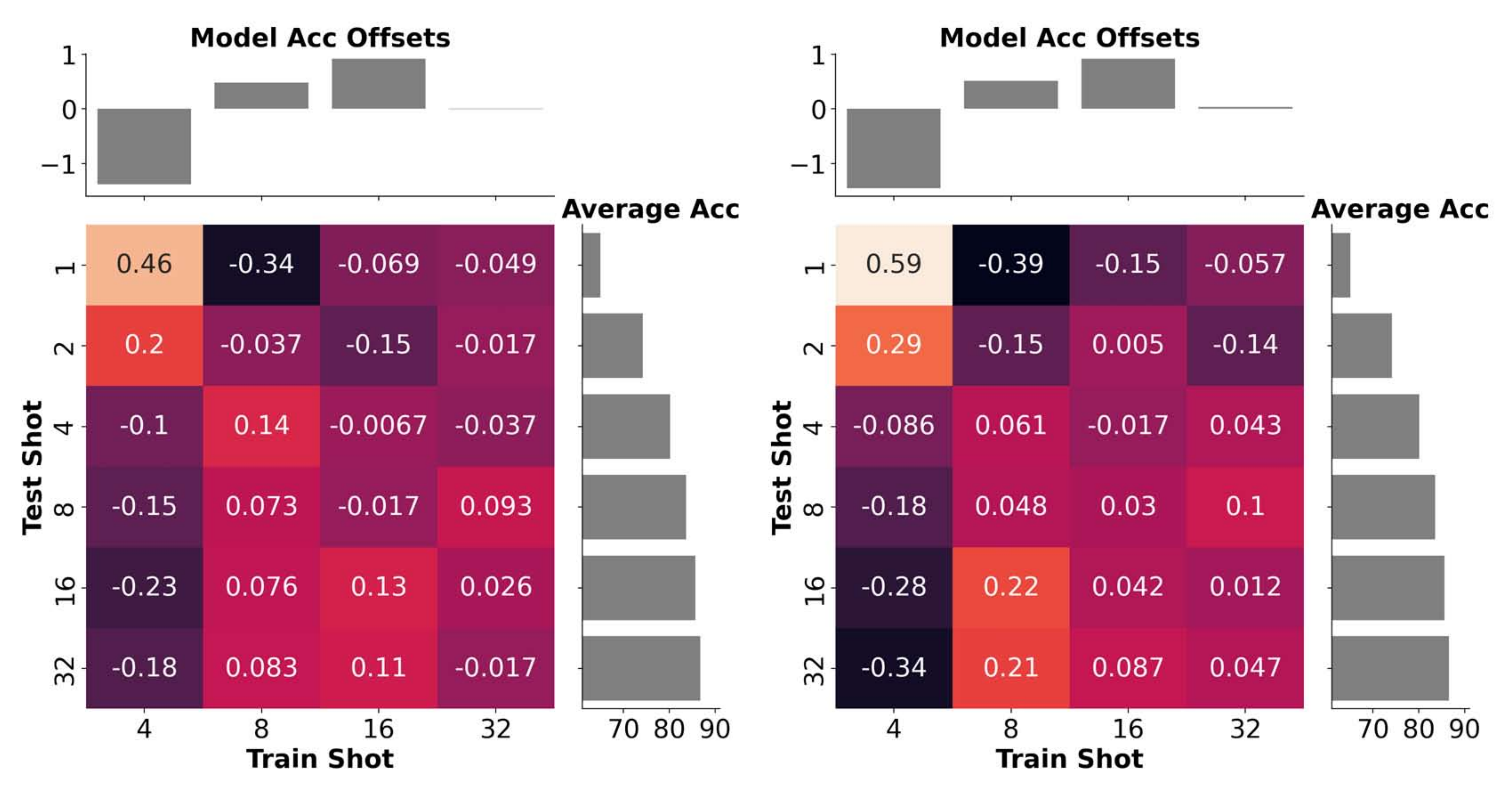}
        \caption{PCA vs EST on tiered-ImageNet, both using a ResNet-12 backbone.}
        \label{fig:supp_heatmap_tin_pca_est}
        \end{figure*}
    
       \begin{table}
      \centering
      \tiny
      \begin{tabular}{c c | c c c c | c}
        \toprule
        \multicolumn{2}{c|}{\textbf{tIN ResNet-12}} &\multicolumn{5}{c}{\textbf{Train Shot}} \\
        \multicolumn{2}{c|}{\textbf{PCA}} & 4 & 8 & 16 & 32 & Mean \\
        \hline
        \Tstrut\multirow{7}{*}{\makecell{\textbf{Test}\\\textbf{Shot}}} 
& 1 & 64.07 & 65.14 & 65.84 & 64.94 & 65.00 \\
& 2 & 73.01 & 74.64 & 74.96 & 74.17 & 74.19 \\
& 4 & 78.71 & 80.82 & 81.10 & 80.15 & 80.19 \\
& 8 & 82.17 & 84.26 & 84.60 & 83.79 & 83.71 \\
& 16 & 84.10 & 86.27 & 86.75 & 85.73 & 85.71 \\
& 32 & 85.16 & 87.29 & 87.75 & 86.70 & 86.73 \\
\cline{2-7}
\Tstrut & Offset & -1.39 & 0.48 & 0.91 & -0.01 & 0.80 \\
        \bottomrule
      \end{tabular}
      \quad
      \begin{tabular}{c c | c c c c | c}
        \toprule
        \multicolumn{2}{c|}{\textbf{tIN ResNet-12}} &\multicolumn{5}{c}{\textbf{Train Shot}} \\
        \multicolumn{2}{c|}{\textbf{EST}} & 4 & 8 & 16 & 32 & Mean \\
        \hline
        \Tstrut\multirow{7}{*}{\makecell{\textbf{Test}\\\textbf{Shot}}} 
& 1 & 64.20 & 65.18 & 65.83 & 65.04 & 65.06 \\
& 2 & 72.94 & 74.47 & 75.03 & 74.00 & 74.11 \\
& 4 & 78.54 & 80.65 & 80.98 & 80.16 & 80.08 \\
& 8 & 81.96 & 84.15 & 84.54 & 83.73 & 83.60 \\
& 16 & 83.85 & 86.31 & 86.54 & 85.63 & 85.58 \\
& 32 & 84.78 & 87.29 & 87.58 & 86.66 & 86.58 \\
\cline{2-7}
\Tstrut & Offset & -1.46 & 0.51 & 0.91 & 0.03 & 0.98 \\
        \bottomrule
      \end{tabular}
      \caption{Accuracy comparison for PCA and EST on tiered-ImageNet. Average group accuracy is displayed in the rightmost column of each table, model offsets are given in the bottom row. Bottom right cell (intersection of Offset and Mean) shows the sensitivity score.}
      \label{tab:supp_tab_tin_pca_est}
    \end{table}
    
    \clearpage

 \begin{figure*}
        \centering
        \includegraphics[width=.8\linewidth]{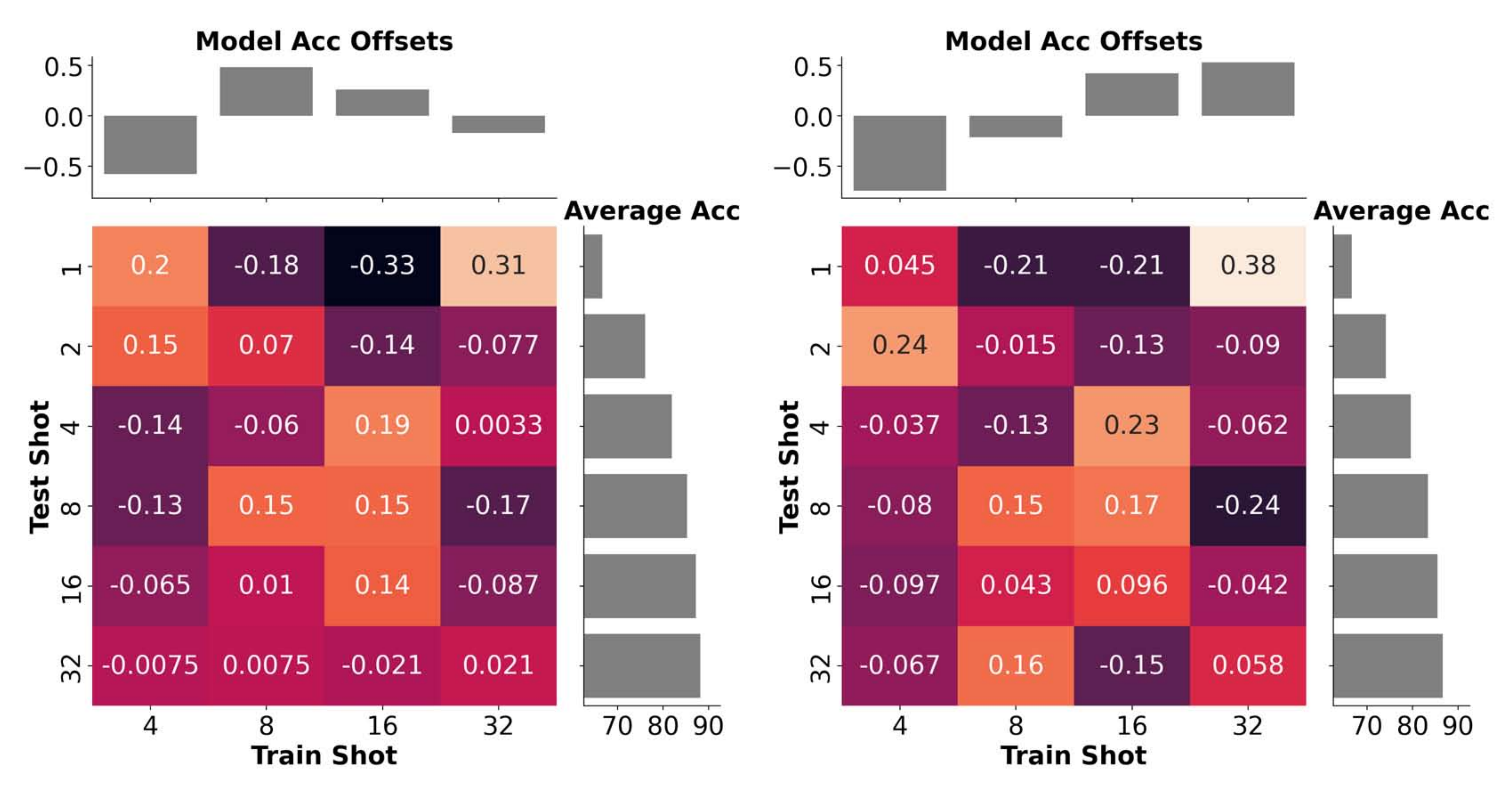}
        \caption{FEAT vs Cosine FEAT on tiered-ImageNet, both using a ResNet-12 backbone.}
        \label{fig:supp_heatmap_tin_feat}
        \end{figure*}
    
       \begin{table}
      \centering
      \tiny
      \begin{tabular}{c c | c c c c | c}
        \toprule
        \multicolumn{2}{c|}{\textbf{tIN ResNet-12}} &\multicolumn{5}{c}{\textbf{Train Shot}} \\
        \multicolumn{2}{c|}{\textbf{FEAT}} & 4 & 8 & 16 & 32 & Mean \\
        \hline
        \Tstrut\multirow{7}{*}{\makecell{\textbf{Test}\\\textbf{Shot}}} 
& 1 & 66.26 & 66.95 & 66.58 & 66.78 & 66.64 \\
& 2 & 75.64 & 76.63 & 76.20 & 75.83 & 76.07 \\
& 4 & 81.20 & 82.34 & 82.37 & 81.75 & 81.92 \\
& 8 & 84.62 & 85.97 & 85.75 & 85.00 & 85.33 \\
& 16 & 86.58 & 87.72 & 87.63 & 86.97 & 87.22 \\
& 32 & 87.62 & 88.70 & 88.45 & 88.06 & 88.21 \\
\cline{2-7}
\Tstrut & Offset & -0.58 & 0.49 & 0.26 & -0.17 & 0.63 \\
        \bottomrule
      \end{tabular}
      \quad
      \begin{tabular}{c c | c c c c | c}
        \toprule
        \multicolumn{2}{c|}{\textbf{tIN ResNet-12}} &\multicolumn{5}{c}{\textbf{Train Shot}} \\
        \multicolumn{2}{c|}{\textbf{Cosine FEAT}} & 4 & 8 & 16 & 32 & Mean \\
        \hline
        \Tstrut\multirow{7}{*}{\makecell{\textbf{Test}\\\textbf{Shot}}} 
& 1 & 65.93 & 66.20 & 66.84 & 67.54 & 66.63 \\
& 2 & 73.60 & 73.88 & 74.40 & 74.55 & 74.11 \\
& 4 & 78.83 & 79.27 & 80.26 & 80.08 & 79.61 \\
& 8 & 82.51 & 83.27 & 83.93 & 83.62 & 83.33 \\
& 16 & 84.61 & 85.28 & 85.97 & 85.94 & 85.45 \\
& 32 & 85.82 & 86.58 & 86.90 & 87.22 & 86.63 \\
\cline{2-7}
\Tstrut & Offset & -0.74 & -0.21 & 0.42 & 0.53 & 0.63 \\
        \bottomrule
      \end{tabular}
      \caption{Accuracy comparison for FEAT and Cosine FEAT on tiered-ImageNet. Average group accuracy is displayed in the rightmost column of each table, model offsets are given in the bottom row. Bottom right cell (intersection of Offset and Mean) shows the sensitivity score.}
      \label{tab:supp_tab_tin_feat}
    \end{table}

       \begin{figure*}
        \centering
        \includegraphics[width=.8\linewidth]{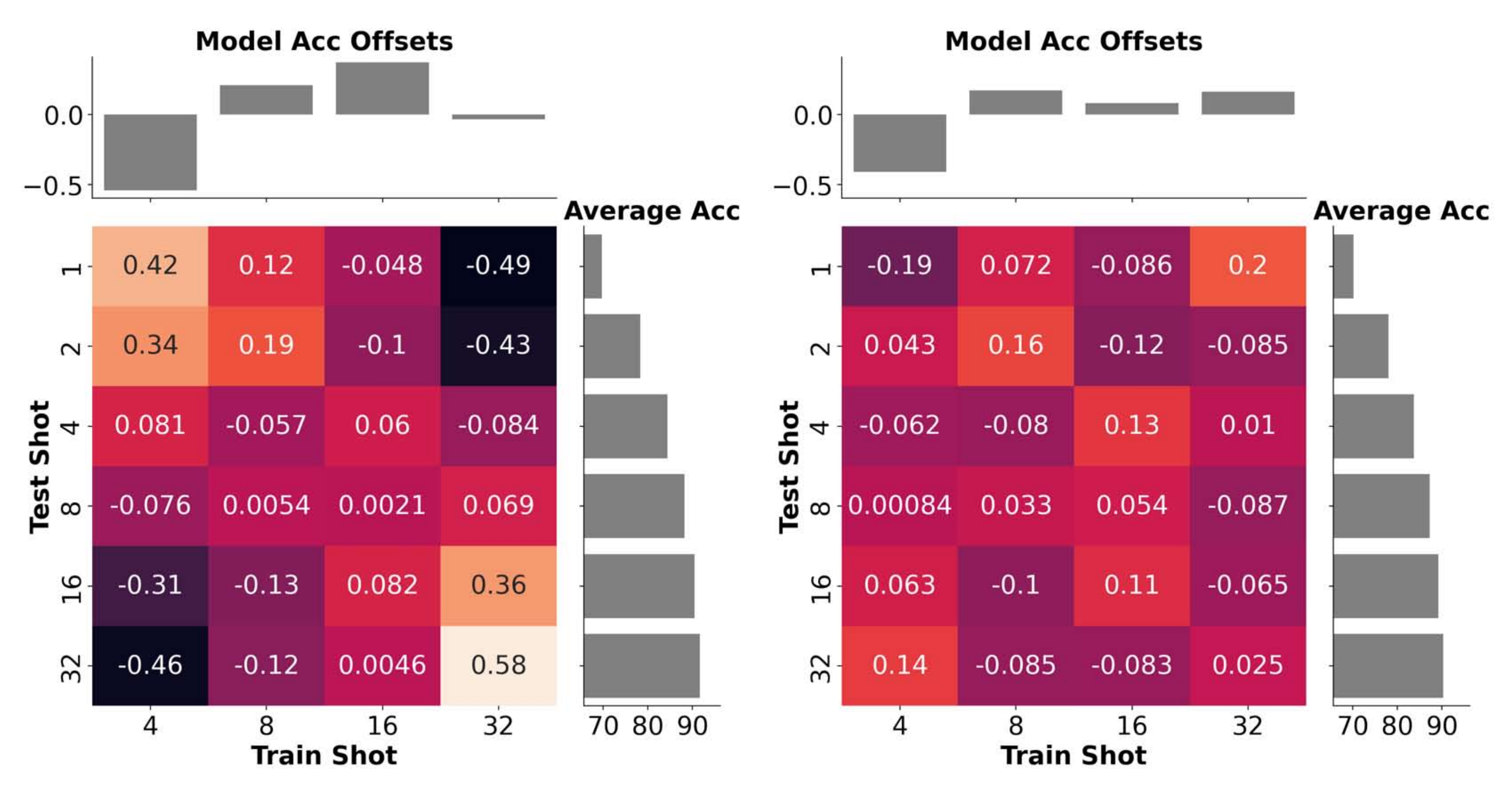}
        \caption{FRN vs Cosine FRN on tiered-ImageNet, both using a ResNet-12 backbone.}
        \label{fig:supp_heatmap_tin_frn}
        \end{figure*}
    
       \begin{table}
      \centering
      \tiny
      \begin{tabular}{c c | c c c c | c}
        \toprule

        \multicolumn{2}{c|}{\textbf{tIN ResNet-12}} &\multicolumn{5}{c}{\textbf{Train Shot}} \\
        \multicolumn{2}{c|}{\textbf{FRN}} & 4 & 8 & 16 & 32 & Mean \\
        \hline
        \Tstrut\multirow{7}{*}{\makecell{\textbf{Test}\\\textbf{Shot}}} 
& 1 & 69.56 & 70.00 & 70.00 & 69.15 & 69.68 \\
& 2 & 78.17 & 78.77 & 78.64 & 77.90 & 78.37 \\
& 4 & 83.94 & 84.55 & 84.83 & 84.28 & 84.40 \\
& 8 & 87.58 & 88.41 & 88.57 & 88.23 & 88.20 \\
& 16 & 89.62 & 90.54 & 90.92 & 90.79 & 90.47 \\
& 32 & 90.70 & 91.79 & 92.08 & 92.25 & 91.71 \\
\cline{2-7}
\Tstrut & Offset & -0.54 & 0.21 & 0.37 & -0.04 & 1.07 \\
        \bottomrule
      \end{tabular}
      \quad
      \begin{tabular}{c c | c c c c | c}
        \toprule
        \multicolumn{2}{c|}{\textbf{tIN ResNet-12}} &\multicolumn{5}{c}{\textbf{Train Shot}} \\
        \multicolumn{2}{c|}{\textbf{Cosine FRN}} & 4 & 8 & 16 & 32 & Mean \\
        \hline
        \Tstrut\multirow{7}{*}{\makecell{\textbf{Test}\\\textbf{Shot}}} 
& 1 & 69.57 & 70.41 & 70.16 & 70.53 & 70.17 \\
& 2 & 77.69 & 78.39 & 78.01 & 78.13 & 78.06 \\
& 4 & 83.23 & 83.79 & 83.91 & 83.87 & 83.70 \\
& 8 & 86.82 & 87.43 & 87.36 & 87.30 & 87.23 \\
& 16 & 88.90 & 89.31 & 89.43 & 89.34 & 89.24 \\
& 32 & 90.02 & 90.37 & 90.28 & 90.47 & 90.28 \\
\cline{2-7}
\Tstrut & Offset & -0.41 & 0.17 & 0.08 & 0.16 & 0.39 \\
        \bottomrule
      \end{tabular}
      \caption{Accuracy comparison for FRN and Cosine FRN on tiered-ImageNet. Average group accuracy is displayed in the rightmost column of each table, model offsets are given in the bottom row. Bottom right cell (intersection of Offset and Mean) shows the sensitivity score.}
      \label{tab:supp_tab_tin_frn}
    \end{table}

\end{document}